\documentclass[conference]{IEEEtran}
\IEEEoverridecommandlockouts

\usepackage{cite}
\usepackage{amsmath,amssymb,amsfonts}
\usepackage{algorithmic}
\usepackage{graphicx}
\usepackage{textcomp}
\usepackage{xcolor}
\usepackage{url}
\usepackage{adjustbox}

\makeatletter
\newcommand{\resdef}[2]{\expandafter\def\csname res@#1\endcsname{#2}}
\newcommand{\Res}[1]{\ifcsname res@#1\endcsname\csname res@#1\endcsname\else\textcolor{red}{??#1}\@latex@warning{Undefined result #1}\fi}
\makeatother
\resdef{ablation/ENA-Combined/adapt/Train-Env-Steps/mean}{4,920,196}
\resdef{ablation/ENA-Combined/compute/candsteps/mean}{4,226,921}
\resdef{ablation/ENA-Combined/compute/candsteps/std}{596,722}
\resdef{ablation/ENA-Combined/detect/test/Brown-dwarf/changes/mean}{16.4}
\resdef{ablation/ENA-Combined/detect/test/Brown-dwarf/changes/std}{18.8}
\resdef{ablation/ENA-Combined/hof/gladiators/max}{24}
\resdef{ablation/ENA-Combined/hof/gladiators/mean}{16.4}
\resdef{ablation/ENA-Combined/hof/gladiators/min}{10}
\resdef{ablation/ENA-Combined/hof/niches/mean}{3.8}
\resdef{ablation/ENA-Combined/recog/Brown-dwarf/unrecognized}{30}
\resdef{ablation/ENA-Combined/recog/Neptune/Earth-Standard}{30}
\resdef{ablation/ENA-Combined/retention/Jupiter-Heavy/mean}{97.4}
\resdef{ablation/ENA-Combined/retention/Mars-Light/mean}{101.5}
\resdef{ablation/ENA-Combined/test/Brown-dwarf/mean}{367.2}
\resdef{ablation/ENA-Combined/test/Brown-dwarf/std}{153.9}
\resdef{ablation/ENA-Combined/test/Neptune/mean}{490.4}
\resdef{ablation/ENA-Combined/test/rel/mean}{83.2}
\resdef{ablation/ENA-Combined/test/rel/std}{16.1}
\resdef{ablation/ENA-Combined/time/min/mean}{38.9}
\resdef{ablation/ENA-HoF+EnvDetection/adapt/Train-Env-Steps/mean}{343,189}
\resdef{ablation/ENA-HoF+EnvDetection/hof/gladiators/max}{23}
\resdef{ablation/ENA-HoF+EnvDetection/hof/gladiators/mean}{17.9}
\resdef{ablation/ENA-HoF+EnvDetection/hof/gladiators/min}{15}
\resdef{ablation/ENA-HoF+EnvDetection/hof/niches/mean}{3.9}
\resdef{ablation/ENA-HoF+EnvDetection/time/min/mean}{7.5}
\resdef{ablation/ENA-HoF-vs-ENA-HoF+EnvDetection/compute/candsteps/ci/high}{1}
\resdef{ablation/ENA-HoF-vs-ENA-HoF+EnvDetection/compute/candsteps/ci/low}{\ensuremath{-}222}
\resdef{ablation/ENA-HoF-vs-ENA-HoF+EnvDetection/compute/candsteps/diff}{\ensuremath{-}111}
\resdef{ablation/ENA-HoF-vs-ENA-HoF+EnvDetection/compute/candsteps/p}{$p=0.051$}
\resdef{ablation/ENA-HoF-vs-ENA-HoF+EnvDetection/test/rel/d}{0.59}
\resdef{ablation/ENA-HoF-vs-ENA-Plasticity+HoF/test/rel/d}{1.25}
\resdef{ablation/ENA-HoF/adapt/Train-Env-Steps/mean}{267,760}
\resdef{ablation/ENA-HoF/time/min/mean}{6.2}
\resdef{ablation/ENA-NoSkill-vs-ENA-Combined/test/rel/ci/high}{55.1}
\resdef{ablation/ENA-NoSkill-vs-ENA-Combined/test/rel/ci/low}{34.2}
\resdef{ablation/ENA-NoSkill-vs-ENA-Combined/test/rel/d}{1.60}
\resdef{ablation/ENA-NoSkill-vs-ENA-Combined/test/rel/diff}{+44.7}
\resdef{ablation/ENA-NoSkill-vs-ENA-Combined/test/rel/p}{$p<0.001$}
\resdef{ablation/ENA-NoSkill-vs-ENA-HoF+EnvDetection/test/rel/ci/high}{24.1}
\resdef{ablation/ENA-NoSkill-vs-ENA-HoF+EnvDetection/test/rel/ci/low}{\ensuremath{-}4.7}
\resdef{ablation/ENA-NoSkill-vs-ENA-HoF+EnvDetection/test/rel/diff}{+9.7}
\resdef{ablation/ENA-NoSkill-vs-ENA-HoF+EnvDetection/test/rel/p}{$p=0.178$}
\resdef{ablation/ENA-NoSkill-vs-ENA-HoF/compute/candsteps/ci/high}{518}
\resdef{ablation/ENA-NoSkill-vs-ENA-HoF/compute/candsteps/ci/low}{\ensuremath{-}64}
\resdef{ablation/ENA-NoSkill-vs-ENA-HoF/compute/candsteps/diff}{+227}
\resdef{ablation/ENA-NoSkill-vs-ENA-HoF/compute/candsteps/p}{$p=0.122$}
\resdef{ablation/ENA-NoSkill-vs-ENA-HoF/test/rel/d}{\ensuremath{-}0.46}
\resdef{ablation/ENA-NoSkill-vs-ENA-HoF/test/rel/diff}{\ensuremath{-}13.8}
\resdef{ablation/ENA-NoSkill-vs-ENA-HoF/test/rel/nreq}{64}
\resdef{ablation/ENA-NoSkill-vs-ENA-HoF/test/rel/pholm}{$p=0.036$}
\resdef{ablation/ENA-NoSkill/adapt/Train-Env-Steps/mean}{266,057}
\resdef{ablation/ENA-NoSkill/compute/candsteps/mean}{28,145}
\resdef{ablation/ENA-NoSkill/end/Jupiter-Heavy/mean}{71}
\resdef{ablation/ENA-NoSkill/test/Jupiter-Heavy/mean}{144.8}
\resdef{ablation/ENA-NoSkill/test/rel/mean}{38.5}
\resdef{ablation/ENA-NoSkill/test/rel/std}{20.0}
\resdef{ablation/ENA-NoSkill/time/min/mean}{6.0}
\resdef{ablation/ENA-Plasticity+HoF-vs-ENA-Combined/compute/candsteps/ci/high}{339,337}
\resdef{ablation/ENA-Plasticity+HoF-vs-ENA-Combined/compute/candsteps/ci/low}{\ensuremath{-}315,239}
\resdef{ablation/ENA-Plasticity+HoF-vs-ENA-Combined/compute/candsteps/diff}{+12,049}
\resdef{ablation/ENA-Plasticity+HoF-vs-ENA-Combined/compute/candsteps/p}{$p=0.940$}
\resdef{ablation/ENA-Plasticity+HoF/adapt/Train-Env-Steps/mean}{4,893,572}
\resdef{ablation/ENA-Plasticity+HoF/retention/Jupiter-Heavy/mean}{58.8}
\resdef{ablation/ENA-Plasticity+HoF/retention/Mars-Light/mean}{100.5}
\resdef{ablation/ENA-Plasticity-vs-ENA-Combined/test/rel/ci/high}{27.5}
\resdef{ablation/ENA-Plasticity-vs-ENA-Combined/test/rel/ci/low}{12.3}
\resdef{ablation/ENA-Plasticity-vs-ENA-Combined/test/rel/diff}{+19.9}
\resdef{ablation/ENA-Plasticity-vs-ENA-Combined/test/rel/p}{$p<0.001$}
\resdef{ablation/ENA-Plasticity-vs-ENA-Plasticity+HoF/compute/candsteps/ci/high}{822,475}
\resdef{ablation/ENA-Plasticity-vs-ENA-Plasticity+HoF/compute/candsteps/ci/low}{\ensuremath{-}76,157}
\resdef{ablation/ENA-Plasticity-vs-ENA-Plasticity+HoF/compute/candsteps/diff}{+373,159}
\resdef{ablation/ENA-Plasticity-vs-ENA-Plasticity+HoF/compute/candsteps/p}{$p=0.100$}
\resdef{ablation/ENA-Plasticity-vs-ENA-Plasticity+HoF/test/rel/d}{\ensuremath{-}0.19}
\resdef{ablation/ENA-Plasticity-vs-ENA-Plasticity+HoF/test/rel/diff}{\ensuremath{-}4.4}
\resdef{ablation/ENA-Plasticity-vs-ENA-Plasticity+HoF/test/rel/nreq}{356}
\resdef{ablation/ENA-Plasticity-vs-ENA-Plasticity+HoF/test/rel/pholm}{$p=0.311$}
\resdef{ablation/ENA-Plasticity/adapt/Train-Env-Steps/mean}{4,532,688}
\resdef{ablation/ENA-Plasticity/compute/candsteps/mean}{3,841,713}
\resdef{ablation/ENA-Plasticity/retention/Jupiter-Heavy/mean}{75.4}
\resdef{ablation/ENA-Plasticity/retention/Mars-Light/mean}{58.1}
\resdef{ablation/ENA-Plasticity/time/min/mean}{34.9}
\resdef{ablation/family}{7}
\resdef{ablation/mde}{0.69}
\resdef{ablation/n}{30}
\resdef{ablation/plannedmde}{0.94}
\resdef{cfg/cartpole/maxsteps}{500}
\resdef{cfg/cartpole/seedoffsetraw}{1000}
\resdef{cfg/cartpole/solved}{475}
\resdef{cfg/cartpole/test}{200}
\resdef{cfg/cartpole/train}{500}
\resdef{cfg/challenge/budget}{2,500,000}
\resdef{cfg/challenge/checkevery}{5,000}
\resdef{cfg/challenge/evalepisodes}{5}
\resdef{cfg/challenge/mlpnet}{64 $\times$ 64 ReLU}
\resdef{cfg/challenge/pponet}{64 $\times$ 64 tanh}
\resdef{cfg/challenge/sacdratio}{0.5}
\resdef{cfg/cheetah/baselinenet}{256 $\times$ 256 ReLU}
\resdef{cfg/cheetah/budget}{1,000,000}
\resdef{cfg/cheetah/chirp/bands}{8}
\resdef{cfg/cheetah/chirp/excitation}{wavelet train}
\resdef{cfg/cheetah/chirp/excitationwavelet}{Mexican hat}
\resdef{cfg/cheetah/chirp/featurewavelet}{Haar}
\resdef{cfg/cheetah/chirp/level}{3}
\resdef{cfg/cheetah/chirp/margin}{3}
\resdef{cfg/cheetah/chirp/minchirps}{8}
\resdef{cfg/cheetah/chirp/share}{6.4}
\resdef{cfg/cheetah/chirp/steps}{64}
\resdef{cfg/cheetah/chirp/transform}{wavelet packet}
\resdef{cfg/cheetah/chirp/window}{4}
\resdef{cfg/cheetah/maxeval}{200}
\resdef{cfg/cheetah/maxsteps}{1,000}
\resdef{cfg/cheetah/rampminsteps}{29}
\resdef{cfg/cheetah/solved}{4,800}
\resdef{cfg/cheetah/test}{50}
\resdef{cfg/cheetahena/brain}{32 $\times$ 16 tanh}
\resdef{cfg/cheetahena/halloffamedriver}{olympic}
\resdef{cfg/cheetahena/maxevalsteps}{200}
\resdef{cfg/cheetahena/maxperniche}{10}
\resdef{cfg/cheetahena/mutationrate}{0.25}
\resdef{cfg/cheetahena/olympicsepisodes}{3}
\resdef{cfg/cheetahena/olympicsinterval}{200,000}
\resdef{cfg/cheetahena/olympicsqualifierseats}{2}
\resdef{cfg/cheetahppo/batchsize}{64}
\resdef{cfg/cheetahppo/clipratio}{0.1}
\resdef{cfg/cheetahppo/entropycoef}{$4.01762\times10^{-4}$}
\resdef{cfg/cheetahppo/gaelambda}{0.92}
\resdef{cfg/cheetahppo/gamma}{0.98}
\resdef{cfg/cheetahppo/learningrate}{$2.0633\times10^{-5}$}
\resdef{cfg/cheetahppo/logstdinit}{\ensuremath{-}2}
\resdef{cfg/cheetahppo/maxgradnorm}{0.8}
\resdef{cfg/cheetahppo/nepochs}{20}
\resdef{cfg/cheetahppo/nsteps}{512}
\resdef{cfg/cheetahppo/valuecoef}{0.58096}
\resdef{cfg/cheetahsac/batchsize}{256}
\resdef{cfg/cheetahsac/buffercapacity}{1,000,000}
\resdef{cfg/cheetahsac/gamma}{0.99}
\resdef{cfg/cheetahsac/gradientsteps}{20}
\resdef{cfg/cheetahsac/learningrate}{$3\times10^{-4}$}
\resdef{cfg/cheetahsac/learningstarts}{10,000}
\resdef{cfg/cheetahsac/tau}{0.005}
\resdef{cfg/cheetahsac/updatefrequency}{20}
\resdef{cfg/detbench/block}{4}
\resdef{cfg/detbench/confirmseeds}{30}
\resdef{cfg/detbench/episodes}{40}
\resdef{cfg/detbench/purity}{90}
\resdef{cfg/detbench/transition/minsamples}{1,000}
\resdef{cfg/detbench/transition/minverifysamples}{1,000}
\resdef{cfg/detbench/transition/noveltymargin}{3}
\resdef{cfg/detbench/window}{5}
\resdef{cfg/dqn/batchsize}{64}
\resdef{cfg/dqn/buffersize}{50,000}
\resdef{cfg/dqn/epsilondecaysteps}{10,000}
\resdef{cfg/dqn/epsilonend}{0.01}
\resdef{cfg/dqn/epsilonstart}{1}
\resdef{cfg/dqn/gamma}{0.99}
\resdef{cfg/dqn/learningrate}{$3\times10^{-4}$}
\resdef{cfg/dqn/learningstarts}{1,000}
\resdef{cfg/dqn/targetupdatefreq}{1,000}
\resdef{cfg/dqn/trainfreq}{4}
\resdef{cfg/ena/bpr/crossprior}{0.5}
\resdef{cfg/ena/bpr/mix}{0.95}
\resdef{cfg/ena/bpr/noise}{0.2}
\resdef{cfg/ena/bpr/ownprior}{1}
\resdef{cfg/ena/bpr/prior}{1}
\resdef{cfg/ena/bpr/switch}{0.05}
\resdef{cfg/ena/brain}{4 $\times$ 2 ReLU}
\resdef{cfg/ena/crossover/share}{0.5}
\resdef{cfg/ena/detect/gammaexp}{8}
\resdef{cfg/ena/detect/margin}{2}
\resdef{cfg/ena/detect/minsamples}{100}
\resdef{cfg/ena/detect/recent}{500}
\resdef{cfg/ena/detect/tau}{3}
\resdef{cfg/ena/detect/verify}{50}
\resdef{cfg/ena/detect/windows}{2, 4}
\resdef{cfg/ena/epsilon}{0.2}
\resdef{cfg/ena/fuzzy/rules}{9}
\resdef{cfg/ena/fuzzy/trendbefore}{45}
\resdef{cfg/ena/fuzzy/trendmin}{20}
\resdef{cfg/ena/fuzzy/trendrecent}{5}
\resdef{cfg/ena/history}{100}
\resdef{cfg/ena/hof/K}{5}
\resdef{cfg/ena/hof/gatefloor}{0.1}
\resdef{cfg/ena/hof/gaterelief}{0.5}
\resdef{cfg/ena/hof/gatetrust}{0.5}
\resdef{cfg/ena/hof/gatez}{1.5}
\resdef{cfg/ena/hof/reinject}{0.5}
\resdef{cfg/ena/hof/similarity}{0.95}
\resdef{cfg/ena/mutation/minnoise}{0.01}
\resdef{cfg/ena/mutation/noise}{0.1}
\resdef{cfg/ena/mutation/rate}{0.05}
\resdef{cfg/ena/pop}{100}
\resdef{cfg/ena/search/crossovers}{40}
\resdef{cfg/ena/search/crossoverspop}{32}
\resdef{cfg/ena/search/elite}{20}
\resdef{cfg/ena/search/fresh}{20}
\resdef{cfg/ena/search/freshpop}{16}
\resdef{cfg/ena/search/mincandidates}{2}
\resdef{cfg/ena/search/mutants}{40}
\resdef{cfg/ena/search/mutantspop}{32}
\resdef{cfg/ena/search/rampmincandidates}{10}
\resdef{cfg/ena/search/rampminsteps}{59}
\resdef{cfg/ena/search/replaced}{80}
\resdef{cfg/ena/signal/min}{0.1}
\resdef{cfg/ena/signal/percentile}{90}
\resdef{cfg/ena/signal/warmup}{10}
\resdef{cfg/ena/smax}{500}
\resdef{cfg/ena/smin}{10}
\resdef{cfg/ena/trust/high}{5}
\resdef{cfg/ena/trust/low}{\ensuremath{-}2}
\resdef{cfg/ena/trust/phigh}{95}
\resdef{cfg/ena/trust/plow}{5}
\resdef{cfg/ena/trust/reset}{0.1}
\resdef{cfg/plot/band}{50}
\resdef{cfg/ppo/anneallr}{no}
\resdef{cfg/ppo/batchsize}{64}
\resdef{cfg/ppo/clipratio}{0.2}
\resdef{cfg/ppo/entropycoef}{0.01}
\resdef{cfg/ppo/gaelambda}{0.95}
\resdef{cfg/ppo/gamma}{0.99}
\resdef{cfg/ppo/learningrate}{$3\times10^{-4}$}
\resdef{cfg/ppo/maxgradnorm}{0.5}
\resdef{cfg/ppo/nepochs}{10}
\resdef{cfg/ppo/nsteps}{2,048}
\resdef{cfg/ppo/valuecoef}{0.5}
\resdef{cfg/sacd/cleanrlratio}{0.89}
\resdef{cfg/sacd/paperratio}{0.98}
\resdef{cfg/sacdiscrete/batchsize}{64}
\resdef{cfg/sacdiscrete/buffercapacity}{100,000}
\resdef{cfg/sacdiscrete/gamma}{0.99}
\resdef{cfg/sacdiscrete/learningrate}{$3\times10^{-4}$}
\resdef{cfg/sacdiscrete/learningstarts}{1,000}
\resdef{cfg/sacdiscrete/targetentropyratio}{0.5}
\resdef{cfg/sacdiscrete/tau}{0.005}
\resdef{cfg/sacdiscrete/trainfreq}{4}
\resdef{cfg/software/gymnasium}{1.3.0}
\resdef{cfg/software/keras}{3.15.1}
\resdef{cfg/software/mujoco}{3.13.0}
\resdef{cfg/software/numpy}{2.4.6}
\resdef{cfg/software/python}{3.11}
\resdef{cfg/software/pywavelets}{1.8.0}
\resdef{cfg/software/scipy}{1.17.1}
\resdef{cfg/software/tensorflow}{2.21.0}
\resdef{challenge/DQN-vs-ENA-Combined/adapt/Earth-Standard:steps/p}{$p<0.001$}
\resdef{challenge/DQN-vs-ENA-Combined/adapt/Jupiter-Heavy:steps/p}{$p=0.625$}
\resdef{challenge/DQN-vs-ENA-Combined/adapt/Mars-Light:steps/p}{$p<0.001$}
\resdef{challenge/DQN-vs-ENA-Combined/test/rel/ci/high}{41.7}
\resdef{challenge/DQN-vs-ENA-Combined/test/rel/ci/low}{24.8}
\resdef{challenge/DQN-vs-ENA-Combined/test/rel/d}{1.12}
\resdef{challenge/DQN-vs-ENA-Combined/test/rel/diff}{+33.3}
\resdef{challenge/DQN-vs-ENA-Combined/test/rel/pholm}{$p<0.001$}
\resdef{challenge/DQN/adapt/Earth-Standard:steps/median}{5,476}
\resdef{challenge/DQN/adapt/Jupiter-Heavy:steps/median}{122,843}
\resdef{challenge/DQN/adapt/Mars-Light:steps/median}{257,626}
\resdef{challenge/DQN/end/Jupiter-Heavy/mean}{404}
\resdef{challenge/DQN/episodes/perenv}{6,759}
\resdef{challenge/DQN/forgetting/Jupiter-Heavy/mean}{329}
\resdef{challenge/DQN/retention/Jupiter-Heavy/mean}{19.9}
\resdef{challenge/DQN/retention/Mars-Light/mean}{98.7}
\resdef{challenge/DQN/test/Jupiter-Heavy/mean}{75.2}
\resdef{challenge/DQN/test/rel/mean}{50.6}
\resdef{challenge/DQN/test/trainrel/mean}{45.9}
\resdef{challenge/DQN/time/min/mean}{97.1}
\resdef{challenge/ENA-Combined/adapt/Earth-Standard:steps/median}{39,288}
\resdef{challenge/ENA-Combined/adapt/Jupiter-Heavy:steps/median}{106,116}
\resdef{challenge/ENA-Combined/adapt/Jupiter-Heavy:steps/unsolved}{1}
\resdef{challenge/ENA-Combined/adapt/Mars-Light:steps/median}{44,841}
\resdef{challenge/ENA-Combined/candshare/mean}{84.9}
\resdef{challenge/ENA-Combined/end/Jupiter-Heavy/mean}{468}
\resdef{challenge/ENA-Combined/episodes/perenv}{815}
\resdef{challenge/ENA-Combined/forgetting/Jupiter-Heavy/mean}{7}
\resdef{challenge/ENA-Combined/played/Jupiter-Heavy/Jupiter-Heavy}{45}
\resdef{challenge/ENA-Combined/recog/Jupiter-Heavy/Jupiter-Heavy}{50}
\resdef{challenge/ENA-Combined/retention/Jupiter-Heavy/mean}{98.6}
\resdef{challenge/ENA-Combined/retention/Mars-Light/mean}{103.2}
\resdef{challenge/ENA-Combined/test/Brown-dwarf/mean}{381.6}
\resdef{challenge/ENA-Combined/test/Earth-Standard/mean}{498.1}
\resdef{challenge/ENA-Combined/test/Jupiter-Heavy/mean}{461.3}
\resdef{challenge/ENA-Combined/test/Mars-Light/mean}{480.2}
\resdef{challenge/ENA-Combined/test/Neptune/mean}{492.9}
\resdef{challenge/ENA-Combined/test/rel/mean}{83.9}
\resdef{challenge/ENA-Combined/test/rel/std}{11.7}
\resdef{challenge/ENA-Combined/test/trainrel/mean}{88.6}
\resdef{challenge/ENA-Combined/time/min/mean}{65.3}
\resdef{challenge/PPO-vs-ENA-Combined/adapt/Earth-Standard:steps/p}{$p<0.001$}
\resdef{challenge/PPO-vs-ENA-Combined/adapt/Jupiter-Heavy:steps/p}{$p<0.001$}
\resdef{challenge/PPO-vs-ENA-Combined/adapt/Mars-Light:steps/p}{$p<0.001$}
\resdef{challenge/PPO-vs-ENA-Combined/test/Brown-dwarf/p}{$p=0.950$}
\resdef{challenge/PPO-vs-ENA-Combined/test/rel/ci/high}{14.4}
\resdef{challenge/PPO-vs-ENA-Combined/test/rel/ci/low}{5.7}
\resdef{challenge/PPO-vs-ENA-Combined/test/rel/d}{0.65}
\resdef{challenge/PPO-vs-ENA-Combined/test/rel/diff}{+10.1}
\resdef{challenge/PPO-vs-ENA-Combined/test/rel/pholm}{$p<0.001$}
\resdef{challenge/PPO/adapt/Earth-Standard:steps/median}{5,000}
\resdef{challenge/PPO/adapt/Jupiter-Heavy:steps/median}{47,743}
\resdef{challenge/PPO/adapt/Mars-Light:steps/median}{10,018}
\resdef{challenge/PPO/end/Jupiter-Heavy/mean}{498}
\resdef{challenge/PPO/episodes/perenv}{5,791}
\resdef{challenge/PPO/forgetting/Jupiter-Heavy/mean}{477}
\resdef{challenge/PPO/retention/Jupiter-Heavy/mean}{4.3}
\resdef{challenge/PPO/retention/Mars-Light/mean}{101.1}
\resdef{challenge/PPO/test/Brown-dwarf/mean}{379.5}
\resdef{challenge/PPO/test/Jupiter-Heavy/mean}{21.6}
\resdef{challenge/PPO/test/rel/mean}{73.8}
\resdef{challenge/PPO/test/trainrel/mean}{65.5}
\resdef{challenge/PPO/time/min/mean}{140.3}
\resdef{challenge/SAC-Discrete-vs-ENA-Combined/adapt/Earth-Standard:steps/p}{$p<0.001$}
\resdef{challenge/SAC-Discrete-vs-ENA-Combined/adapt/Jupiter-Heavy:steps/p}{$p=0.126$}
\resdef{challenge/SAC-Discrete-vs-ENA-Combined/adapt/Mars-Light:steps/p}{$p=0.071$}
\resdef{challenge/SAC-Discrete-vs-ENA-Combined/test/Brown-dwarf/p}{$p=0.694$}
\resdef{challenge/SAC-Discrete-vs-ENA-Combined/test/Earth-Standard/p}{$p=0.049$}
\resdef{challenge/SAC-Discrete-vs-ENA-Combined/test/Mars-Light/p}{$p=0.032$}
\resdef{challenge/SAC-Discrete-vs-ENA-Combined/test/Neptune/p}{$p<0.001$}
\resdef{challenge/SAC-Discrete-vs-ENA-Combined/test/rel/ci/high}{14.0}
\resdef{challenge/SAC-Discrete-vs-ENA-Combined/test/rel/ci/low}{4.7}
\resdef{challenge/SAC-Discrete-vs-ENA-Combined/test/rel/d}{0.57}
\resdef{challenge/SAC-Discrete-vs-ENA-Combined/test/rel/diff}{+9.4}
\resdef{challenge/SAC-Discrete-vs-ENA-Combined/test/rel/pholm}{$p<0.001$}
\resdef{challenge/SAC-Discrete/adapt/Earth-Standard:steps/median}{5,000}
\resdef{challenge/SAC-Discrete/adapt/Jupiter-Heavy:steps/median}{162,720}
\resdef{challenge/SAC-Discrete/adapt/Mars-Light:steps/median}{127,650}
\resdef{challenge/SAC-Discrete/end/Jupiter-Heavy/mean}{493}
\resdef{challenge/SAC-Discrete/episodes/perenv}{5,927}
\resdef{challenge/SAC-Discrete/forgetting/Jupiter-Heavy/mean}{441}
\resdef{challenge/SAC-Discrete/retention/Jupiter-Heavy/mean}{10.5}
\resdef{challenge/SAC-Discrete/retention/Mars-Light/mean}{100.1}
\resdef{challenge/SAC-Discrete/test/Brown-dwarf/mean}{393.5}
\resdef{challenge/SAC-Discrete/test/Earth-Standard/mean}{500.0}
\resdef{challenge/SAC-Discrete/test/Jupiter-Heavy/mean}{52.1}
\resdef{challenge/SAC-Discrete/test/Mars-Light/mean}{495.7}
\resdef{challenge/SAC-Discrete/test/Neptune/mean}{500.0}
\resdef{challenge/SAC-Discrete/test/rel/mean}{74.5}
\resdef{challenge/SAC-Discrete/test/trainrel/mean}{66.3}
\resdef{challenge/SAC-Discrete/time/min/mean}{119.0}
\resdef{challenge/family}{3}
\resdef{challenge/n}{50}
\resdef{cheetah/ENA-Combined/compute/candsteps/mean}{2,376,493}
\resdef{cheetah/ENA-Combined/detect/test/Ice-Floor/unrec/mean}{52.2}
\resdef{cheetah/ENA-Combined/detect/test/changes/total}{9.1}
\resdef{cheetah/ENA-Combined/detect/train/changes/total}{3.9}
\resdef{cheetah/ENA-Combined/end/Earth-Standard/mean}{1,805}
\resdef{cheetah/ENA-Combined/end/Heavy-Rough/mean}{2,133}
\resdef{cheetah/ENA-Combined/end/Mars-Slippery/mean}{1,143}
\resdef{cheetah/ENA-Combined/episodes/perenv}{214}
\resdef{cheetah/ENA-Combined/forgetting/left/mean}{\ensuremath{-}239}
\resdef{cheetah/ENA-Combined/forgetting/left/std}{229}
\resdef{cheetah/ENA-Combined/played/Heavy-Rough/Heavy-Rough}{20}
\resdef{cheetah/ENA-Combined/recog/Earth-Standard/Earth-Standard}{20}
\resdef{cheetah/ENA-Combined/recog/Heavy-Rough/Heavy-Rough}{17}
\resdef{cheetah/ENA-Combined/recog/High-Wind/Earth-Standard}{20}
\resdef{cheetah/ENA-Combined/recog/Mars-Slippery/Mars-Slippery}{18}
\resdef{cheetah/ENA-Combined/recog/distinct}{20}
\resdef{cheetah/ENA-Combined/retention/left/mean}{115.6}
\resdef{cheetah/ENA-Combined/retention/left/std}{16.6}
\resdef{cheetah/ENA-Combined/test/Earth-Standard/mean}{2,000}
\resdef{cheetah/ENA-Combined/test/Heavy-Rough/mean}{2,417}
\resdef{cheetah/ENA-Combined/test/High-Wind/mean}{996}
\resdef{cheetah/ENA-Combined/test/Ice-Floor/mean}{1,915}
\resdef{cheetah/ENA-Combined/test/Mars-Slippery/mean}{1,773}
\resdef{cheetah/ENA-Combined/test/rel/mean}{0.0}
\resdef{cheetah/ENA-Combined/time/min/mean}{32.0}
\resdef{cheetah/ENA-Combined/time/min/std}{1.7}
\resdef{cheetah/PPO-vs-ENA-Combined/forgetting/left/ci/high}{\ensuremath{-}2,654}
\resdef{cheetah/PPO-vs-ENA-Combined/forgetting/left/ci/low}{\ensuremath{-}3,651}
\resdef{cheetah/PPO-vs-ENA-Combined/forgetting/left/diff}{\ensuremath{-}3,152}
\resdef{cheetah/PPO-vs-ENA-Combined/forgetting/left/pholm}{$p<0.001$}
\resdef{cheetah/PPO-vs-ENA-Combined/retention/left/ci/high}{89.9}
\resdef{cheetah/PPO-vs-ENA-Combined/retention/left/ci/low}{62.8}
\resdef{cheetah/PPO-vs-ENA-Combined/retention/left/d}{2.63}
\resdef{cheetah/PPO-vs-ENA-Combined/retention/left/diff}{+76.3}
\resdef{cheetah/PPO-vs-ENA-Combined/retention/left/pholm}{$p<0.001$}
\resdef{cheetah/PPO-vs-ENA-Combined/test/Heavy-Rough/diff}{+1,115}
\resdef{cheetah/PPO-vs-ENA-Combined/test/Heavy-Rough/p}{$p<0.001$}
\resdef{cheetah/PPO/end/Heavy-Rough/mean}{4,793}
\resdef{cheetah/PPO/episodes/perenv}{1,000}
\resdef{cheetah/PPO/forgetting/left/mean}{2,913}
\resdef{cheetah/PPO/forgetting/left/std}{965}
\resdef{cheetah/PPO/retention/Heavy-Rough/mean}{28.3}
\resdef{cheetah/PPO/retention/left/mean}{39.2}
\resdef{cheetah/PPO/retention/left/std}{17.9}
\resdef{cheetah/PPO/test/Heavy-Rough/mean}{1,302}
\resdef{cheetah/PPO/test/rel/mean}{0.9}
\resdef{cheetah/PPO/time/min/mean}{115.3}
\resdef{cheetah/SAC-vs-ENA-Combined/forgetting/left/ci/high}{\ensuremath{-}4,898}
\resdef{cheetah/SAC-vs-ENA-Combined/forgetting/left/ci/low}{\ensuremath{-}6,106}
\resdef{cheetah/SAC-vs-ENA-Combined/forgetting/left/diff}{\ensuremath{-}5,502}
\resdef{cheetah/SAC-vs-ENA-Combined/forgetting/left/pholm}{$p<0.001$}
\resdef{cheetah/SAC-vs-ENA-Combined/retention/left/ci/high}{91.6}
\resdef{cheetah/SAC-vs-ENA-Combined/retention/left/ci/low}{70.9}
\resdef{cheetah/SAC-vs-ENA-Combined/retention/left/d}{3.67}
\resdef{cheetah/SAC-vs-ENA-Combined/retention/left/diff}{+81.2}
\resdef{cheetah/SAC-vs-ENA-Combined/retention/left/pholm}{$p<0.001$}
\resdef{cheetah/SAC-vs-ENA-Combined/test/Earth-Standard/p}{$p<0.001$}
\resdef{cheetah/SAC-vs-ENA-Combined/test/Heavy-Rough/diff}{+1,116}
\resdef{cheetah/SAC-vs-ENA-Combined/test/Heavy-Rough/p}{$p<0.001$}
\resdef{cheetah/SAC-vs-ENA-Combined/test/High-Wind/p}{$p<0.001$}
\resdef{cheetah/SAC-vs-ENA-Combined/test/Ice-Floor/p}{$p<0.001$}
\resdef{cheetah/SAC-vs-ENA-Combined/test/Mars-Slippery/p}{$p<0.001$}
\resdef{cheetah/SAC/end/Earth-Standard/mean}{8,635}
\resdef{cheetah/SAC/end/Heavy-Rough/mean}{7,544}
\resdef{cheetah/SAC/end/Mars-Slippery/mean}{9,405}
\resdef{cheetah/SAC/episodes/perenv}{1,000}
\resdef{cheetah/SAC/forgetting/left/mean}{5,263}
\resdef{cheetah/SAC/forgetting/left/std}{1,234}
\resdef{cheetah/SAC/retention/Heavy-Rough/mean}{18.0}
\resdef{cheetah/SAC/retention/left/mean}{34.4}
\resdef{cheetah/SAC/retention/left/std}{13.1}
\resdef{cheetah/SAC/test/Earth-Standard/mean}{4,353}
\resdef{cheetah/SAC/test/Heavy-Rough/mean}{1,301}
\resdef{cheetah/SAC/test/High-Wind/mean}{2,642}
\resdef{cheetah/SAC/test/Ice-Floor/mean}{6,774}
\resdef{cheetah/SAC/test/Mars-Slippery/mean}{10,146}
\resdef{cheetah/SAC/test/rel/mean}{47.1}
\resdef{cheetah/SAC/time/min/mean}{281.1}
\resdef{cheetah/archived/ENA-Combined/forgetting/left/mean}{\ensuremath{-}26}
\resdef{cheetah/archived/ENA-Combined/retention/left/mean}{103.2}
\resdef{cheetah/archived/ENA-Combined/retention/left/std}{10.5}
\resdef{cheetah/archived/n}{18}
\resdef{cheetah/hardware}{Linux aarch64, 4 cores; TensorFlow threads all; ENA threads 2}
\resdef{cheetah/n}{20}
\resdef{compute/ENA-Combined-PlasticityFull/compute/candsteps/mean}{57,890,361}
\resdef{compute/ENA-Combined-PlasticityFull/test/Brown-dwarf/mean}{410.0}
\resdef{compute/ENA-Combined-PlasticityFull/test/rel/mean}{90.3}
\resdef{compute/ENA-Combined/compute/candsteps/mean}{4,617,335}
\resdef{compute/ENA-Combined/test/Brown-dwarf/mean}{61.5}
\resdef{compute/ENA-Combined/test/rel/mean}{74.1}
\resdef{compute/ENA-HoF+EnvDetection/compute/candsteps/mean}{27,816}
\resdef{compute/ENA-HoF+EnvDetection/test/rel/mean}{55.3}
\resdef{compute/rank/atleast}{15}
\resdef{compute/rank/of}{30}
\resdef{compute/ratio/full-ramp}{13}
\resdef{compute/ratio/ramp-zero}{166}
\resdef{controller/ENA-Combined-FuzzyPlasticity/test/rel/mean}{83.9}
\resdef{controller/ENA-Combined-FuzzyPlasticity/test/rel/std}{11.4}
\resdef{controller/ENA-Combined-QuadraticPlasticity/test/rel/mean}{89.3}
\resdef{controller/ENA-Combined-QuadraticPlasticity/test/rel/std}{7.7}
\resdef{controller/ENA-Combined-vs-ENA-Combined-QuadraticPlasticity/test/rel/ci/high}{12.4}
\resdef{controller/ENA-Combined-vs-ENA-Combined-QuadraticPlasticity/test/rel/ci/low}{\ensuremath{-}0.3}
\resdef{controller/ENA-Combined-vs-ENA-Combined-QuadraticPlasticity/test/rel/d}{0.35}
\resdef{controller/ENA-Combined-vs-ENA-Combined-QuadraticPlasticity/test/rel/nreq}{79}
\resdef{controller/mde}{0.59}
\resdef{controller/n}{30}
\resdef{controller/verdict}{linear ramp}
\resdef{detbench/bestother/excitation}{exponential sweep}
\resdef{detbench/bestother/features}{SWT Haar}
\resdef{detbench/bestother/passes}{4}
\resdef{detbench/cartpole/passes}{26}
\resdef{detbench/cartpole/seeds}{30}
\resdef{detbench/chosen/falseseeds}{1}
\resdef{detbench/chosen/lower}{82.8}
\resdef{detbench/chosen/minpurity}{75}
\resdef{detbench/chosen/passes}{29}
\resdef{detbench/chosen/puritymean}{99.2}
\resdef{detbench/chosen/upper}{99.9}
\resdef{detbench/chosenexcitation/minpasses}{29}
\resdef{detbench/cmp/chirp/p}{$p<0.001$}
\resdef{detbench/cmp/other/p}{$p<0.001$}
\resdef{detbench/passing}{3}
\resdef{detbench/seeds}{30}
\resdef{detbench/transition/changes}{26.3}
\resdef{detbench/transition/inside}{24.5}
\resdef{detbench/transition/niches}{21.8}
\resdef{detbench/transition/passes}{0}
\resdef{detbench/transition/purity}{21.9}
\resdef{detbench/transitionfixed/changes}{16.2}
\resdef{detbench/transitionfixed/merged}{14}
\resdef{detbench/transitionfixed/niches}{6.5}
\resdef{sacd/0.5/alphamax}{3.7}
\resdef{sacd/0.5/valuemax}{121}
\resdef{sacd/0.89/alphamax}{4,907}
\resdef{sacd/0.89/valuemax}{40,174}
\resdef{sacd/0.98/alphamax}{21,048}
\resdef{sacd/0.98/alphastart}{1.0}
\resdef{sacd/0.98/seeds}{3}
\resdef{sacd/0.98/steps}{148k}
\resdef{sacd/0.98/valuemax}{125,066}
\resdef{seeds/ablation/first}{1000}
\resdef{seeds/ablation/last}{1029}
\resdef{seeds/challenge/first}{1000}
\resdef{seeds/challenge/last}{1049}
\resdef{seeds/cheetah/first}{1000}
\resdef{seeds/cheetah/last}{1019}
\resdef{selection/ENA-Combined-BestTrust/played/mean}{1.0}
\resdef{selection/ENA-Combined-BestTrust/test/Jupiter-Heavy/mean}{206.8}
\resdef{selection/ENA-Combined-BestTrust/test/rel/mean}{53.8}
\resdef{selection/ENA-Combined-BestTrust/test/rel/std}{32.3}
\resdef{selection/ENA-Combined-NicheMatch-vs-ENA-Combined-BestTrust/test/Jupiter-Heavy/ci/high}{\ensuremath{-}131.6}
\resdef{selection/ENA-Combined-NicheMatch-vs-ENA-Combined-BestTrust/test/Jupiter-Heavy/ci/low}{\ensuremath{-}269.6}
\resdef{selection/ENA-Combined-NicheMatch-vs-ENA-Combined-BestTrust/test/Jupiter-Heavy/diff}{\ensuremath{-}200.6}
\resdef{selection/ENA-Combined-NicheMatch-vs-ENA-Combined-BestTrust/test/Jupiter-Heavy/p}{$p<0.001$}
\resdef{selection/ENA-Combined-NicheMatch-vs-ENA-Combined-BestTrust/test/rel/ci/high}{2.5}
\resdef{selection/ENA-Combined-NicheMatch-vs-ENA-Combined-BestTrust/test/rel/ci/low}{\ensuremath{-}13.3}
\resdef{selection/ENA-Combined-NicheMatch-vs-ENA-Combined-BestTrust/test/rel/d}{\ensuremath{-}0.26}
\resdef{selection/ENA-Combined-NicheMatch-vs-ENA-Combined-BestTrust/test/rel/diff}{\ensuremath{-}5.4}
\resdef{selection/ENA-Combined-NicheMatch-vs-ENA-Combined-BestTrust/test/rel/higher}{15}
\resdef{selection/ENA-Combined-NicheMatch-vs-ENA-Combined-BestTrust/test/rel/lower}{15}
\resdef{selection/ENA-Combined-NicheMatch-vs-ENA-Combined-BestTrust/test/rel/pholm}{$p=0.170$}
\resdef{selection/ENA-Combined-NicheMatch/played/mean}{3.2}
\resdef{selection/ENA-Combined-NicheMatch/test/Jupiter-Heavy/mean}{407.4}
\resdef{selection/ENA-Combined-NicheMatch/test/rel/mean}{59.2}
\resdef{selection/ENA-Combined-NicheMatch/test/rel/std}{25.9}
\resdef{selection/ENA-Combined-vs-ENA-Combined-NicheMatch/test/rel/ci/high}{\ensuremath{-}15.1}
\resdef{selection/ENA-Combined-vs-ENA-Combined-NicheMatch/test/rel/ci/low}{\ensuremath{-}32.8}
\resdef{selection/ENA-Combined-vs-ENA-Combined-NicheMatch/test/rel/d}{\ensuremath{-}1.01}
\resdef{selection/ENA-Combined-vs-ENA-Combined-NicheMatch/test/rel/diff}{\ensuremath{-}24.0}
\resdef{selection/ENA-Combined-vs-ENA-Combined-NicheMatch/test/rel/lower}{29}
\resdef{selection/ENA-Combined-vs-ENA-Combined-NicheMatch/test/rel/pairs}{30}
\resdef{selection/ENA-Combined-vs-ENA-Combined-NicheMatch/test/rel/pholm}{$p<0.001$}
\resdef{selection/ENA-Combined/played/mean}{7.2}
\resdef{selection/ENA-Combined/test/rel/mean}{83.2}
\resdef{selection/ENA-Combined/test/rel/std}{16.1}
\resdef{selection/ENA-Plasticity-vs-ENA-Combined-BestTrust/test/rel/ci/high}{3.2}
\resdef{selection/ENA-Plasticity-vs-ENA-Combined-BestTrust/test/rel/ci/low}{\ensuremath{-}22.2}
\resdef{selection/ENA-Plasticity-vs-ENA-Combined-BestTrust/test/rel/diff}{\ensuremath{-}9.5}
\resdef{selection/ENA-Plasticity-vs-ENA-Combined-BestTrust/test/rel/p}{$p=0.137$}
\resdef{selection/ENA-Plasticity-vs-ENA-Combined-NicheMatch/test/rel/ci/high}{6.5}
\resdef{selection/ENA-Plasticity-vs-ENA-Combined-NicheMatch/test/rel/ci/low}{\ensuremath{-}14.6}
\resdef{selection/ENA-Plasticity-vs-ENA-Combined-NicheMatch/test/rel/diff}{\ensuremath{-}4.1}
\resdef{selection/ENA-Plasticity-vs-ENA-Combined-NicheMatch/test/rel/p}{$p=0.437$}
\resdef{selection/ENA-Plasticity/test/rel/mean}{63.3}
\resdef{selection/audit/bpr}{90.2}
\resdef{selection/audit/bprhigher}{23}
\resdef{selection/audit/nm}{64.0}
\resdef{selection/audit/nmhigher}{2}
\resdef{selection/audit/of}{90}
\resdef{selection/audit/phases}{37}
\resdef{selection/identity/checked}{60}
\resdef{selection/identity/identical}{60}
\resdef{selection/mde}{0.59}
\resdef{selection/n}{30}
\resdef{stats/alpha}{0.05}
\resdef{stats/ci}{95}
\resdef{stats/power}{80}
\resdef{stats/welch/agree}{20}
\resdef{stats/welch/planned}{20}
 
\begin{document}

\title{Mitigating Catastrophic Forgetting with Physics-Indexed Specialist Memory: the Evolutionary Neuroplastic Agent}

\author{
\IEEEauthorblockN{Blas Eugenio Vicco Zarate}
\IEEEauthorblockA{\textit{Facultad de Ingenier\'{\i}a y Ciencias H\'{\i}dricas}\\\textit{Universidad Nacional del Litoral}\\Santa Fe, Argentina \\}
\IEEEauthorblockA{\texttt{blasvicco@gmail.com}}
\and
\IEEEauthorblockN{Miguel Hern\'an Vicco}
\IEEEauthorblockA{\textit{Facultad de Ciencias M\'edicas}\\\textit{Universidad Nacional del Litoral}\\Santa Fe, Argentina \\}
\IEEEauthorblockA{\texttt{mvicco@santafe-conicet.gov.ar}}
}

\maketitle

\begin{abstract}
Continual reinforcement learning requires adaptation to changing conditions while preserving performance under previously encountered dynamics. It was investigated whether specialists generated by evolutionary search can provide a reusable memory when combined with environment detection and failure-driven search. The Evolutionary Neuroplastic Agent (ENA) evolves small neural networks without gradients, archives selected policies by detected dynamic regime, and retrieves them through Bayesian policy reuse. The architecture was evaluated in non-stationary CartPole and HalfCheetah environments. In the \Res{ablation/n}-replicate CartPole ablation, the complete architecture achieved a mean success rate of \Res{ablation/ENA-Combined/test/rel/mean}\% across the five test environments. Among variants that learned the harder CartPole environments, the complete architecture showed the most consistent retention. A selection ablation also showed that the retrieval rule affected performance. In the evaluated sequences, ENA adapted more slowly than PPO on CartPole and achieved lower returns than SAC on HalfCheetah. The evidence concerns the complete architecture and does not isolate gradient-free optimization as the cause of retention or establish superiority over other continual-learning methods.
 \end{abstract}


\section{Introduction}

Reinforcement learning (RL) agents are usually trained and deployed on the assumption that the environment does not change. When it does, for instance when a robot's physics shift between conditions, an agent with a single policy adapts by overwriting the parameters that encoded its previous behavior. This is \textit{catastrophic forgetting}~\cite{mccloskey1989catastrophic}. When the original conditions return, the agent has to learn them again.

Continual reinforcement learning studies agents that keep adapting without losing what they have already learned~\cite{khetarpal2022towards}. One family of methods protects or partitions the weights of a shared network, by regularization~\cite{kirkpatrick2017overcoming}, replay~\cite{rolnick2019experience} or task-specific heads~\cite{kessler2022same}. Another keeps a separate policy for each detected context and recalls it when that context returns~\cite{dasilva2006dealing,alegre2021minimum,dick2024statistical}. Neuroevolution offers a third route. It trains neural networks by mutation and selection over a population of candidates, instead of by gradient descent. Nisioti et al.~\cite{nisioti2026continual} showed that evolution strategies balance adapting to new tasks and remembering old ones well, while genetic algorithms learn new tasks best but forget more. In these methods, however, the whole population evolves toward the current task. The population receives no signal that the environment has changed, nor indication of which earlier solution to restore when an earlier environment returns.

Evolutionary methods learn by trying many candidate solutions at once. Along the way, they produce many solutions that already work in the conditions they were tried in. The question addressed in this paper is whether a small set of skills can turn those solutions into a memory that mitigates forgetting. The skills are inspired by how organisms vary under stress and consolidate memories. The \textbf{main research question} is:
\begin{itemize}[\IEEEsetlabelwidth{\textbf{RQ1a}}]
\item[\textbf{RQ1}] When an agent encounters a sequence of non-stationary environments without explicit change notifications, does preserving effective policies indexed by detected physical dynamics---combined with failure-driven modulation of search effort---mitigate catastrophic forgetting?
\end{itemize}
Three narrower questions examine this memory mechanism in detail:
\begin{itemize}[\IEEEsetlabelwidth{\textbf{RQ1a}}]
\item[\textbf{RQ1a}] Is preserving successful policies sufficient on its own, or must the archive be explicitly indexed by the physical regime under which each policy was validated?
\item[\textbf{RQ1b}] Does policy archiving require additional environment interaction, or does it operate as an emergent by-product of population evaluations?
\item[\textbf{RQ1c}] Can dynamic shifts and previously encountered regimes be autonomously detected and identified purely from unassisted agent experience?
\end{itemize}
Searching incurs computational cost, as candidate policies require evaluation in the environment. The \textbf{secondary research question} is therefore:
\begin{itemize}[\IEEEsetlabelwidth{\textbf{RQ1a}}]
\item[\textbf{RQ2}] Does dynamically concentrating search effort during performance failure achieve competitive retention and reliability while significantly reducing the computational cost incurred by uniform full-effort search?
\end{itemize}

To address these questions, the \textbf{Evolutionary Neuroplastic Agent (ENA)} is introduced. ENA evolves a population of small neural networks, termed \textit{specialists}. Each specialist is scored with a \textit{Trust} metric tracking recent return history. Search effort, termed \textit{Plasticity}, scales dynamically with the performance deficit of the active specialist. Validated specialists are archived in a \textit{Hall of Fame}. \textit{Environment detection} identifies dynamic transitions and recognizes previously encountered physical regimes strictly from unassisted agent observations. On CartPole, it evaluates predictive accuracy under transition models of each known regime. On HalfCheetah, it applies a structured diagnostic perturbation---a \textit{chirp}---and analyzes mechanical response features. \textit{Specialist selection} governs which archived specialist to deploy via Bayesian policy reuse~\cite{rosman2016bayesian}, integrating detector evidence with observed episodic returns. As a structural design constraint, no component of ENA computes gradients.

Two domains are studied. In CartPole~\cite{barto1983neuronlike}, the agent balances a pole on a moving cart. In HalfCheetah, a two-legged robot simulated in MuJoCo~\cite{todorov2012mujoco} learns to run. In each domain, the agent trains on three environments in turn, which differ in gravity and other physical parameters (Section~\ref{sec:setup}). The evidence for the claim is a \textit{skill ablation}: the same agent, with each skill switched off in turn. Alongside it, standard single-policy learners were trained on the same sequences of environments, with the same number of environment steps: DQN~\cite{mnih2015human}, PPO~\cite{schulman2017proximal} and SAC~\cite{haarnoja2018soft}. They show how much standard learners forget on these streams. They are a reference, not opponents. \textit{It is not claimed that ENA outperforms other continual-RL methods}, nor that gradient-free learning is what prevents forgetting. The question is only whether these skills mitigate it. Comparing ENA with continual-RL systems built for the same problem is left to future work (Section~\ref{sec:scope}).

The contributions, each tied to the question it answers, are:
\begin{itemize}
    \item \textbf{A skill ablation} on CartPole, which switches each skill off in turn (RQ1, RQ1a, RQ1b, RQ2).
    \item \textbf{A selection ablation}, which compares Bayesian policy reuse with simpler ways of choosing an archived specialist (RQ1a).
    \item \textbf{A measurement of forgetting} in standard single-policy learners, on CartPole and HalfCheetah, as the reference for RQ1.
    \item \textbf{Gradient-free environment detection} for both domains, including the chirp for HalfCheetah, evaluated on a detection benchmark (RQ1c).
    \item \textbf{A study of what failure-driven search costs}: one run at full effort in every episode, and two alternative rules for turning failure into search effort (RQ2).
    \item \textbf{Reproducibility}: every result and setting reported here is generated by scripts, from the released code and experiment outputs.
\end{itemize}

Section~\ref{sec:related} reviews related work. Section~\ref{sec:method} states the problem and describes ENA. Section~\ref{sec:setup} presents the environments, protocols and baselines, and Section~\ref{sec:results} the results. Section~\ref{sec:discussion} answers the research questions and discusses the limitations. Section~\ref{sec:conclusion} concludes.
\section{Related Work}
\label{sec:related}

\subsection{Non-Stationary and Continual Reinforcement Learning}
A neural network trained by gradient descent on one task after another degrades on the earlier tasks. McCloskey and Cohen named this failure catastrophic interference~\cite{mccloskey1989catastrophic}. Continual reinforcement learning studies agents that must keep learning in such conditions without losing what they already know~\cite{khetarpal2022towards}. Elastic Weight Consolidation~\cite{kirkpatrick2017overcoming} slows learning on the weights that mattered for earlier tasks. CLEAR~\cite{rolnick2019experience} replays past experience with behavioral cloning, and needs no task boundaries. When tasks are incompatible, a single network cannot serve them all. For such tasks, OWL~\cite{kessler2022same} gives each task its own policy head on a shared feature extractor, and chooses a head at test time with a multi-armed bandit. These three methods protect or partition the weights of one gradient-trained network. ENA, like the context-based methods below, preserves whole policies instead. An archived specialist is never trained again, so there is nothing to protect.

\subsection{Evolutionary Reinforcement Learning and Continual Neuroevolution}
Neuroevolution optimizes neural networks by mutation and selection over a population, instead of by gradients~\cite{stanley2019designing}. Evolution strategies~\cite{salimans2017evolution} and genetic algorithms~\cite{such2017deep} are competitive with gradient-based RL on standard benchmarks. They need many samples, however~\cite{khadka2018evolutionary,pourchot2019cemrl}, and optimizing many parameters remains hard for them~\cite{khadka2018evolutionary}. Hybrid methods use a population to diversify the experience of a gradient-trained learner~\cite{khadka2018evolutionary}. Nisioti et al.~\cite{nisioti2026continual} compared evolution strategies and genetic algorithms with PPO, continual RL methods and population-based RL, under changes of task and environment. Evolution strategies achieved the best balance between adapting and remembering. Genetic algorithms were the most plastic, but forgot more. The authors related this balance to how much the neighborhoods of good solutions in weight space overlap between consecutive tasks.

Earlier work on evolutionary algorithms for changing optimization problems developed remedies such as increasing diversity after a change, maintaining diversity throughout the run, and memory~\cite{yazdani2021survey}. ENA uses these three (Section~\ref{sec:inspiration}). In the methods Nisioti et al. study, as in standard neuroevolution, the population evolves as a whole toward the current task. Changes are not detected, and earlier solutions are not stored and recalled. ENA adds both without gradients: it identifies the dynamics online, and it preserves specialists indexed by their physics.

\subsection{Quality-Diversity and Repertoire-Based Adaptation}
Quality-Diversity algorithms such as MAP-Elites~\cite{mouret2015illuminating} search for many diverse, high-performing solutions instead of a single optimum, and store them in an archive. Intelligent Trial and Error (ITE)~\cite{cully2015robots} computes such an archive offline, in simulation. With it, a damaged robot finds a working gait within minutes, by trying archived behaviors guided by a Bayesian model. TeLAPA~\cite{lillo2026beyond} brings the same idea to continual RL. For each task, it trains a base policy with PPO and fills an archive of behaviorally diverse policies around it with MAP-Elites. When a new or revisited task begins, short adaptation probes pick a starting point from all archives, so that competent alternatives remain available when a task is revisited. TeLAPA is evaluated on MiniGrid and on cyclic friction shifts in MuJoCo, HalfCheetah included, with the task boundaries given. ENA's Hall of Fame is also an archive of specialists, but it differs in two ways. It is built online, while the environment changes. It is organized by the physics the agent detects, not by a behavior descriptor.

\subsection{Change Detection and Policy Reuse}
Sequential change detection asks when a data stream has changed. Classic methods range from Page's CUSUM~\cite{page1954continuous} to Bayesian online changepoint detection~\cite{adams2007bayesian}. On CartPole, ENA uses the Predictive-Mixture CuSum~\cite{halme2026sequential}. It detects a change away from a known distribution without a model of the distribution after the change, by mixing predictors fitted on recent windows of data.

In RL, RL-CD~\cite{dasilva2006dealing} learned one model of the environment per \textit{context}, and recognized contexts by which model predicts best. ENA follows this idea. MBCD~\cite{alegre2021minimum} combined an ensemble of learned probabilistic dynamics models with a high-confidence change-point detector whose false-alarm rate is bounded. It created a new policy for each new context, and reused an earlier one when a context recurred. SWOKS~\cite{dick2024statistical} detects contexts with optimal-transport distances and Kolmogorov--Smirnov tests, and trains one policy per context. These systems are the closest to ENA in purpose, and they too preserve a dedicated policy per context. RL-CD is tabular, while MBCD trains its models and policies by gradient descent, and SWOKS its policies.

ENA differs from these systems in how memory is populated and recalled, rather than in representation. Its specialists are individuals generated inherently by population search, rendering policy archiving free of dedicated training overhead. The search that produces them is scaled to failure. Recall weighs the detector's evidence together with the observed returns. Section~\ref{sec:scope} explains why these systems are not baselines here. To choose which stored policy to deploy, Bayesian Policy Reuse~\cite{rosman2016bayesian} keeps a belief over task types. The belief is updated from a signal of each episode, such as its return or its transitions. ENA adapts it to an online, non-stationary setting, and updates the belief from the returns and the detector's likelihoods together (Section~\ref{sec:selection}).

\subsection{Positioning}
Table~\ref{tab:capabilities} summarizes how related methods preserve and recall earlier behavior. OWL and TeLAPA are told the task boundaries during training, so they detect nothing online. ITE starts adapting when it senses a performance drop. TeLAPA is partly gradient-free: it trains with PPO, and fills its archives by mutation. ENA does not introduce any of these capabilities on its own. Change detection, recurrence recognition, whole-policy preservation, gradient-free learning and probing all have precedents. What ENA combines is threefold. Its archive is populated with candidate solutions generated naturally during evolutionary search (RQ1b). The archive is indexed online by physics the agent detects without labels (RQ1c). The search that feeds it scales its effort to failure (RQ2). No component computes a gradient. Unlike most methods in the table, ENA also assumes that it can reset the environment: to evaluate candidates and, on HalfCheetah, to play its chirp.

\begin{table*}[t]
\caption{How Related Methods Keep and Recall Earlier Behavior}
\begin{center}
\footnotesize
\begin{tabular}{|l|l|c|c|c|c|}
\hline
\textbf{Method} & \textbf{Memory} & \textbf{Online detection} & \textbf{Recurrence} & \textbf{Gradient-free} & \textbf{Probing} \\
\hline
EWC~\cite{kirkpatrick2017overcoming} & regularized shared weights & -- & -- & -- & -- \\
CLEAR~\cite{rolnick2019experience} & replay into shared weights & -- & -- & -- & -- \\
OWL~\cite{kessler2022same} & shared extractor (EWC), one head per task & -- & bandit at test & -- & -- \\
RL-CD~\cite{dasilva2006dealing} & model and policy per context & \checkmark & \checkmark & tabular & -- \\
MBCD~\cite{alegre2021minimum} & model and policy per context & \checkmark & \checkmark & -- & -- \\
SWOKS~\cite{dick2024statistical} & policy per context & \checkmark & \checkmark & -- & -- \\
TeLAPA~\cite{lillo2026beyond} & per-task archives of initializations & -- & by label & partly & few-shot probes \\
Nisioti et al.~\cite{nisioti2026continual} & population only & -- & -- & \checkmark & -- \\
ITE~\cite{cully2015robots} & offline behavior archive & performance drop & -- & \checkmark & trials \\
\textbf{ENA} & \textbf{frozen specialists per niche} & \checkmark & \checkmark & \checkmark & \textbf{chirp (HalfCheetah)} \\
\hline
\end{tabular}
\label{tab:capabilities}
\end{center}
\end{table*}
\section{Methodology}
\label{sec:method}

\subsection{Problem Setting}
\label{sec:problem}
The agent meets a sequence of environments $\mathcal{M}_1, \dots, \mathcal{M}_E$. Each environment is an episodic Markov decision process $\mathcal{M}_e = (\mathcal{S}, \mathcal{A}, P_e, r, \mu_0)$, with these components:
\begin{itemize}
\item the state space $\mathcal{S}$: the vectors the agent observes, whose components Appendix~\ref{app:information} lists;
\item the action space $\mathcal{A}$: on CartPole, a push of the cart to the left or to the right; on HalfCheetah, a vector of joint torques within fixed bounds;
\item the transition kernel $P_e(s' \mid s, a)$: the distribution of the next state $s'$ after action $a$ in state $s$, which encodes the physics;
\item the reward function $r(s, a, s')$: a real number for each transition, given for each domain in Section~\ref{sec:setup};
\item the initial-state distribution $\mu_0$.
\end{itemize}
All the environments share $\mathcal{S}$, $\mathcal{A}$, $r$ and $\mu_0$. They differ only in their transition kernels $P_e$, that is, in their physics.

An episode starts in a state $s_0 \sim \mu_0$. At each step $t$, the agent plays an action $a_t \in \mathcal{A}$. The environment moves to $s_{t+1} \sim P_e(\cdot \mid s_t, a_t)$, and the agent receives the reward $r_t = r(s_t, a_t, s_{t+1})$. The episode ends when a terminal state is reached, such as a fallen pole, or when it is truncated at the time limit of $t_{\max}$ steps. The \textit{return} of an episode is the undiscounted sum of its rewards,
\begin{equation}
R = \sum_{t} r_t,
\label{eq:return}
\end{equation}
where $t$ runs over the steps of the episode. A \textit{policy} $\pi : \mathcal{S} \to \mathcal{A}$ maps each state to an action. Every policy in ENA is deterministic. ENA scores its policies by their returns~\eqref{eq:return}. The single-policy baselines of Section~\ref{sec:setup} train on discounted returns, with the discount factors of Appendix~\ref{app:hyperparameters}. Every agent is evaluated on $R$.

During training, the environments are presented one after the other, each for a fixed budget. In the skill ablation, the budget is a number of episodes; when ENA is compared with other learners, it is a number of environment steps (Section~\ref{sec:setup}). The agent observes only states, actions, rewards and episode ends. It is never told when the environment changes, which environment it is in, or how many there are (Appendix~\ref{app:information}), and it keeps all of its state across the changes. After training, the agent is tested, frozen, on a sequence that contains every training environment and some it has never seen. Again, it is not told where one environment ends and the next begins.

An agent \textit{forgets} environment $\mathcal{M}_e$ to the extent that its test return there falls below its return at the end of $\mathcal{M}_e$'s training phase. Section~\ref{sec:metrics} measures this as retention and absolute forgetting. An agent \textit{adapts} when it learns each new environment within its budget. The goal is to adapt to every new environment without forgetting the earlier ones. This setting is narrower than continual RL in general~\cite{khetarpal2022towards}: the observation space, the action space and the reward stay the same, and only the physics change.

\subsection{Overview and Terms}
The Evolutionary Neuroplastic Agent (ENA) never computes a gradient. It maintains a population of $N$ small neural-network policies, the \textit{specialists}, and each specialist carries a scalar \textit{Trust} score. In every training episode, one specialist acts, and its return updates its Trust. A \textit{Plasticity} value then decides how much of the population is re-evaluated. The acting specialist may be archived in a \textit{Hall of Fame}, and the population evolves. Two more skills make the archive usable when the environment changes. \textit{Environment detection} notices when the physics change and recognizes physics seen before. \textit{Specialist selection} chooses which archived specialist to deploy. Each skill is an interchangeable module with an ``off'' setting, which makes the ablation of Section~\ref{sec:results} possible. Section~\ref{sec:procedure} gives the training and test episodes as pseudocode, once every quantity in them has been defined.

The biological names are labels, not claims. In operational terms:
\begin{itemize}
\item a \textit{specialist} is one policy network of the population;
\item a \textit{gladiator} is a frozen copy of a specialist, stored in the Hall of Fame;
\item a \textit{niche} is a set of physics that the detector treats as one environment, numbered by ENA itself and never given by the experiment;
\item a \textit{regime} is the stretch of training between two changes of physics confirmed by environment detection, or before the first one;
\item \textit{Trust} is a running score of a specialist's recent normalized returns, used to rank specialists;
\item \textit{Plasticity} is a number in $[0,1]$ that sets how much search effort an episode receives.
\end{itemize}

\subsection{Inspiration}
\label{sec:inspiration}
ENA borrows two strategies from nature. Each one works through a mechanism that evolutionary algorithms for changing environments already use.

\subsubsection{Variation in proportion to stress (Plasticity)}
Bacteria under stress increase their mutation rate. The response is regulated and transient. Variation rises when the current genotype is failing and stays low otherwise, because most mutations in a well-adapted organism are harmful or neutral~\cite{galhardo2007mutation}. Evolutionary computation adopted the idea as \textit{triggered hypermutation}, which raises the mutation rate when performance drops in a non-stationary environment~\cite{cobb1990investigation}. More generally, it studies the control of an algorithm's parameters during the run~\cite{karafotias2015parameter}.

ENA's Plasticity extends this response from the mutation rate to the whole search effort. A failure signal, defined in Section~\ref{sec:trust}, sets three things: how many specialists are re-evaluated (a random subset), how long each is evaluated, and how far mutants move from their parents. The share of the population replaced in each generation stays constant. What scales with failure is the evidence gathered before selection, and the size of the steps taken after it. The rationale is cost. Every candidate step is an environment step, so a converged agent should spend few of them, while an agent that has just met new physics needs to search more broadly. Freshly initialized specialists also enter the population in every generation, as \textit{random immigrants}~\cite{tinos2007self} do. Their purpose is to keep the population diverse enough to follow a moving optimum.

\subsubsection{Consolidated memory (Hall of Fame)}
The complementary learning systems theory~\cite{mcclelland1995there,kumaran2016learning} holds that the brain learns quickly in the hippocampus and consolidates slowly into the neocortex. There, knowledge is protected from interference by new learning. The theory was proposed in part to explain why the brain does not suffer the catastrophic interference of gradient-trained networks~\cite{mccloskey1989catastrophic}. ENA follows the same division of labor. The population is the fast learner, overwritten in every generation. The Hall of Fame is the consolidated store. It is entered only through a gate, it is never modified, and evolution never replaces it.

In evolutionary computation, a \textit{hall of fame} first kept the best individuals of past generations in competitive coevolution~\cite{rosin1997new}. Memory-enhanced evolutionary algorithms store good solutions and reinsert them when earlier conditions return~\cite{branke1999memory}. ENA adds two things to this kind of memory. The memory is indexed by the physics in which each specialist succeeded (Section~\ref{sec:detection}). Recall is a Bayesian decision rather than a reinsertion (Section~\ref{sec:selection}).

\subsection{Population and Specialists}
Each specialist $i$ is a policy $\pi_i$, computed by a small multi-layer perceptron. On CartPole, its hidden layers have \Res{cfg/ena/brain} units, with orthogonal initialization. Its output layer is linear, and the action with the highest output is played. On HalfCheetah, the hidden layers have \Res{cfg/cheetahena/brain} units. Tanh hidden layers follow the evolution-strategies policy of Salimans et al.~\cite{salimans2017evolution}, and normc initialization follows its released code~\cite{openai2017esstarter}. The output is also a tanh there, as in the actor networks of CEM-RL~\cite{pourchot2019cemrl}, which keeps actions in their range. A specialist's inputs are normalized with a running mean and variance. These statistics keep drifting while the agent explores. Before a specialist acts, its first layer is re-scaled analytically to the current statistics, so that its behavior is preserved without retraining. During training, the acting specialist is chosen $\epsilon$-greedily, with $\epsilon = \Res{cfg/ena/epsilon}$. With probability $\epsilon$, it is a random non-archived specialist; otherwise, it is the specialist with the highest Trust $T_i$ (Section~\ref{sec:trust}).

\subsection{Trust and Plasticity}
\label{sec:trust}
Trust and Plasticity both derive from the same \textit{failure signal}, which grades a return against the agent's recent returns. Let $H$ be the history of the last \Res{cfg/ena/history} returns of the acting specialists, and $H_x$ its $x$-th percentile. Let $R_{\min}$ be the worst return of the current regime. Both $H$ and $R_{\min}$ are cleared when a new regime starts, that is, when a change of physics is confirmed (Section~\ref{sec:detection}). For a return $R$~\eqref{eq:return}, the failure signal is
\begin{equation}
f(R) =
\begin{cases}
1 & |H| < \Res{cfg/ena/signal/warmup}, \\
f_{\min} & R \ge H_{\Res{cfg/ena/signal/percentile}}, \\
1 - (1 - f_{\min})\,\rho(R) & \text{otherwise,}
\end{cases}
\label{eq:ramp}
\end{equation}
\begin{equation}
\rho(R) = \mathrm{clip}\!\left(\frac{R - R_{\min}}{H_{\Res{cfg/ena/signal/percentile}} - R_{\min}},\, 0,\, 1\right),
\label{eq:rho}
\end{equation}
where $f_{\min} = \Res{cfg/ena/signal/min}$ is the floor of the signal. During the first \Res{cfg/ena/signal/warmup} returns of a regime, $H$ is too short for reliable percentiles, and the signal is $1$. A return at or above the \Res{cfg/ena/signal/percentile}th percentile gives the floor, so that a confident agent still searches a little. Below that percentile, the signal grows linearly, up to $1$ at the worst return of the regime. Its slope, $1 - f_{\min}$, is therefore fixed by the floor, not chosen separately. The function $f$ is termed the \textit{linear ramp}.

\subsubsection{Trust}
Trust is built on a second scale, the \textit{normalized return} $\tilde{R}$. It places a return within the spread of recent returns, between the \Res{cfg/ena/trust/plow}th and \Res{cfg/ena/trust/phigh}th percentiles of $H$:
\begin{equation}
\tilde{R}(R) = \mathrm{clip}\!\left(\frac{R - H_{\Res{cfg/ena/trust/plow}}}{H_{\Res{cfg/ena/trust/phigh}} - H_{\Res{cfg/ena/trust/plow}}},\, \Res{cfg/ena/trust/low},\, \Res{cfg/ena/trust/high}\right).
\label{eq:normreturn}
\end{equation}
A small constant keeps the denominator positive. When $H_{\Res{cfg/ena/trust/phigh}} = H_{\Res{cfg/ena/trust/plow}}$, as when every recent CartPole return is the maximum, $\tilde{R}$ is therefore $0$ at that return. Below it, $\tilde{R}$ takes the lower clip bound, and above it the upper one. The clip is deliberately wider than $[0,1]$, so that a specialist that beats the current record still ranks above the rest. During the warm-up, while $|H| < \Res{cfg/ena/signal/warmup}$, there are no reliable percentiles. $\tilde{R}$ is then the return relative to the solved return $R_{\text{solved}}$, the task's registered success threshold (Appendix~\ref{app:information}), with the same clip.

The acting specialist's Trust $T_i$ is an exponential average of its normalized returns, whose rate is the failure signal $D = f(R)$:
\begin{equation}
T_i \leftarrow (1 - D)\,T_i + D\,\tilde{R}(R).
\label{eq:trust}
\end{equation}
A poor episode moves Trust quickly, and a good one only nudges it. Most episodes also re-open the competition. The \textit{reference Trust} $T_{\mathrm{ref}} = \tilde{R}(\bar{H})$ is the normalized mean $\bar{H}$ of the history. Until $H$ holds a few returns, $T_{\mathrm{ref}}$ is a fixed placeholder. Whenever $D > \Res{cfg/ena/trust/reset}$, that is, while $|H| < \Res{cfg/ena/signal/warmup}$ or whenever the return falls below $H_{\Res{cfg/ena/signal/percentile}}$, the Trust $T$ of every specialist except the gladiators is blended toward it:
\begin{equation}
T \leftarrow (1 - D)\,T + D\,T_{\mathrm{ref}}.
\label{eq:blend}
\end{equation}

\subsubsection{Plasticity}
Plasticity $\mathcal{P} \in [0, 1]$ sets the search effort of an episode, and a \textit{controller} sets it from the episode's return (Section~\ref{sec:controllers}). After each training episode, Plasticity chooses $k$ of the other $N-1$ specialists at random as \textit{candidates}, and evaluates each one for at most $h$ steps:
\begin{equation}
\begin{split}
k &= \min\left(N-1,\, \max\left(k_{\min}, \lfloor N\mathcal{P} \rfloor\right)\right), \\
h &= \left\lfloor h_{\min} + (h_{\max} - h_{\min})\,\mathcal{P} \right\rfloor.
\end{split}
\label{eq:plasticity}
\end{equation}
Here $k_{\min} = \Res{cfg/ena/search/mincandidates}$ and $h_{\min} = \Res{cfg/ena/smin}$. $h_{\max}$ is the time limit $t_{\max}$ on CartPole (\Res{cfg/ena/smax} steps), and \Res{cfg/cheetah/maxeval} steps on HalfCheetah. Candidates play in a copy of the current environment. A candidate still playing after $h$ steps has its return rescaled to a full-episode equivalent: the return is multiplied by $t_{\max}/h$ (Appendix~\ref{app:information}). A candidate whose episode ends earlier keeps its return. Each candidate's Trust is then updated with~\eqref{eq:trust}, at the rate $f$ of its own return. Plasticity also scales the mutation noise (Section~\ref{sec:search}).

On CartPole, the return counts the steps survived. A candidate that survives all $h$ steps is therefore credited with \Res{cfg/cartpole/maxsteps}, the maximum return. At $h_{\min}$, surviving \Res{cfg/ena/smin} steps is enough. On HalfCheetah, every candidate is rescaled, from at most \Res{cfg/cheetah/maxeval} of \Res{cfg/cheetah/maxsteps} steps. Whether these short returns rank specialists as full episodes would was not checked. There, the gladiators deployed at test are those confirmed on full episodes, in the Olympiads (Section~\ref{sec:hof}).

In short, under the linear ramp, a confident agent evaluates \Res{cfg/ena/search/rampmincandidates} candidates for at most \Res{cfg/ena/search/rampminsteps} steps each (\Res{cfg/cheetah/rampminsteps} on HalfCheetah), and mutates gently. A failing agent searches broadly and deeply. The minimum search, $k_{\min}$ candidates for $h_{\min}$ steps, is the zero controller's (Section~\ref{sec:controllers}). Every candidate step is a real environment step. When ENA is compared with other learners, these steps count against ENA's budget. In the skill ablation, whose budget is a number of episodes, they are extra interaction.

\subsubsection{Plasticity controllers}
\label{sec:controllers}
A Plasticity controller maps the return of an episode to $\mathcal{P}$. Trust always uses the linear ramp~\eqref{eq:ramp} as its rate, so only the Plasticity controller changes. Five controllers are defined:
\begin{itemize}
\item \textit{Linear ramp}, ENA's controller: $\mathcal{P} = f(R)$~\eqref{eq:ramp}.
\item \textit{Zero}: $\mathcal{P} = 0$ in every episode. This is the minimum search: $k=k_{\min}$ candidates for $h_{\min}$ steps each, and the lowest mutation noise. The ablation uses it to switch Plasticity off.
\item \textit{Full}: $\mathcal{P} = 1$ in every episode. Every other specialist is evaluated for $h_{\max}$ steps, and mutants receive the largest noise. It is the reference for a fixed maximum effort (Section~\ref{sec:compute}).
\item \textit{Quadratic}:
\[
\mathcal{P} = \mathrm{clip}\left(\left(\frac{R_{\max} - R}{R_{\max} - R_{\min}}\right)^{2},\, f_{\min},\, 1\right),
\]
where $R_{\max}$ is the best return of the current regime. A small constant keeps the denominator positive, so $\mathcal{P} = f_{\min}$ when $R_{\max} = R_{\min}$. It stays low for moderate failures, and rises steeply only near the worst return.
\item \textit{Fuzzy}: a Mamdani controller~\cite{mamdani1975experiment} with two inputs. The first input is $\rho(R)$~\eqref{eq:rho}, the place of the return between $R_{\min}$ and $H_{\Res{cfg/ena/signal/percentile}}$. The second is the return's recent trend: the mean of the last \Res{cfg/ena/fuzzy/trendrecent} returns, minus the mean of up to \Res{cfg/ena/fuzzy/trendbefore} returns before them. The trend is divided by $H_{\Res{cfg/ena/signal/percentile}} - R_{\min}$, the span of $\rho$, and clipped to $[-1, 1]$. It is $0$ until \Res{cfg/ena/fuzzy/trendmin} returns have been seen. When the span is near zero, $\rho$ is $1$ if $R \ge H_{\Res{cfg/ena/signal/percentile}}$ and $0$ otherwise, and the trend is $0$. The controller's \Res{cfg/ena/fuzzy/rules} rules raise Plasticity when the return is low or falling, and lower it when the return is high and steady or rising. Inference is min--max. The output is the centroid of the aggregated set, where Mamdani and Assilian took the mean of maximum. In fuzzy control, the centroid is widely used, and generally gives a lower mean square error than the mean of maximum~\cite{lee1990fuzzy}.
\end{itemize}
Like the ramp, the quadratic and fuzzy controllers return $1$ until \Res{cfg/ena/signal/warmup} returns have been seen, and they stay within $[f_{\min}, 1]$. Section~\ref{sec:controller} compares them. At test, $\mathcal{P}=0$ whatever the controller. When environment detection confirms a change of physics, $H$ is cleared. The ramp, quadratic and fuzzy controllers then search at full effort until they have seen \Res{cfg/ena/signal/warmup} returns of the new regime.

\subsection{Evolutionary Search}
\label{sec:search}
After every training episode, the specialists are ranked by Trust, and ties go to the older specialist. A specialist's age counts the episodes it has played, and a mutant inherits its parent's age. The top \Res{cfg/ena/search/elite}\% form the elite pool. The bottom \Res{cfg/ena/search/replaced}\%, except archived specialists, are replaced. Of the replaced specialists, \Res{cfg/ena/search/mutants}\% become mutants of a random elite and \Res{cfg/ena/search/crossovers}\% become crossovers of two elites. The remaining \Res{cfg/ena/search/fresh}\% are freshly initialized, which keeps the population diverse. In all, with no archived specialist among them, each generation replaces \Res{cfg/ena/search/mutantspop}\%, \Res{cfg/ena/search/crossoverspop}\% and \Res{cfg/ena/search/freshpop}\% of the population with mutants, crossovers and fresh specialists.

A mutant perturbs each weight of its parent with probability $p_m$, by Gaussian noise of standard deviation $\sigma_{\min} + (\sigma_m - \sigma_{\min})\,\mathcal{P}$, with $\sigma_{\min} = \Res{cfg/ena/mutation/minnoise}$ (Appendix~\ref{app:hyperparameters} gives $p_m$ and $\sigma_m$). A crossover takes each weight from either parent with probability $\Res{cfg/ena/crossover/share}$, after both parents are re-scaled to the current input statistics. New specialists start at the reference Trust $T_{\mathrm{ref}}$.

\subsection{Hall of Fame}
\label{sec:hof}
The Hall of Fame stores frozen copies of proven specialists, the \textit{gladiators}. It holds at most $K$ gladiators per \textit{niche}, a niche being a set of physics recognized by environment detection (Section~\ref{sec:detection}). The evolutionary step never replaces a gladiator. After each training episode, the acting specialist $i$ is considered for induction, in five steps:
\begin{enumerate}
\item \textit{Gate.} Its Trust must stand out from the population, $(T_i - \mu_T)/\max(\sigma_T, \Res{cfg/ena/hof/gatefloor}) \ge \Res{cfg/ena/hof/gatez}\,(1 - \Res{cfg/ena/hof/gaterelief}\,\mathcal{P})$, or reach $T_i \ge \Res{cfg/ena/hof/gatetrust}$. Here $\mu_T$ and $\sigma_T$ are the mean and standard deviation of Trust over the population. High Plasticity lowers the induction threshold during active search.
\item \textit{Niche.} Environment detection resolves which niche the episode was played in.
\item \textit{Cloning, not relabeling.} When a gladiator also succeeds in a different niche, it is cloned into that niche. The original retains its home niche assignment.
\item \textit{Diversity.} A candidate may be a near-copy of a gladiator of the same niche, with a cosine similarity of weights above $\Res{cfg/ena/hof/similarity}$. It then replaces that twin only if its Trust is higher. Weight similarity is a cheap but imperfect proxy for behavioral similarity. It catches near-copies, such as a mutant and its parent. Two networks can still behave alike with different weights, for instance with permuted hidden units.
\item \textit{Capacity.} A candidate fills a free slot of its niche, or replaces the niche's weakest gladiator if it is better.
\end{enumerate}
This design decouples policy preservation from ongoing adaptation:
\begin{itemize}
\item \textit{Retention.} A gladiator is a frozen copy. The Trust stored with it never decays. It rises only when the same gladiator passes the gate again in its niche, with a higher Trust. Only a candidate of its own niche with strictly higher Trust can displace it. Gladiators of other niches can never evict it, however long its own physics stay away. Gladiators in the population are also exempt from the evolutionary replacement and from the Trust blend~\eqref{eq:blend}.
\item \textit{Within-niche improvement.} Within a niche, the archive continues to improve. A better candidate takes a free slot or the weakest gladiator's place, and high Plasticity lowers the gate. Success in a new niche creates a clone there, instead of moving the original.
\end{itemize}
Stored Trust is relative to the return history $H$ at the time it was computed~\eqref{eq:normreturn}. $H$ changes as the population improves, and it is cleared at each confirmed change. Trust values stored at different times are therefore not on one scale. A gladiator stored early in a regime, against weak recent returns, can outrank a later and stronger candidate, and keep it out of a full niche. Section~\ref{sec:selection_ablation} compares this ranking with the test returns.

On HalfCheetah, returns vary widely with initial conditions. Admission on a single episode then risks keeping specialists that were merely lucky. There, ENA uses an \textit{Olympic} variant of the Hall of Fame. Every \Res{cfg/cheetahena/olympicsinterval} environment steps, all gladiators of the current niche replay \Res{cfg/cheetahena/olympicsepisodes} full episodes on shared seeds, an \textit{Olympiad}. A gladiator is evicted if its mean Olympic return, normalized like Trust, falls below the reference Trust. The others are confirmed, with that mean return. Each niche holds \Res{cfg/cheetahena/olympicsqualifierseats} extra seats for gladiators that have not competed yet. Only confirmed gladiators are deployed at test.

Before testing, gladiators are re-injected into the population, round-robin across niches, up to a share of \Res{cfg/ena/hof/reinject} of the population. The selection of Section~\ref{sec:selection} then chooses among them.

\subsection{Environment Detection}
\label{sec:detection}
The context-detection idea of da Silva et al.~\cite{dasilva2006dealing} is followed. Each niche $j$ owns a model of the environment's dynamics, and a niche is recognized by how well its model predicts the transitions the agent actually experiences. The models are linear-Gaussian:
\begin{equation}
s_{t+1} - s_t = W_j^{\top}[s_t, a_t, 1] + \varepsilon_t, \qquad \varepsilon_t \sim \mathcal{N}\!\left(0, \mathrm{diag}(\sigma_j^2)\right).
\label{eq:dynamics}
\end{equation}
Here $W_j$ is the niche's matrix of coefficients, and $\sigma_j^2$ holds the variance of its noise in each state dimension. The models are fitted in closed form, by ridge least squares on accumulated sufficient statistics, so the detector is gradient-free as well. The action is conditioned upon so that the evidence does not depend on which specialist is playing, as long as the model fits the states that specialist visits. Every real transition of every episode is evidence.

\subsubsection{Detection}
The detector monitors the standardized residuals of the current niche $j$'s model,
\begin{equation}
z_t = \frac{s_{t+1} - s_t - W_j^{\top}[s_t, a_t, 1]}{\tau\,\sigma_j},
\end{equation}
taken per state dimension. The tolerance $\tau = \Res{cfg/ena/detect/tau}$ widens their variance by $\tau^2$, which absorbs the linear model's own error on nonlinear physics. The monitor is the Predictive-Mixture CuSum (PM-CuSum) of Halme, Poor and Koivunen~\cite{halme2026sequential}, a recent successor of Page's CUSUM~\cite{page1954continuous}. With $q = \mathcal{N}(0, I)$ the pre-change distribution of the residuals $z_t$,
\begin{equation}
S_t = \max(S_{t-1}, 0) + \log \frac{\hat{p}_t(z_t)}{q(z_t)},
\end{equation}
where $\hat{p}_t$ is a Fixed-Share~\cite{herbster1998tracking} mixture of Gaussian predictors, fitted on sliding windows of $\Res{cfg/ena/detect/windows}, \dots$ recent residuals. An alarm is raised when $S_t > \log\Gamma$. Here $\Gamma$ is a target average run length to a false alarm, set to $10^{\Res{cfg/ena/detect/gammaexp}}$ steps. For an exact pre-change distribution, Halme et al. show that this threshold keeps the average run length above $\Gamma$. Here, the pre-change distribution is an online-fitted linear-Gaussian approximation of nonlinear dynamics, so $\log\Gamma$ is only a decision threshold. The values of $\tau$ and $\Gamma$ were set on preliminary CartPole runs, before the study, so that an unchanged environment raised no alarm. Table~\ref{tab:detection} reports the false alarms of the study itself.

A departure from the Gaussian example of~\cite{halme2026sequential} is made in one respect. Each window also estimates a variance, and that variance is one-sided: it is never below the pre-change variance. The rationale is that a wrong dynamics model can change the spread of the residuals, not only their mean, and should fit worse than the right one, not better.

\subsubsection{Identification}
The recent transitions, up to \Res{cfg/ena/detect/recent}, are matched to the niche $j^{*}$ with the highest mean per-step log-likelihood. They are recognized as $j^{*}$ if it explains them within a margin of $m = \Res{cfg/ena/detect/margin}$ nats per step of a model fitted on those same transitions. Otherwise, the physics are new, and a niche is created. Only niches fitted on at least \Res{cfg/ena/detect/minsamples} transitions take part.

\subsubsection{Detect, then verify}
An alarm is only a candidate. After \Res{cfg/ena/detect/verify} post-alarm transitions, the identification test is run against the alarmed niche. If that niche still explains them, the detector treats the alarm as coming from states that its model had not yet covered. The transitions are then added to the model, and the CuSum restarts. Otherwise, a change of physics is declared, and the post-alarm transitions seed the new regime's evidence. During the test, the dynamics models are never updated. The detector still detects and identifies, but it learns nothing new.

\subsubsection{Late identity binding}
A niche is assigned only when a specialist is considered for the Hall of Fame (step~2 above). The assignment uses the transitions of the episode the specialist just played. The resolved niche's model is refined with them, or a new niche is fitted on them.

\subsubsection{Chirp detection (HalfCheetah)}
\label{sec:chirp}
On HalfCheetah, the transition models misfire. On the detection benchmark of Section~\ref{sec:detection_benchmark}, they confirm many more changes, and create many more niches, than there are environments. This happens even when a single fixed gait is played throughout. It happens more often still when the gait changes, as it does in a learning population (another gait, a flip, standing still). The cause was not isolated. Two plausible ones are that a linear model fits the contact-rich dynamics of a running body poorly, and that its evidence is the acting specialist's own transitions.

On HalfCheetah, ENA plays a \textit{chirp} instead: a fixed probing sequence of actions, as in system identification with a known input signal~\cite{ljung1999system}. At every episode end, ENA resets the environment and plays the same \Res{cfg/cheetah/chirp/steps}-step action sequence, whatever any specialist would do. It then compares the observations this produces across niches. Probing an environment to infer its dynamics has also been learned in RL~\cite{zhou2019environment}. ENA's chirp is fixed, so its response cannot depend on the agent.

\textit{What is played.} The excitation is a multiresolution wavelet train. The \Res{cfg/cheetah/chirp/excitationwavelet} wavelet is played at dyadic scales, from a segment of half the chirp down to the finest. The aim is to excite both the slow and the fast responses of the body.

\textit{What is measured.} The response is the observations minus the chirp's starting observation. Each observation channel is decomposed into a \Res{cfg/cheetah/chirp/featurewavelet} \Res{cfg/cheetah/chirp/transform} tree of depth \Res{cfg/cheetah/chirp/level}~\cite{coifman1992entropy}. Like the discrete wavelet transform~\cite{mallat1989theory}, the tree is dyadic and non-overlapping, but it splits every band. This gives \Res{cfg/cheetah/chirp/bands} frequency bands of equal width, and one log-energy is kept per band. In machine fault diagnosis, the energies of wavelet-packet bands are widely used to characterize faults~\cite{yan2014wavelets}. The transforms are computed with PyWavelets~\cite{lee2019pywavelets}. The excitation and the features were chosen by the ablation of Section~\ref{sec:detection_benchmark}, in the three HalfCheetah training environments. ENA's other HalfCheetah settings were chosen on tuning seeds disjoint from the study's replicates.

\textit{How niches are compared.} Each niche $j$ keeps the mean $\bar{\phi}_j$ of its $n_j$ chirps' feature vectors. The variance $\sigma_c^2$ of each feature $c$ is pooled over niches, on the assumption that it comes mainly from the environment's reset noise. The mean $\bar{\phi}$ of the last $w = \Res{cfg/cheetah/chirp/window}$ chirps is tested against the current niche with
\begin{equation}
\Delta_j = \Bigg(\frac{1}{d}\sum_{c=1}^{d} \frac{(\bar{\phi}_c - \bar{\phi}_{j,c})^2}{\sigma_c^2\,(1/w + 1/n_j)}\Bigg)^{1/2},
\label{eq:chirp}
\end{equation}
where $d$ is the number of features. $\Delta_j$ is about 1 under the same physics. Above the margin $\Res{cfg/cheetah/chirp/margin}$, a change is confirmed, and the most recent chirps identify the new physics. They are assigned to the known niche of lowest $\Delta_j$ within the margin, or else to a new niche (in training only). A niche is judged once it holds \Res{cfg/cheetah/chirp/minchirps} chirps. At test, each chirp's Gaussian log-likelihood under each niche is the dynamics evidence $L_j$ of the selection~\eqref{eq:bpr}.

The chirp costs \Res{cfg/cheetah/chirp/steps} steps per episode, \Res{cfg/cheetah/chirp/share}\,\% of a HalfCheetah episode, and these steps count in ENA's budget. The environment's random stream is restored after the chirp, so episode starts are unchanged. CartPole keeps the transition models, which detect its changes (Appendix~\ref{sec:appendix_detection}).

\subsection{Specialist Selection}
\label{sec:selection}
At test, the weights are frozen and $\mathcal{P}=0$. The only adaptation left is the choice of which gladiator acts. This choice is made with Bayesian Policy Reuse (BPR)~\cite{rosman2016bayesian}. ENA's BPR combines two kinds of evidence in one posterior: the dynamics evidence $L_j$ and the observed utility $u$. The niches play the role of BPR's types, and every gladiator is a policy. The agent keeps a belief $\beta$ over the set $\mathcal{J}$ of niches, and plays the gladiator $\pi$ of highest expected utility, $\sum_{j \in \mathcal{J}}\beta(j)\,\mu_{j\pi}$. Here $u = R/R_{\text{solved}}$ is the return relative to the one at which the environment counts as solved, so solving reads as $u = 1$. $\mu_{j\pi}$ is the mean utility of policy $\pi$ in niche $j$. After each episode, the belief is updated:
\begin{equation}
\beta'(j) \propto \beta(j)\; \mathcal{N}\!\left(u;\, \mu_{j\pi}, \Res{cfg/ena/bpr/noise}^2\right)\; \exp(L_j).
\label{eq:bpr}
\end{equation}
$L_j$ is the log-likelihood of the episode's transitions under niche $j$'s dynamics model~\eqref{eq:dynamics}. It is summed over the episode, with the variance widened by $\tau^2$ as in detection. On HalfCheetah, it is the chirp's log-likelihood (Section~\ref{sec:chirp}). The return term is bounded by the range of the utilities, while $L_j$ grows with the number of transitions in the episode. When the niches' models differ, the dynamics can therefore dominate the belief over niches. The returns also update the performance model $\mu_{j\pi}$, which decides which gladiator of a niche plays.

Two changes adapt BPR to this setting.
\begin{itemize}
\item BPR builds its performance model offline, by playing every policy in every type. The model used here is learned online. Its prior is a utility of $\Res{cfg/ena/bpr/ownprior}$ for a gladiator in its own niche and $\Res{cfg/ena/bpr/crossprior}$ elsewhere, worth \Res{cfg/ena/bpr/prior} observation. Every observed utility then updates it, weighted by the posterior. On CartPole, every gladiator thus starts from the same prior in its own niche, and stored Trust only breaks ties among the gladiators not yet tried. On HalfCheetah, the solved return is out of reach. There, a gladiator's own-niche prior is its mean Olympic return relative to $R_{\text{solved}}$, and its prior elsewhere is $\Res{cfg/ena/bpr/crossprior}$ times that.
\item BPR assumes one fixed type per trial. To follow a changing environment, a Fixed-Share step~\cite{herbster1998tracking} mixes back a uniform share after each update: $\beta \leftarrow \Res{cfg/ena/bpr/mix}\,\beta' + \Res{cfg/ena/bpr/switch}/|\mathcal{J}|$. A confirmed change of physics also resets the belief to uniform.
\end{itemize}
At test, the belief starts uniform when the test begins. It is reset to uniform only when the detector confirms a change of physics. As in training, the test environments follow one another unannounced. Three quantities still adapt at test: the belief, the selection's performance model, and the detector's alarms. The weights, Plasticity ($\mathcal{P}=0$) and the detector's models stay fixed. A confirmed change still resets the input statistics. This matters only for a specialist outside the archive, because a gladiator keeps the statistics it was archived with.

Without a Hall of Fame, there is no gladiator to choose from. The agent then plays a random member of its final population in each test episode. This is how the ablation's variants without a Hall of Fame are tested (Section~\ref{sec:variants}). Section~\ref{sec:selection_ablation} compares BPR with two simpler rules: deploying the highest-Trust gladiator of the niche the detector recognizes, and deploying the highest-Trust gladiator overall. Both rank the gladiators by their stored Trust.

\subsection{The Complete Procedure}
\label{sec:procedure}
Fig.~\ref{fig:algorithm} puts the pieces together. Training repeats the training episode in each environment of the training sequence, until that environment's budget is spent. Before the test, the gladiators are re-injected into the population (Section~\ref{sec:hof}), and the belief $\beta$ is set to uniform. The test then repeats the test episode, with the weights and the dynamics models frozen and $\mathcal{P} = 0$. Appendix~\ref{app:hyperparameters} lists every setting.

\begin{figure}[htbp]
\centering
\fbox{\begin{minipage}{0.94\linewidth}
\small
\textit{Training episode}
\begin{algorithmic}[1]
\STATE choose the acting specialist $i$: with probability $\epsilon$, a random specialist that is not a gladiator; otherwise, the one of highest Trust $T_i$
\STATE play one episode with $\pi_i$, feeding every transition to the detector (on HalfCheetah, play the chirp at the episode's end); let $R$ be its return~\eqref{eq:return}
\IF{the detector confirmed a change of physics}
\STATE start a new regime: reset the detector's evidence, the input statistics, $H$ and $R_{\min}$
\ENDIF
\STATE add $R$ to $H$; \ $R_{\min} \leftarrow \min(R_{\min}, R)$
\STATE $D \leftarrow f(R)$~\eqref{eq:ramp}; \ $T_i \leftarrow (1-D)\,T_i + D\,\tilde{R}(R)$~\eqref{eq:trust}
\STATE $\mathcal{P} \leftarrow$ the controller's output for $R$ (Section~\ref{sec:controllers})
\STATE Hall of Fame: gate $i$, resolve its niche, then archive or clone it (Section~\ref{sec:hof})
\IF{$D > \Res{cfg/ena/trust/reset}$}
\STATE blend the Trust of every specialist except the gladiators toward $T_{\mathrm{ref}}$~\eqref{eq:blend}
\ENDIF
\STATE on HalfCheetah, hold an Olympiad when one is due (Section~\ref{sec:hof})
\STATE draw $k$ candidates among the other $N-1$ specialists, and play each for at most $h$ steps~\eqref{eq:plasticity}; update each candidate's Trust with~\eqref{eq:trust}, at the rate $f$ of its own return
\STATE keep the elites, and replace the rest by mutants, crossovers and fresh specialists; gladiators are never replaced (Section~\ref{sec:search})
\end{algorithmic}
\smallskip
\textit{Test episode}
\begin{algorithmic}[1]
\STATE play the gladiator $\pi$ of highest expected utility $\sum_{j \in \mathcal{J}}\beta(j)\,\mu_{j\pi}$ for one episode, feeding every transition to the detector (on HalfCheetah, play the chirp at the episode's end)
\STATE update the belief with~\eqref{eq:bpr} and the performance model $\mu_{j\pi}$; mix the belief with the Fixed-Share step
\IF{the detector confirmed a change of physics}
\STATE reset $\beta$ to uniform, and reset the input statistics
\ENDIF
\end{algorithmic}
\end{minipage}}
\caption{One ENA training episode and one test episode. Every symbol is defined in Sections~\ref{sec:problem} to~\ref{sec:selection}. Without a Hall of Fame, a test episode plays a random member of the final population instead.}
\label{fig:algorithm}
\end{figure}
\section{Experimental Setup}
\label{sec:setup}

ENA was evaluated in two non-stationary domains built on Gymnasium~\cite{towers2024gymnasium}: the discrete-action \texttt{CartPole-v1}~\cite{barto1983neuronlike}, and the continuous-control MuJoCo~\cite{todorov2012mujoco} \texttt{HalfCheetah-v5}. In both domains, the agent trains on three environments in sequence, without being told when the environment changes. It is then tested, frozen, on five environments, two of which it has never seen.

\subsection{Environments}

\subsubsection{CartPole}
The agent balances a pole hinged on a cart by pushing the cart left or right. Every step earns the same reward, so the return~\eqref{eq:return} counts the steps the episode lasts. Each environment changes the gravity $g$, the pole mass $m_p$ and the pole half-length $\ell$ (Table~\ref{tab:cartpole_worlds}). The environments are named after the planet whose gravity they use. The training environments are also named after their pole (Mars-Light, Jupiter-Heavy, Earth-Standard), and in the text they are abbreviated to Mars, Jupiter and Earth. The training sequence is Mars, Jupiter, Earth. The test sequence interleaves them with two unseen environments: Mars, Neptune, Jupiter, Brown-dwarf, Earth. Neptune is close to Earth. Brown-dwarf is a nominal high-gravity case, with about twice Jupiter's gravity. An episode ends when the pole falls, or after \Res{cfg/cartpole/maxsteps} steps. A return of \Res{cfg/cartpole/solved} or more counts as a success, which is Gymnasium's registered reward threshold for \texttt{CartPole-v1}.

\begin{table}[htbp]
\caption{CartPole Environments}
\begin{center}
\footnotesize
\setlength{\tabcolsep}{3pt}
\begin{adjustbox}{max width=\columnwidth}
\begin{tabular}{|l|c|c|c|c|}
\hline
\textbf{Environment} & \textbf{$g$ (m/s$^2$)} & \textbf{$m_p$ (kg)} & \textbf{$\ell$ (m)} & \textbf{Seen in training} \\
\hline
Mars-Light & 3.7 & 0.05 & 0.1 & yes \\
Jupiter-Heavy & 24.8 & 0.5 & 1.5 & yes \\
Earth-Standard & 9.8 & 0.1 & 0.5 & yes \\
Neptune & 11.15 & 0.1 & 0.5 & no \\
Brown-dwarf & 50 & 0.1 & 0.5 & no \\
\hline
\end{tabular}
\end{adjustbox}
\label{tab:cartpole_worlds}
\end{center}
\end{table}
 
\subsubsection{HalfCheetah}
The agent controls the six joints of a planar two-legged robot, and is rewarded for running forward. The reward of each step is the robot's forward velocity minus a cost on the squared joint torques, with Gymnasium's default weights. Each environment changes the gravity, the ground friction and the wind (Table~\ref{tab:cheetah_worlds}). The friction is applied to every contact surface, since MuJoCo uses the larger friction of two touching bodies. The wind acts through air, which only the windy environments have. The training sequence is Earth (Earth-Standard in the tables) with a soft head wind, then Heavy-Rough (heavier gravity, on rough ground), then Mars-Slippery (Mars's gravity on a slippery floor). The test sequence is Earth-Standard, Ice-Floor, Heavy-Rough, High-Wind and Mars-Slippery. The two unseen environments retain Earth's gravity. Ice-Floor has a much more slippery floor, and neither wind nor air. High-Wind differs from Earth-Standard only by a stronger head wind.

Gymnasium's registered reward threshold for \texttt{HalfCheetah-v5} is \Res{cfg/cheetah/solved}, and the figures show it as a reference line. On the non-stationary stream, ENA never reaches it at test, PPO almost never, and SAC in only part of its test episodes (Section~\ref{sec:cheetah}). Reliability therefore cannot show forgetting on HalfCheetah. \textit{Retention} and \textit{absolute forgetting} (Section~\ref{sec:metrics}) replace it as the primary metrics.

\begin{table}[htbp]
\caption{HalfCheetah Environments}
\begin{center}
\footnotesize
\setlength{\tabcolsep}{3pt}
\begin{adjustbox}{max width=\columnwidth}
\begin{tabular}{|l|c|c|c|c|c|}
\hline
\textbf{Environment} & \textbf{$g$ (m/s$^2$)} & \textbf{Friction} & \textbf{Wind (m/s)} & \textbf{Air (kg/m$^3$)} & \textbf{Seen} \\
\hline
Earth-Standard & 9.8 & 0.4 & \ensuremath{-}5 & 1.2 & yes \\
Heavy-Rough & 14.7 & 1 & 0 & 0 & yes \\
Mars-Slippery & 3.7 & 0.2 & 0 & 0 & yes \\
Ice-Floor & 9.8 & 0.05 & 0 & 0 & no \\
High-Wind & 9.8 & 0.4 & \ensuremath{-}20 & 1.2 & no \\
\hline
\end{tabular}
\end{adjustbox}
\label{tab:cheetah_worlds}
\end{center}
\end{table}
 ENA detects HalfCheetah's environments with its chirp (Section~\ref{sec:chirp}), instead of the transition models.

\subsection{Protocols}
\label{sec:protocols}

\subsubsection{Skill ablation (CartPole)}
Each ENA variant (Section~\ref{sec:variants}, Table~\ref{tab:variants}) trains for \Res{cfg/cartpole/train} episodes per environment. It is then tested for \Res{cfg/cartpole/test} episodes per environment. The budget is a number of episodes, not of environment steps. The variants with Plasticity evaluate many more candidates, so they also interact far more with the environment (Section~\ref{sec:ablation}).

A total of \Res{ablation/n} independent replicates were run. Replicate $k$ ($k = 1, 2, \dots$) seeds Python's, NumPy's and TensorFlow's random number generators with $\Res{cfg/cartpole/seedoffsetraw} + k - 1$ before each agent is built, trained and tested. Every environment is seeded with numbers derived from the same seed. All the agents of a replicate therefore start from the same random state and meet the same initial states: they differ only in what they are. Every study in this paper seeds its replicates the same way.

\subsubsection{Forgetting in single-policy learners}
\label{sec:forgetting_protocol}
To measure how much standard learners forget on these streams, single-policy learners train on the same streams as ENA. Comparing their forgetting with ENA's is only fair at equal interaction. Every agent therefore trains on each environment for the same budget of environment steps: \Res{cfg/challenge/budget} on CartPole and \Res{cfg/cheetah/budget} on HalfCheetah. ENA's budget includes every step it spends evaluating candidates~\eqref{eq:plasticity}, holding Olympiads and playing its chirp (Section~\ref{sec:chirp}), not only the episodes of its acting specialist. Each agent stops at the end of the episode in which it reaches the budget. ENA's episode includes these extra steps, so ENA can exceed the budget slightly.

ENA spends \Res{challenge/ENA-Combined/candshare/mean}\% of its CartPole training steps evaluating candidates. It therefore plays far fewer episodes within the same budget: about \Res{challenge/ENA-Combined/episodes/perenv} per environment on CartPole, against \Res{challenge/DQN/episodes/perenv} for DQN, \Res{challenge/PPO/episodes/perenv} for PPO and \Res{challenge/SAC-Discrete/episodes/perenv} for SAC-Discrete. On HalfCheetah, candidate evaluation, Olympiads and chirps take most of its steps. ENA plays about \Res{cheetah/ENA-Combined/episodes/perenv} episodes per environment there, against \Res{cheetah/PPO/episodes/perenv} for PPO and \Res{cheetah/SAC/episodes/perenv} for SAC. Training is compared against environment steps, not episodes, so that every agent spans the same budget. The frozen test needs no such normalization, since every agent plays the same number of test episodes. No agent observes the physical parameters, or is told when the environment changes. Appendix~\ref{app:information} lists what each agent receives.

On CartPole, the agent is evaluated, frozen, every \Res{cfg/challenge/checkevery} training steps; ENA is evaluated through its highest-Trust specialist. Each check plays \Res{cfg/challenge/evalepisodes} episodes on a separate copy of the current environment. The checks never stop or alter training, and their steps do not count as training. They record the \textit{first solve}: the number of training steps into an environment at which the evaluation mean first reaches the success threshold. A replicate that never solves an environment is counted at the end of its phase. HalfCheetah uses no intermediate checks. First solve is not measured there, since ENA never reaches its threshold and PPO almost never does.

After training, all agents are tested frozen. On CartPole they play \Res{cfg/cartpole/test} episodes per environment, as in the ablation, and on HalfCheetah \Res{cfg/cheetah/test}. The CartPole comparison runs \Res{challenge/n} replicates. Its \Res{challenge/family} comparisons, ENA-Combined against each baseline on test reliability, form one Holm family. This family was planned before the runs. Since the single-policy learners serve only as a reference, the result is read as the overall cost of their forgetting, not as a ranking of the agents. The HalfCheetah comparison runs \Res{cheetah/n} replicates. Its two comparisons, ENA-Combined against PPO and against SAC on left-behind retention, form one Holm family.

\subsection{Agents}

\subsubsection{ENA variants}
\label{sec:variants}
The three skills under study are switched on and off as in Table~\ref{tab:variants}, and every other setting is shared. ``Plasticity off'' means the zero controller (Section~\ref{sec:controllers}): $\mathcal{P} = 0$, the minimum search. Besides fewer and shorter candidate evaluations, this changes the candidate fitness signal. On CartPole, a candidate that survives \Res{cfg/ena/smin} steps is then credited with a full episode (Section~\ref{sec:trust}). ``Plasticity on'' means the linear ramp.

Environment detection organizes the Hall of Fame's archive into niches, so it is never tested without it. Switching it on changes three things at once:
\begin{itemize}
\item \textit{the index}: each gladiator is stored, and recalled at test, under the niche it was proven in;
\item \textit{the capacity}: the archive holds $K$ gladiators per niche, instead of $K$ in all (Section~\ref{sec:ablation} gives the sizes reached);
\item \textit{the resets}: each confirmed change clears $H$ and $R_{\min}$ and resets the input statistics. Trust's rate, and under the ramp Plasticity, then return to 1 for \Res{cfg/ena/signal/warmup} episodes (Section~\ref{sec:trust}).
\end{itemize}
Without detection, no change is ever confirmed, so none of these resets happens. The ablation does not separate the three.

At test, the variants with a Hall of Fame choose among their gladiators by Bayesian policy reuse. The variants without one have no gladiator to choose from. They play a random member of their final population in each test episode (Section~\ref{sec:selection}). Adding the Hall of Fame therefore changes both what is kept and how the test policy is chosen. A random agent gives the floor. The full configuration is listed in Appendix~\ref{app:hyperparameters}.

\begin{table}[htbp]
\caption{ENA Variants in the Skill Ablation}
\begin{center}
\begin{tabular}{|l|c|c|c|}
\hline
\textbf{Variant} & \textbf{Plasticity} & \textbf{HoF} & \textbf{Detection} \\
\hline
ENA-NoSkill & -- & -- & -- \\
ENA-Plasticity & \checkmark & -- & -- \\
ENA-HoF & -- & \checkmark & -- \\
ENA-HoF+EnvDetection & -- & \checkmark & \checkmark \\
ENA-Plasticity+HoF & \checkmark & \checkmark & -- \\
ENA-Combined & \checkmark & \checkmark & \checkmark \\
\hline
\end{tabular}
\label{tab:variants}
\end{center}
\end{table}

\subsubsection{Single-policy baselines}
On CartPole, ENA-Combined was compared with DQN~\cite{mnih2015human}, PPO~\cite{schulman2017proximal} and the discrete-action variant of SAC~\cite{christodoulou2019soft}. On HalfCheetah, it was compared with PPO and SAC~\cite{haarnoja2018soft}. All are custom implementations, with the default hyperparameters of these implementations (Appendix~\ref{app:hyperparameters}). The non-stationary stream made two changes necessary, both found in full-length runs before the reported study:
\begin{itemize}
\item \textit{PPO uses a constant learning rate.} With the usual linear annealing to zero over the training budget, PPO reaches its last environment with almost no learning rate left. A learner in a non-stationary stream has no reason to learn less in a later environment than in an earlier one.
\item \textit{SAC-Discrete uses a target entropy of $\Res{cfg/challenge/sacdratio}\log|\mathcal{A}|$}, instead of $\Res{cfg/sacd/paperratio}\log|\mathcal{A}|$. With CartPole's two actions, $\Res{cfg/sacd/paperratio}\log 2$ is nearly the maximum entropy. The policy cannot reach it, and the temperature keeps growing. Over \Res{sacd/0.98/steps} steps of the first training environment (median of \Res{sacd/0.98/seeds} seeds), the temperature rose from \Res{sacd/0.98/alphastart} to \Res{sacd/0.98/alphamax}, and the critics' values to \Res{sacd/0.98/valuemax}. A ratio of $\Res{cfg/sacd/cleanrlratio}$ diverged in the same way, with a temperature of \Res{sacd/0.89/alphamax} and values of \Res{sacd/0.89/valuemax}. At $\Res{cfg/challenge/sacdratio}$, the temperature peaked at \Res{sacd/0.5/alphamax} and the values at \Res{sacd/0.5/valuemax}.
\end{itemize}
During the frozen test, the baselines act deterministically with their final parameters, SAC with the mean of its policy. Section~\ref{sec:selection} describes what still adapts in ENA at test.

\subsection{Metrics}
\label{sec:metrics}
\begin{itemize}
\item \textit{Reliability} (CartPole): the percentage of test episodes whose return reaches the success threshold, over all five test environments. The tables also give it over the three training environments alone, which leaves out the unseen ones.
\item \textit{Retention}: the test return as a percentage of the end-of-phase training return. It is computed for each training environment, and averaged over the environments left behind, that is, all training environments except the last. It is a ratio per replicate, so its mean can differ from the ratio of the mean returns. The end-of-phase return is the mean return of the training episodes that end in the last tenth of the phase: of its step budget in the comparisons, and of its episodes in the ablation. Retention is the primary metric on HalfCheetah, and it is reported descriptively on CartPole. End-of-phase returns include every agent's training exploration, so retention can exceed 100\%. For an environment the agent never learned, retention measures nothing.
\item \textit{Absolute forgetting}: the end-of-phase training return minus the test return, in return units. It is computed for each training environment, and averaged over the environments left behind. Unlike retention, it does not reward an agent for having learned little. Its returns tell a weak policy kept intact from a strong policy partly lost.
\item \textit{Mean return} per environment.
\item \textit{Steps to first solve} per training environment (CartPole comparison only), at the \Res{cfg/challenge/checkevery}-step resolution of the evaluation checks.
\item \textit{Compute}: training environment steps, and ENA's candidate-evaluation steps among them, which are properties of the algorithms; and wall-clock training time, which also depends on the implementations and hardware (Appendix~\ref{app:compute}).
\end{itemize}
Every metric is computed per replicate, and reported as the mean and standard deviation over replicates. All the agents of a replicate share its seed (Section~\ref{sec:protocols}), so every comparison is paired by replicate. Each difference between two agents is reported in three ways: the mean of the within-replicate differences, with its \Res{stats/ci}\% confidence interval; the effect size $d_z$, which is that mean over the standard deviation of the differences; and a paired $t$-test at $\alpha = \Res{stats/alpha}$. Steps to first solve are the exception. A few slow replicates dominate their mean, so they are reported as the median and interquartile range, and compared with the two-sided Wilcoxon signed-rank test.

The skill ablation makes \Res{ablation/family} planned comparisons: each skill is added on its own, and on top of the others. They are treated as one family, and their $p$-values are corrected with Holm's method. The number of replicates was fixed before the runs, by a power analysis for two independent groups. With \Res{ablation/n} replicates per variant, a two-sided test at the family's most stringent level ($\Res{stats/alpha}/\Res{ablation/family}$) has \Res{stats/power}\% power for effects of $d \ge \Res{ablation/plannedmde}$. Here $d$ is the difference of means over the root mean square of the two standard deviations. Analyzed paired, the family has \Res{stats/power}\% power for $d_z \ge \Res{ablation/mde}$. Welch's unpaired $t$-test reaches the same decision in \Res{stats/welch/agree} of the \Res{stats/welch/planned} Holm-corrected comparisons of this paper.

The selection study forms its own Holm family (Section~\ref{sec:selection_ablation}). The Plasticity-controller study forms one family per metric (Section~\ref{sec:controller}). The HalfCheetah comparison on absolute forgetting, ENA-Combined against PPO and SAC, forms another family, defined after the runs. Retention and absolute forgetting on CartPole, in the ablation and in the comparison, are descriptive and uncorrected, and so is every per-environment comparison.

\subsection{Research Questions and Evidence}
Table~\ref{tab:rq_map} maps each research question to the study that answers it. The skill ablation is the primary evidence for RQ1.

\begin{table}[htbp]
\caption{Research Questions and the Evidence for Each}
\begin{center}
\footnotesize
\setlength{\tabcolsep}{3pt}
\begin{tabular}{|l|p{0.62\columnwidth}|l|}
\hline
\textbf{RQ} & \textbf{Evidence} & \textbf{Section} \\
\hline
RQ1 & skill ablation (reliability and retention); forgetting of single-policy learners & \ref{sec:ablation}, \ref{sec:forgetting_cartpole}, \ref{sec:cheetah} \\
RQ1a & ablation: Hall of Fame with and without detection; selection ablation & \ref{sec:ablation}, \ref{sec:selection_ablation} \\
RQ1b & ablation: candidate steps with and without the archive & \ref{sec:ablation} \\
RQ1c & recognition per phase (both domains); detection benchmark (HalfCheetah) & \ref{sec:appendix_detection}, \ref{sec:detection_benchmark} \\
RQ2 & ablation; compute study; controller study & \ref{sec:ablation}, \ref{sec:compute}, \ref{sec:controller} \\
\hline
\end{tabular}
\label{tab:rq_map}
\end{center}
\end{table}
\section{Results}
\label{sec:results}

This section first evaluates the individual contributions of ENA's constituent skills and its specialist selection mechanism through ablation studies on CartPole. Performance is subsequently analyzed on previously encountered and unseen environments alongside single-policy baselines, accompanied by the detection benchmarks on HalfCheetah. Finally, the interaction costs of failure-driven search and alternative controller formulations are assessed. Table~\ref{tab:rq_map} maps each subsection to the research question it informs. Reporting conventions and statistical tests are detailed in Section~\ref{sec:metrics}; unless stated otherwise, reported values are means over independent replicates with standard deviations in parentheses, and significance is assessed with the two-sided paired $t$-test.

\subsection{Skill Ablation}
\label{sec:ablation}
\begin{figure}[htbp]
\centerline{\includegraphics[width=\linewidth]{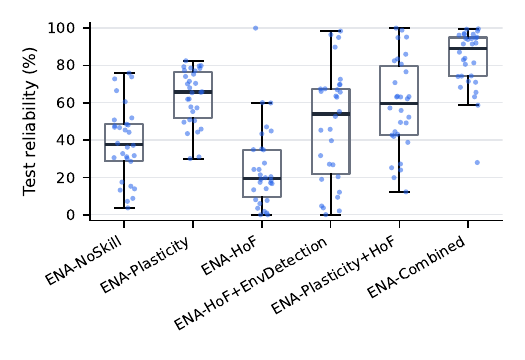}}
\caption{Test reliability of each ENA variant; each dot is one of the \Res{ablation/n} replicates.}
\label{fig:ablation_reliability}
\end{figure}

Table~\ref{tab:ablation_test} and Fig.~\ref{fig:ablation_reliability} report the frozen test over \Res{ablation/n} replicates. With all three skills, \textbf{ENA-Combined} achieved a test reliability of \Res{ablation/ENA-Combined/test/rel/mean}\% (\Res{ablation/ENA-Combined/test/rel/std}). Without any skill, ENA-NoSkill achieved \Res{ablation/ENA-NoSkill/test/rel/mean}\% (\Res{ablation/ENA-NoSkill/test/rel/std}). The difference is \Res{ablation/ENA-NoSkill-vs-ENA-Combined/test/rel/diff} points (\Res{stats/ci}\% CI \Res{ablation/ENA-NoSkill-vs-ENA-Combined/test/rel/ci/low} to \Res{ablation/ENA-NoSkill-vs-ENA-Combined/test/rel/ci/high}; $d_z = \Res{ablation/ENA-NoSkill-vs-ENA-Combined/test/rel/d}$; \Res{ablation/ENA-NoSkill-vs-ENA-Combined/test/rel/p}, uncorrected). ENA-Combined is more reliable than every other variant. Its smallest margin is over ENA-Plasticity: \Res{ablation/ENA-Plasticity-vs-ENA-Combined/test/rel/diff} points (\Res{ablation/ENA-Plasticity-vs-ENA-Combined/test/rel/ci/low} to \Res{ablation/ENA-Plasticity-vs-ENA-Combined/test/rel/ci/high}; \Res{ablation/ENA-Plasticity-vs-ENA-Combined/test/rel/p}, uncorrected).

Table~\ref{tab:ablation_family} gives the \Res{ablation/family} planned comparisons, which add one skill at a time. They form one Holm family, and each difference is in points of test reliability. In all three pairs of variants that differ only by Plasticity, the variant with Plasticity is more reliable. In both pairs that differ only by environment detection, the variant with detection is more reliable. ENA-HoF is less reliable than ENA-NoSkill (\Res{ablation/ENA-NoSkill-vs-ENA-HoF/test/rel/diff} points, Holm \Res{ablation/ENA-NoSkill-vs-ENA-HoF/test/rel/pholm}), and ENA-Plasticity+HoF does not differ significantly from ENA-Plasticity (\Res{ablation/ENA-Plasticity-vs-ENA-Plasticity+HoF/test/rel/diff}, Holm \Res{ablation/ENA-Plasticity-vs-ENA-Plasticity+HoF/test/rel/pholm}). As Section~\ref{sec:variants} notes, each Hall of Fame comparison also changes how the test policy is chosen. Each detection comparison also changes the archive's capacity and adds the resets. Without detection, the archive holds at most $K = \Res{cfg/ena/hof/K}$ gladiators, and ENA-HoF and ENA-Plasticity+HoF fill it in every replicate. With detection, ENA-HoF+EnvDetection ends training with \Res{ablation/ENA-HoF+EnvDetection/hof/gladiators/mean} gladiators on average (\Res{ablation/ENA-HoF+EnvDetection/hof/gladiators/min} to \Res{ablation/ENA-HoF+EnvDetection/hof/gladiators/max}), in \Res{ablation/ENA-HoF+EnvDetection/hof/niches/mean} niches. ENA-Combined ends with \Res{ablation/ENA-Combined/hof/gladiators/mean} (\Res{ablation/ENA-Combined/hof/gladiators/min} to \Res{ablation/ENA-Combined/hof/gladiators/max}), in \Res{ablation/ENA-Combined/hof/niches/mean} niches.

\textit{Acquisition and retention.} Reliability reflects both performance acquisition within an environment and retention across subsequent environment switches. Table~\ref{tab:ablation_retention} separates these aspects on the environments left behind, Mars and Jupiter. It reports each variant's end-of-phase and test returns, its retention, and its absolute forgetting. Without Plasticity, ENA achieves minimal learning on Jupiter. ENA-NoSkill ends its Jupiter phase at a return of \Res{ablation/ENA-NoSkill/end/Jupiter-Heavy/mean}, and scores \Res{ablation/ENA-NoSkill/test/Jupiter-Heavy/mean} there at test. Its retention above 100\% indicates only that no acquired capability was degraded. The three variants with Plasticity acquire policies in both environments, but retention differs across architectures. ENA-Plasticity retains \Res{ablation/ENA-Plasticity/retention/Mars-Light/mean}\% of its Mars return and \Res{ablation/ENA-Plasticity/retention/Jupiter-Heavy/mean}\% of its Jupiter return. ENA-Plasticity+HoF retains Mars (\Res{ablation/ENA-Plasticity+HoF/retention/Mars-Light/mean}\%), but only \Res{ablation/ENA-Plasticity+HoF/retention/Jupiter-Heavy/mean}\% of Jupiter. ENA-Combined maintains performance in both, with \Res{ablation/ENA-Combined/retention/Mars-Light/mean}\% and \Res{ablation/ENA-Combined/retention/Jupiter-Heavy/mean}\%. These retention values are descriptive.

The archive organized by detection can also be compared with no archive at all. With Plasticity, ENA-Combined is more reliable than ENA-Plasticity, as reported above. Without Plasticity, the difference is not resolved: ENA-HoF+EnvDetection is \Res{ablation/ENA-NoSkill-vs-ENA-HoF+EnvDetection/test/rel/diff} points above ENA-NoSkill (\Res{ablation/ENA-NoSkill-vs-ENA-HoF+EnvDetection/test/rel/ci/low} to \Res{ablation/ENA-NoSkill-vs-ENA-HoF+EnvDetection/test/rel/ci/high}; \Res{ablation/ENA-NoSkill-vs-ENA-HoF+EnvDetection/test/rel/p}, uncorrected).

\textit{Unseen environments.} On Neptune, ENA-Combined's mean return is \Res{ablation/ENA-Combined/test/Neptune/mean}. Its detector recognizes Neptune as Earth's niche in \Res{ablation/ENA-Combined/recog/Neptune/Earth-Standard} of the \Res{ablation/n} replicates. On Brown-dwarf, its mean return is \Res{ablation/ENA-Combined/test/Brown-dwarf/mean} (\Res{ablation/ENA-Combined/test/Brown-dwarf/std}), and the detector leaves the environment unrecognized in \Res{ablation/ENA-Combined/recog/Brown-dwarf/unrecognized} of them. There, it also confirms \Res{ablation/ENA-Combined/detect/test/Brown-dwarf/changes/mean} changes per replicate on average (\Res{ablation/ENA-Combined/detect/test/Brown-dwarf/changes/std}). At most one of them follows the real switch, and each one resets the selection's belief to uniform. Table~\ref{tab:detection} (Appendix~\ref{sec:appendix_detection}) gives the recognition per phase: confirmed changes, unrecognized episodes and niches per environment.

\textit{Interaction and cost of the archive.} The ablation fixes episodes, not steps, so the variants with Plasticity interact far more with the environment. ENA-NoSkill, ENA-HoF and ENA-HoF+EnvDetection take \Res{ablation/ENA-NoSkill/adapt/Train-Env-Steps/mean}, \Res{ablation/ENA-HoF/adapt/Train-Env-Steps/mean} and \Res{ablation/ENA-HoF+EnvDetection/adapt/Train-Env-Steps/mean} training steps per replicate. ENA-Plasticity, ENA-Plasticity+HoF and ENA-Combined take \Res{ablation/ENA-Plasticity/adapt/Train-Env-Steps/mean}, \Res{ablation/ENA-Plasticity+HoF/adapt/Train-Env-Steps/mean} and \Res{ablation/ENA-Combined/adapt/Train-Env-Steps/mean}.

Adding the archive and detection changes the candidate-evaluation steps as follows (exploratory, uncorrected). Without Plasticity, ENA-NoSkill spends \Res{ablation/ENA-NoSkill/compute/candsteps/mean} candidate steps per replicate. Adding the Hall of Fame changes this by \Res{ablation/ENA-NoSkill-vs-ENA-HoF/compute/candsteps/diff} (\Res{ablation/ENA-NoSkill-vs-ENA-HoF/compute/candsteps/ci/low} to \Res{ablation/ENA-NoSkill-vs-ENA-HoF/compute/candsteps/ci/high}; \Res{ablation/ENA-NoSkill-vs-ENA-HoF/compute/candsteps/p}), and adding detection on top changes it by \Res{ablation/ENA-HoF-vs-ENA-HoF+EnvDetection/compute/candsteps/diff} (\Res{ablation/ENA-HoF-vs-ENA-HoF+EnvDetection/compute/candsteps/ci/low} to \Res{ablation/ENA-HoF-vs-ENA-HoF+EnvDetection/compute/candsteps/ci/high}; \Res{ablation/ENA-HoF-vs-ENA-HoF+EnvDetection/compute/candsteps/p}). With Plasticity, ENA-Plasticity spends \Res{ablation/ENA-Plasticity/compute/candsteps/mean} candidate steps. Adding the Hall of Fame changes this by \Res{ablation/ENA-Plasticity-vs-ENA-Plasticity+HoF/compute/candsteps/diff} (\Res{ablation/ENA-Plasticity-vs-ENA-Plasticity+HoF/compute/candsteps/ci/low} to \Res{ablation/ENA-Plasticity-vs-ENA-Plasticity+HoF/compute/candsteps/ci/high}; \Res{ablation/ENA-Plasticity-vs-ENA-Plasticity+HoF/compute/candsteps/p}), and adding detection on top changes it by \Res{ablation/ENA-Plasticity+HoF-vs-ENA-Combined/compute/candsteps/diff} (\Res{ablation/ENA-Plasticity+HoF-vs-ENA-Combined/compute/candsteps/ci/low} to \Res{ablation/ENA-Plasticity+HoF-vs-ENA-Combined/compute/candsteps/ci/high}; \Res{ablation/ENA-Plasticity+HoF-vs-ENA-Combined/compute/candsteps/p}). Wall-clock training takes \Res{ablation/ENA-NoSkill/time/min/mean}, \Res{ablation/ENA-HoF/time/min/mean} and \Res{ablation/ENA-HoF+EnvDetection/time/min/mean} minutes per replicate for ENA-NoSkill, ENA-HoF and ENA-HoF+EnvDetection.

\textit{Statistical power.} With \Res{ablation/n} replicates, the family detects $d_z \ge \Res{ablation/mde}$ with \Res{stats/power}\% power (Section~\ref{sec:metrics}). The five comparisons that add Plasticity or detection have $d_z$ between \Res{ablation/ENA-HoF-vs-ENA-HoF+EnvDetection/test/rel/d} and \Res{ablation/ENA-HoF-vs-ENA-Plasticity+HoF/test/rel/d}. Each of their intervals excludes zero, as does ENA-Combined's margin over each variant. The Hall of Fame's effect on ENA-NoSkill is significant, but smaller than the design detects ($d_z = \Res{ablation/ENA-NoSkill-vs-ENA-HoF/test/rel/d}$). Measuring it with \Res{stats/power}\% power would take \Res{ablation/ENA-NoSkill-vs-ENA-HoF/test/rel/nreq} replicates. Its effect on ENA-Plasticity ($d_z = \Res{ablation/ENA-Plasticity-vs-ENA-Plasticity+HoF/test/rel/d}$) would take \Res{ablation/ENA-Plasticity-vs-ENA-Plasticity+HoF/test/rel/nreq}.

\begin{table}[htbp]
\caption{Skill Ablation: Frozen Test (30 Replicates)}
\begin{center}
\footnotesize
\setlength{\tabcolsep}{3pt}
\begin{adjustbox}{max width=\columnwidth}
\begin{tabular}{|l|c|c|c|c|c|c|c|}
\hline
\textbf{Agent} & \textbf{Mars} & \textbf{Neptune} & \textbf{Jupiter} & \textbf{Brown} & \textbf{Earth} & \textbf{Rel. (\%)} & \textbf{Train rel. (\%)} \\
\hline
Random & 9.0 & 21.8 & 33.7 & 16.3 & 22.2 & 0.0 $\pm$ 0.0 & 0.0 \\
ENA-NoSkill & 311.8 & 321.7 & 144.8 & 170.1 & 321.4 & 38.5 $\pm$ 20.0 & 40.1 \\
ENA-Plasticity & 279.8 & 416.3 & 350.7 & 313.1 & 419.4 & 63.3 $\pm$ 14.8 & 63.6 \\
ENA-HoF & 326.6 & 252.4 & 128.3 & 108.3 & 299.9 & 24.7 $\pm$ 21.9 & 30.2 \\
ENA-HoF+EnvDetection & 404.5 & 393.7 & 272.5 & 208.5 & 414.8 & 48.3 $\pm$ 30.3 & 53.7 \\
ENA-Plasticity+HoF & 474.8 & 352.1 & 277.9 & 281.5 & 450.7 & 58.9 $\pm$ 24.6 & 69.9 \\
ENA-Combined & 475.5 & 490.4 & 464.9 & 367.2 & 490.5 & 83.2 $\pm$ 16.1 & 87.5 \\
\hline
\end{tabular}
\end{adjustbox}
\label{tab:ablation_test}
\end{center}
\end{table}
\begin{table}[htbp]
\caption{Skill Ablation: Planned Comparisons (30 Replicates)}
\begin{center}
\footnotesize
\setlength{\tabcolsep}{3pt}
\begin{adjustbox}{max width=\columnwidth}
\begin{tabular}{|l|l|c|c|c|c|}
\hline
\textbf{Skill added} & \textbf{To} & \textbf{Difference} & \textbf{95\% CI} & \textbf{$d_z$} & \textbf{Holm} \\
\hline
Plasticity & ENA-NoSkill & +24.8 & 14.3 to 35.2 & 0.88 & $p<0.001$ \\
Hall of Fame & ENA-NoSkill & \ensuremath{-}13.8 & \ensuremath{-}25.0 to \ensuremath{-}2.5 & \ensuremath{-}0.46 & $p=0.036$ \\
Detection & ENA-HoF & +23.5 & 8.7 to 38.3 & 0.59 & $p=0.009$ \\
Hall of Fame & ENA-Plasticity & \ensuremath{-}4.4 & \ensuremath{-}13.0 to 4.3 & \ensuremath{-}0.19 & $p=0.311$ \\
Plasticity & ENA-HoF & +34.2 & 24.0 to 44.4 & 1.25 & $p<0.001$ \\
Detection & ENA-Plasticity+HoF & +24.3 & 13.5 to 35.1 & 0.84 & $p<0.001$ \\
Plasticity & ENA-HoF+EnvDetection & +35.0 & 22.9 to 47.0 & 1.08 & $p<0.001$ \\
\hline
\end{tabular}
\end{adjustbox}
\label{tab:ablation_family}
\end{center}
\end{table}
\begin{table*}[t]
\caption{Skill Ablation: Retention of the Environments Left Behind (30 Replicates)}
\begin{center}
\footnotesize
\setlength{\tabcolsep}{3pt}
\begin{adjustbox}{max width=\textwidth}
\begin{tabular}{|l|c|c|c|c|c|c|c|c|}
\hline
\textbf{Agent} & \textbf{Mars end} & \textbf{Mars test} & \textbf{Mars ret. (\%)} & \textbf{Jupiter end} & \textbf{Jupiter test} & \textbf{Jupiter ret. (\%)} & \textbf{Retention (\%)} & \textbf{Forgetting} \\
\hline
ENA-NoSkill & 146 & 312 & 499.5 & 71 & 145 & 250.8 & 375.1 & \ensuremath{-}120 \\
ENA-Plasticity & 483 & 280 & 58.1 & 468 & 351 & 75.4 & 66.7 & 160 \\
ENA-HoF & 274 & 327 & 178.3 & 60 & 128 & 272.9 & 225.6 & \ensuremath{-}61 \\
ENA-HoF+EnvDetection & 272 & 404 & 249.0 & 159 & 273 & 236.3 & 242.6 & \ensuremath{-}123 \\
ENA-Plasticity+HoF & 473 & 475 & 100.5 & 468 & 278 & 58.8 & 79.6 & 94 \\
ENA-Combined & 469 & 476 & 101.5 & 478 & 465 & 97.4 & 99.5 & 3 \\
\hline
\end{tabular}
\end{adjustbox}
\label{tab:ablation_retention}
\end{center}
\end{table*}
 
\subsection{Compute Cost of Plasticity}
\label{sec:compute}
Variants with Plasticity train in \Res{ablation/ENA-Plasticity/time/min/mean}--\Res{ablation/ENA-Combined/time/min/mean} minutes per replicate, far longer than those without it, and they spend far more environment steps evaluating candidates (Section~\ref{sec:ablation}).

ENA-Combined was evaluated once with the full controller ($\mathcal{P}=1$, Section~\ref{sec:controllers}) on the first replicate of the ablation. This execution was compared with the same replicate under the linear ramp (ENA-Combined) and under the zero controller (ENA-HoF+EnvDetection). During training, the three runs spent \Res{compute/ENA-Combined-PlasticityFull/compute/candsteps/mean}, \Res{compute/ENA-Combined/compute/candsteps/mean} and \Res{compute/ENA-HoF+EnvDetection/compute/candsteps/mean} environment steps evaluating candidates. Full search costs \Res{compute/ratio/full-ramp} times the linear ramp, which costs \Res{compute/ratio/ramp-zero} times the minimum. At full effort, every generation evaluates every other specialist for up to $h_{\max}$ steps, so its cost is fixed by construction. The linear ramp's cost varies across replicates, with a standard deviation of \Res{ablation/ENA-Combined/compute/candsteps/std} steps over the ablation's \Res{ablation/n}. The ratios above are therefore those of this replicate.

In this replicate, test reliability was \Res{compute/ENA-Combined-PlasticityFull/test/rel/mean}\% at full effort, \Res{compute/ENA-Combined/test/rel/mean}\% with the ramp, and \Res{compute/ENA-HoF+EnvDetection/test/rel/mean}\% at minimum search. Most of the gap between full effort and the ramp comes from Brown-dwarf, the extrapolation environment, with mean returns of \Res{compute/ENA-Combined-PlasticityFull/test/Brown-dwarf/mean} and \Res{compute/ENA-Combined/test/Brown-dwarf/mean}. A single replicate supports no test. Among the ramp's \Res{compute/rank/of} ablation replicates, \Res{compute/rank/atleast} reached at least the reliability of full search, and this replicate is one of the ramp's weaker runs. Whether the ramp matches the reliability of full search therefore remains open.

\subsection{Shape of the Plasticity Controller}
\label{sec:controller}
The linear ramp controller~\eqref{eq:ramp} was compared with the two alternatives from Section~\ref{sec:controllers}: the quadratic and the fuzzy controller. Each alternative replaced only ENA-Combined's Plasticity controller, and ran for \Res{controller/n} replicates with the ablation's seeds. Every agent of a replicate starts from the same random state. The ablation's ENA-Combined is therefore the linear-ramp run of the same replicates, and serves as the reference without being re-run.

The decision rule was fixed before the runs. An alternative replaces the ramp if its test reliability is significantly higher, or if it is not significantly lower and uses significantly fewer candidate-evaluation steps. For each metric, the two comparisons with the ramp form one Holm family. At \Res{controller/n} replicates, each family has \Res{stats/power}\% power for $d_z \ge \Res{controller/mde}$. Table~\ref{tab:controller} gives the comparisons.

The ramp achieved a test reliability of \Res{ablation/ENA-Combined/test/rel/mean}\% (\Res{ablation/ENA-Combined/test/rel/std}), with \Res{ablation/ENA-Combined/compute/candsteps/mean} candidate steps per replicate. The fuzzy controller achieved \Res{controller/ENA-Combined-FuzzyPlasticity/test/rel/mean}\% (\Res{controller/ENA-Combined-FuzzyPlasticity/test/rel/std}), which is not distinguishable from the ramp's, and it spent significantly more candidate steps. The quadratic controller achieved \Res{controller/ENA-Combined-QuadraticPlasticity/test/rel/mean}\% (\Res{controller/ENA-Combined-QuadraticPlasticity/test/rel/std}) with fewer candidate steps, but neither difference is significant. Its effect on reliability ($d_z = \Res{controller/ENA-Combined-vs-ENA-Combined-QuadraticPlasticity/test/rel/d}$) is below the detectable one. Resolving it with \Res{stats/power}\% power would take about \Res{controller/ENA-Combined-vs-ENA-Combined-QuadraticPlasticity/test/rel/nreq} replicates per controller.

The rule's second clause reads a loss that is not significant as no loss, which is not evidence of equivalence. It did not decide this outcome, since neither alternative used significantly fewer candidate steps. A confirmatory test of a cheaper controller would need a non-inferiority margin fixed in advance. For the quadratic controller, the \Res{stats/ci}\% interval of the reliability difference runs from \Res{controller/ENA-Combined-vs-ENA-Combined-QuadraticPlasticity/test/rel/ci/low} to \Res{controller/ENA-Combined-vs-ENA-Combined-QuadraticPlasticity/test/rel/ci/high} points.

By this criterion, the controller retained is the \Res{controller/verdict}. As an exploratory observation outside the planned comparisons, the quadratic controller's reliability varied less across replicates, with a standard deviation of \Res{controller/ENA-Combined-QuadraticPlasticity/test/rel/std} against \Res{ablation/ENA-Combined/test/rel/std}.

\begin{table}[htbp]
\caption{Plasticity Controllers Against the Linear Ramp (30 Replicates)}
\begin{center}
\footnotesize
\setlength{\tabcolsep}{3pt}
\begin{adjustbox}{max width=\columnwidth}
\begin{tabular}{|l|l|c|c|c|c|}
\hline
\textbf{Controller} & \textbf{Metric} & \textbf{Difference} & \textbf{95\% CI} & \textbf{$d_z$} & \textbf{Holm} \\
\hline
Quadratic & Reliability (points) & +6.0 & \ensuremath{-}0.3 to 12.4 & 0.35 & $p=0.124$ \\
Quadratic & Candidate steps & \ensuremath{-}294,127 & \ensuremath{-}596,865 to 8,612 & \ensuremath{-}0.36 & $p=0.056$ \\
Fuzzy & Reliability (points) & +0.7 & \ensuremath{-}6.5 to 7.9 & 0.03 & $p=0.851$ \\
Fuzzy & Candidate steps & +481,852 & 210,827 to 752,878 & 0.66 & $p=0.002$ \\
\hline
\end{tabular}
\end{adjustbox}
\label{tab:controller}
\end{center}
\end{table}
 
\subsection{Selection Ablation}
\label{sec:selection_ablation}
ENA-Combined deploys its archived specialists at test with Bayesian policy reuse (Section~\ref{sec:selection}). Two simpler drivers replaced only that choice. \textit{Niche match} plays the highest-Trust gladiator of the niche the detector currently recognizes. When that niche has no gladiator, it plays the highest-Trust gladiator overall. It ignores the test returns. \textit{Best Trust} always plays the highest-Trust gladiator overall, ignoring both the detector and the returns. Both drivers rank the gladiators by their stored Trust (Section~\ref{sec:hof}).

Training selection is the same for every driver. Each alternative therefore ran for \Res{selection/n} replicates with the ablation's seeds, and the ablation's ENA-Combined is the Bayesian reference without being re-run. In \Res{selection/identity/identical} of the \Res{selection/identity/checked} alternative runs, the training log is byte-identical to the reference's. Two comparisons, fixed before the runs, form one Holm family: Bayesian policy reuse against niche match, and niche match against best Trust. At \Res{selection/n} replicates, the family has \Res{stats/power}\% power for $d_z \ge \Res{selection/mde}$.

Table~\ref{tab:selection_test} reports the frozen test. Test reliability is \Res{selection/ENA-Combined/test/rel/mean}\% (\Res{selection/ENA-Combined/test/rel/std}) with Bayesian policy reuse, \Res{selection/ENA-Combined-NicheMatch/test/rel/mean}\% (\Res{selection/ENA-Combined-NicheMatch/test/rel/std}) with niche match, and \Res{selection/ENA-Combined-BestTrust/test/rel/mean}\% (\Res{selection/ENA-Combined-BestTrust/test/rel/std}) with best Trust. Each difference is given with its \Res{stats/ci}\% interval and Holm-corrected $p$-value:
\begin{itemize}
\item \textit{Niche match against Bayesian policy reuse}: \Res{selection/ENA-Combined-vs-ENA-Combined-NicheMatch/test/rel/diff} points (\Res{selection/ENA-Combined-vs-ENA-Combined-NicheMatch/test/rel/ci/low} to \Res{selection/ENA-Combined-vs-ENA-Combined-NicheMatch/test/rel/ci/high}; $d_z = \Res{selection/ENA-Combined-vs-ENA-Combined-NicheMatch/test/rel/d}$; \Res{selection/ENA-Combined-vs-ENA-Combined-NicheMatch/test/rel/pholm}). Niche match is lower in \Res{selection/ENA-Combined-vs-ENA-Combined-NicheMatch/test/rel/lower} of the \Res{selection/ENA-Combined-vs-ENA-Combined-NicheMatch/test/rel/pairs} replicates.
\item \textit{Best Trust against niche match}: \Res{selection/ENA-Combined-NicheMatch-vs-ENA-Combined-BestTrust/test/rel/diff} points (\Res{selection/ENA-Combined-NicheMatch-vs-ENA-Combined-BestTrust/test/rel/ci/low} to \Res{selection/ENA-Combined-NicheMatch-vs-ENA-Combined-BestTrust/test/rel/ci/high}; $d_z = \Res{selection/ENA-Combined-NicheMatch-vs-ENA-Combined-BestTrust/test/rel/d}$; \Res{selection/ENA-Combined-NicheMatch-vs-ENA-Combined-BestTrust/test/rel/pholm}). Best Trust is lower in \Res{selection/ENA-Combined-NicheMatch-vs-ENA-Combined-BestTrust/test/rel/lower} replicates and higher in \Res{selection/ENA-Combined-NicheMatch-vs-ENA-Combined-BestTrust/test/rel/higher}.
\end{itemize}
Over a test, Bayesian policy reuse plays \Res{selection/ENA-Combined/played/mean} distinct individuals on average, niche match \Res{selection/ENA-Combined-NicheMatch/played/mean}, and best Trust \Res{selection/ENA-Combined-BestTrust/played/mean}. On Jupiter, the training environment left behind, niche match's mean return is \Res{selection/ENA-Combined-NicheMatch/test/Jupiter-Heavy/mean} and best Trust's is \Res{selection/ENA-Combined-BestTrust/test/Jupiter-Heavy/mean} (\Res{selection/ENA-Combined-NicheMatch-vs-ENA-Combined-BestTrust/test/Jupiter-Heavy/diff}; \Res{selection/ENA-Combined-NicheMatch-vs-ENA-Combined-BestTrust/test/Jupiter-Heavy/ci/low} to \Res{selection/ENA-Combined-NicheMatch-vs-ENA-Combined-BestTrust/test/Jupiter-Heavy/ci/high}; \Res{selection/ENA-Combined-NicheMatch-vs-ENA-Combined-BestTrust/test/Jupiter-Heavy/p}, uncorrected). On every other environment, the interval of their difference includes zero.

\textit{Stored Trust at test.} In each replicate, Bayesian policy reuse and niche match play the same archive on the same seeded test episodes. Niche match plays the gladiator of highest stored Trust in a niche. In \Res{selection/audit/phases} of the \Res{selection/audit/of} test phases in the training environments, Bayesian policy reuse mostly played another gladiator of that same niche. In those phases, it reached the success threshold in \Res{selection/audit/bpr}\% of episodes, against \Res{selection/audit/nm}\% for niche match. It did better in \Res{selection/audit/bprhigher} of these phases, and worse in \Res{selection/audit/nmhigher}. The stored Trust ranking thus does not always order a niche's gladiators as their test returns do. This comparison is descriptive.

\textit{Archive without Bayesian policy reuse.} ENA-Plasticity maintains no archive, and plays a random member of its population at test (Section~\ref{sec:variants}). It was evaluated with the same seeds, allowing a comparison with the two simpler drivers. Its test reliability is \Res{selection/ENA-Plasticity/test/rel/mean}\%. Against it, niche match differs by \Res{selection/ENA-Plasticity-vs-ENA-Combined-NicheMatch/test/rel/diff} points (\Res{selection/ENA-Plasticity-vs-ENA-Combined-NicheMatch/test/rel/ci/low} to \Res{selection/ENA-Plasticity-vs-ENA-Combined-NicheMatch/test/rel/ci/high}; \Res{selection/ENA-Plasticity-vs-ENA-Combined-NicheMatch/test/rel/p}, uncorrected), and best Trust by \Res{selection/ENA-Plasticity-vs-ENA-Combined-BestTrust/test/rel/diff} points (\Res{selection/ENA-Plasticity-vs-ENA-Combined-BestTrust/test/rel/ci/low} to \Res{selection/ENA-Plasticity-vs-ENA-Combined-BestTrust/test/rel/ci/high}; \Res{selection/ENA-Plasticity-vs-ENA-Combined-BestTrust/test/rel/p}, uncorrected). With either driver, the archive does not provide greater reliability than operating without an archive. These comparisons are exploratory. The agents also differ during training, so they do not isolate the archive.

\begin{table}[htbp]
\caption{Selection Ablation: Frozen Test (30 Replicates)}
\begin{center}
\footnotesize
\setlength{\tabcolsep}{3pt}
\begin{adjustbox}{max width=\columnwidth}
\begin{tabular}{|l|c|c|c|c|c|c|c|}
\hline
\textbf{Agent} & \textbf{Mars} & \textbf{Neptune} & \textbf{Jupiter} & \textbf{Brown} & \textbf{Earth} & \textbf{Rel. (\%)} & \textbf{Train rel. (\%)} \\
\hline
ENA-Combined & 475.5 & 490.4 & 464.9 & 367.2 & 490.5 & 83.2 $\pm$ 16.1 & 87.5 \\
ENA-Combined-NicheMatch & 396.5 & 384.8 & 407.4 & 235.6 & 397.1 & 59.2 $\pm$ 25.9 & 67.7 \\
ENA-Combined-BestTrust & 403.2 & 388.3 & 206.8 & 239.7 & 390.4 & 53.8 $\pm$ 32.3 & 55.9 \\
\hline
\end{tabular}
\end{adjustbox}
\label{tab:selection_test}
\end{center}
\end{table}
 
\subsection{Forgetting in Single-Policy Learners: CartPole}
\label{sec:forgetting_cartpole}
Tables~\ref{tab:challenge_test} and~\ref{tab:challenge_adapt} and Fig.~\ref{fig:challenge_reliability} report \Res{challenge/n} replicates, under the equal budget of \Res{cfg/challenge/budget} environment steps per environment. Appendix~\ref{app:dynamics} shows the episode-by-episode test curves across all five environments (Fig.~\ref{fig:challenge_curves}).

\subsubsection{Forgetting}
The single-policy learners end training on Earth. On Jupiter, the environment they trained on just before it, their training return at the end of the phase is \Res{challenge/PPO/end/Jupiter-Heavy/mean} (PPO), \Res{challenge/SAC-Discrete/end/Jupiter-Heavy/mean} (SAC-Discrete) and \Res{challenge/DQN/end/Jupiter-Heavy/mean} (DQN). At test, they score \Res{challenge/PPO/test/Jupiter-Heavy/mean}, \Res{challenge/SAC-Discrete/test/Jupiter-Heavy/mean} and \Res{challenge/DQN/test/Jupiter-Heavy/mean} there. They retain \Res{challenge/PPO/retention/Jupiter-Heavy/mean}\%, \Res{challenge/SAC-Discrete/retention/Jupiter-Heavy/mean}\% and \Res{challenge/DQN/retention/Jupiter-Heavy/mean}\% of their return, an absolute forgetting of \Res{challenge/PPO/forgetting/Jupiter-Heavy/mean}, \Res{challenge/SAC-Discrete/forgetting/Jupiter-Heavy/mean} and \Res{challenge/DQN/forgetting/Jupiter-Heavy/mean}. ENA-Combined ends Jupiter at \Res{challenge/ENA-Combined/end/Jupiter-Heavy/mean}, and scores \Res{challenge/ENA-Combined/test/Jupiter-Heavy/mean} there at test (\Res{challenge/ENA-Combined/retention/Jupiter-Heavy/mean}\%; forgetting \Res{challenge/ENA-Combined/forgetting/Jupiter-Heavy/mean}). At test, it recognizes Jupiter's niche in \Res{challenge/ENA-Combined/recog/Jupiter-Heavy/Jupiter-Heavy} of the \Res{challenge/n} replicates, and mostly plays Jupiter's gladiators in \Res{challenge/ENA-Combined/played/Jupiter-Heavy/Jupiter-Heavy}. On Mars, the first training environment, every agent preserves its end-of-phase return: PPO \Res{challenge/PPO/retention/Mars-Light/mean}\%, SAC-Discrete \Res{challenge/SAC-Discrete/retention/Mars-Light/mean}\%, DQN \Res{challenge/DQN/retention/Mars-Light/mean}\% and ENA-Combined \Res{challenge/ENA-Combined/retention/Mars-Light/mean}\%. Over the three training environments alone, test reliability is \Res{challenge/ENA-Combined/test/trainrel/mean}\% for ENA-Combined, against \Res{challenge/SAC-Discrete/test/trainrel/mean}\% for SAC-Discrete, \Res{challenge/PPO/test/trainrel/mean}\% for PPO and \Res{challenge/DQN/test/trainrel/mean}\% for DQN (Table~\ref{tab:challenge_test}). These values are descriptive.

\subsubsection{Reliability}
\begin{figure}[htbp]
\centerline{\includegraphics[width=\linewidth]{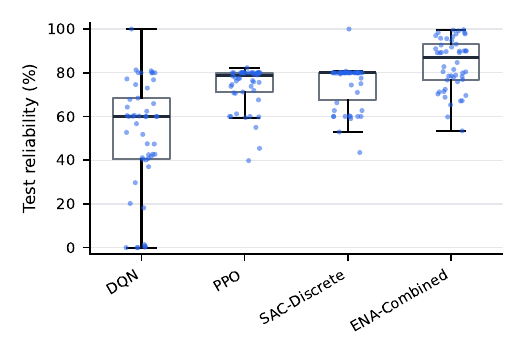}}
\caption{Test reliability on CartPole across \Res{challenge/n} replicates, under equal budgets of environment steps; each dot is one replicate.}
\label{fig:challenge_reliability}
\end{figure}

ENA-Combined's test reliability is \Res{challenge/ENA-Combined/test/rel/mean}\% (\Res{challenge/ENA-Combined/test/rel/std}), against \Res{challenge/SAC-Discrete/test/rel/mean}\% for SAC-Discrete, \Res{challenge/PPO/test/rel/mean}\% for PPO and \Res{challenge/DQN/test/rel/mean}\% for DQN (Fig.~\ref{fig:challenge_reliability}). Each of the three differences is significant after Holm correction over the \Res{challenge/family} comparisons:
\begin{itemize}
\item \textit{SAC-Discrete:} \Res{challenge/SAC-Discrete-vs-ENA-Combined/test/rel/diff} points (\Res{stats/ci}\% CI \Res{challenge/SAC-Discrete-vs-ENA-Combined/test/rel/ci/low} to \Res{challenge/SAC-Discrete-vs-ENA-Combined/test/rel/ci/high}; $d_z = \Res{challenge/SAC-Discrete-vs-ENA-Combined/test/rel/d}$; Holm \Res{challenge/SAC-Discrete-vs-ENA-Combined/test/rel/pholm}).
\item \textit{PPO:} \Res{challenge/PPO-vs-ENA-Combined/test/rel/diff} points (\Res{stats/ci}\% CI \Res{challenge/PPO-vs-ENA-Combined/test/rel/ci/low} to \Res{challenge/PPO-vs-ENA-Combined/test/rel/ci/high}; $d_z = \Res{challenge/PPO-vs-ENA-Combined/test/rel/d}$; Holm \Res{challenge/PPO-vs-ENA-Combined/test/rel/pholm}).
\item \textit{DQN:} \Res{challenge/DQN-vs-ENA-Combined/test/rel/diff} points (\Res{stats/ci}\% CI \Res{challenge/DQN-vs-ENA-Combined/test/rel/ci/low} to \Res{challenge/DQN-vs-ENA-Combined/test/rel/ci/high}; $d_z = \Res{challenge/DQN-vs-ENA-Combined/test/rel/d}$; Holm \Res{challenge/DQN-vs-ENA-Combined/test/rel/pholm}).
\end{itemize}
On the environments other than Jupiter, SAC-Discrete's test return is slightly higher than ENA-Combined's. On Mars, it is \Res{challenge/SAC-Discrete/test/Mars-Light/mean} against \Res{challenge/ENA-Combined/test/Mars-Light/mean} (\Res{challenge/SAC-Discrete-vs-ENA-Combined/test/Mars-Light/p}). On Earth, it is \Res{challenge/SAC-Discrete/test/Earth-Standard/mean} against \Res{challenge/ENA-Combined/test/Earth-Standard/mean} (\Res{challenge/SAC-Discrete-vs-ENA-Combined/test/Earth-Standard/p}), and on the unseen Neptune \Res{challenge/SAC-Discrete/test/Neptune/mean} against \Res{challenge/ENA-Combined/test/Neptune/mean} (\Res{challenge/SAC-Discrete-vs-ENA-Combined/test/Neptune/p}). On Brown-dwarf, the other unseen environment, ENA-Combined scores \Res{challenge/ENA-Combined/test/Brown-dwarf/mean}, against \Res{challenge/PPO/test/Brown-dwarf/mean} for PPO and \Res{challenge/SAC-Discrete/test/Brown-dwarf/mean} for SAC-Discrete (\Res{challenge/PPO-vs-ENA-Combined/test/Brown-dwarf/p} and \Res{challenge/SAC-Discrete-vs-ENA-Combined/test/Brown-dwarf/p}). These per-environment comparisons are uncorrected and descriptive.

\subsubsection{Adaptation}
Table~\ref{tab:challenge_adapt} gives the median steps to first solve, with their interquartile range. ENA-Combined never solved Jupiter in \Res{challenge/ENA-Combined/adapt/Jupiter-Heavy:steps/unsolved} of the \Res{challenge/n} replicates. PPO's median first solve takes \Res{challenge/PPO/adapt/Mars-Light:steps/median}, \Res{challenge/PPO/adapt/Jupiter-Heavy:steps/median} and \Res{challenge/PPO/adapt/Earth-Standard:steps/median} steps on the three training environments. ENA-Combined's takes \Res{challenge/ENA-Combined/adapt/Mars-Light:steps/median}, \Res{challenge/ENA-Combined/adapt/Jupiter-Heavy:steps/median} and \Res{challenge/ENA-Combined/adapt/Earth-Standard:steps/median}, later on all three (\Res{challenge/PPO-vs-ENA-Combined/adapt/Mars-Light:steps/p}, \Res{challenge/PPO-vs-ENA-Combined/adapt/Jupiter-Heavy:steps/p} and \Res{challenge/PPO-vs-ENA-Combined/adapt/Earth-Standard:steps/p}). Against the other two baselines:
\begin{itemize}
\item \textit{Mars:} ENA-Combined solves it sooner than DQN (\Res{challenge/DQN/adapt/Mars-Light:steps/median} steps; \Res{challenge/DQN-vs-ENA-Combined/adapt/Mars-Light:steps/p}). Its difference from SAC-Discrete (\Res{challenge/SAC-Discrete/adapt/Mars-Light:steps/median}; \Res{challenge/SAC-Discrete-vs-ENA-Combined/adapt/Mars-Light:steps/p}) is not significant.
\item \textit{Jupiter:} its first solve does not differ significantly from SAC-Discrete's (\Res{challenge/SAC-Discrete/adapt/Jupiter-Heavy:steps/median}; \Res{challenge/SAC-Discrete-vs-ENA-Combined/adapt/Jupiter-Heavy:steps/p}) or DQN's (\Res{challenge/DQN/adapt/Jupiter-Heavy:steps/median}; \Res{challenge/DQN-vs-ENA-Combined/adapt/Jupiter-Heavy:steps/p}).
\item \textit{Earth:} later than both (\Res{challenge/DQN/adapt/Earth-Standard:steps/median} and \Res{challenge/SAC-Discrete/adapt/Earth-Standard:steps/median}; \Res{challenge/DQN-vs-ENA-Combined/adapt/Earth-Standard:steps/p} and \Res{challenge/SAC-Discrete-vs-ENA-Combined/adapt/Earth-Standard:steps/p}).
\end{itemize}

\subsubsection{Compute}
With the same number of environment steps, ENA-Combined trained in \Res{challenge/ENA-Combined/time/min/mean} minutes of wall-clock time per replicate. DQN took \Res{challenge/DQN/time/min/mean} minutes, SAC-Discrete \Res{challenge/SAC-Discrete/time/min/mean} and PPO \Res{challenge/PPO/time/min/mean}, on the same CPU and with implementations in the same framework. ENA computes no gradients: its cost is forward passes and environment steps.

\begin{table}[htbp]
\caption{CartPole Comparison: Frozen Test (50 Replicates)}
\begin{center}
\footnotesize
\setlength{\tabcolsep}{3pt}
\begin{adjustbox}{max width=\columnwidth}
\begin{tabular}{|l|c|c|c|c|c|c|c|}
\hline
\textbf{Agent} & \textbf{Mars} & \textbf{Neptune} & \textbf{Jupiter} & \textbf{Brown} & \textbf{Earth} & \textbf{Rel. (\%)} & \textbf{Train rel. (\%)} \\
\hline
DQN & 401.6 & 443.7 & 75.2 & 238.2 & 442.0 & 50.6 $\pm$ 26.4 & 45.9 \\
PPO & 488.0 & 497.9 & 21.6 & 379.5 & 498.0 & 73.8 $\pm$ 9.6 & 65.5 \\
SAC-Discrete & 495.7 & 500.0 & 52.1 & 393.5 & 500.0 & 74.5 $\pm$ 10.2 & 66.3 \\
ENA-Combined & 480.2 & 492.9 & 461.3 & 381.6 & 498.1 & 83.9 $\pm$ 11.7 & 88.6 \\
\hline
\end{tabular}
\end{adjustbox}
\label{tab:challenge_test}
\end{center}
\end{table}
\begin{table}[htbp]
\caption{CartPole Comparison: Training Steps to First Solve, Median (IQR) (50 Replicates)}
\begin{center}
\footnotesize
\setlength{\tabcolsep}{3pt}
\begin{adjustbox}{max width=\columnwidth}
\begin{tabular}{|l|c|c|c|}
\hline
\textbf{Agent} & \textbf{Mars} & \textbf{Jupiter} & \textbf{Earth} \\
\hline
DQN & \begin{tabular}{@{}c@{}}257,626\\(201,536--295,276)\end{tabular} & \begin{tabular}{@{}c@{}}122,843\\(100,210--173,903)\end{tabular} & \begin{tabular}{@{}c@{}}5,476\\(5,000--30,069)\end{tabular} \\
PPO & \begin{tabular}{@{}c@{}}10,018\\(6,282--10,046)\end{tabular} & \begin{tabular}{@{}c@{}}47,743\\(35,289--75,043)\end{tabular} & \begin{tabular}{@{}c@{}}5,000\\(5,000--5,000)\end{tabular} \\
SAC-Discrete & \begin{tabular}{@{}c@{}}127,650\\(25,366--184,064)\end{tabular} & \begin{tabular}{@{}c@{}}162,720\\(146,504--175,217)\end{tabular} & \begin{tabular}{@{}c@{}}5,000\\(5,000--5,000)\end{tabular} \\
ENA-Combined & \begin{tabular}{@{}c@{}}44,841\\(20,205--69,308)\end{tabular} & \begin{tabular}{@{}c@{}}106,116\\(36,117--204,861)\end{tabular} & \begin{tabular}{@{}c@{}}39,288\\(26,760--55,594)\end{tabular} \\
\hline
\end{tabular}
\end{adjustbox}
\label{tab:challenge_adapt}
\end{center}
\end{table}
 
\subsection{Environment Detection on HalfCheetah}
\label{sec:detection_benchmark}
The detector's inputs are the agent's own, so detection can be evaluated without training. The detection benchmark plays scripted episodes in Earth, Heavy-Rough and Mars-Slippery, \Res{cfg/detbench/episodes} in each. Every \Res{cfg/detbench/block} episodes, it changes the acting policy among near-still, uniformly random and sinusoidal gaits. The same family of gaits is used in every environment, so that only the physics differ. The benchmark feeds the detector exactly the calls ENA makes, with a niche assignment after every episode. A run meets the target when three conditions hold. It confirms both switches within \Res{cfg/detbench/window} episodes. It raises no false alarm inside an environment. It gives each environment its own niche, covering at least \Res{cfg/detbench/purity}\,\% of that environment's episodes.

Each detector was evaluated on \Res{cfg/detbench/confirmseeds} independent benchmark runs. Each run has its own random seed, which draws different policies and resets, so each run is identified by its seed. Intervals are exact (Clopper--Pearson) at \Res{stats/ci}\,\%. Comparisons are exact McNemar tests on seed-paired outcomes, Holm-adjusted as one family. As a check of the benchmark's criteria, CartPole's transition-model detector meets the target on \Res{detbench/cartpole/passes} of \Res{detbench/cartpole/seeds} runs of the corresponding CartPole benchmark.

On HalfCheetah, the transition-model detector is calibrated for the domain: margin $m = \Res{cfg/detbench/transition/noveltymargin}$, at least \Res{cfg/detbench/transition/minsamples} transitions to fit a niche, and \Res{cfg/detbench/transition/minverifysamples} to verify an alarm. It meets the target on \Res{detbench/transition/passes} of \Res{cfg/detbench/confirmseeds} runs. On average, it confirms \Res{detbench/transition/changes} changes, \Res{detbench/transition/inside} of them inside an environment, and creates \Res{detbench/transition/niches} niches. Only \Res{detbench/transition/purity}\,\% of the worst environment's episodes fall in its niche. With one fixed gait in every environment, the same detector still confirms \Res{detbench/transitionfixed/changes} changes and creates \Res{detbench/transitionfixed/niches} niches on average. It also merges two environments into one niche on \Res{detbench/transitionfixed/merged} of the \Res{cfg/detbench/confirmseeds} seeds.

Table~\ref{tab:chirp_ablation} ablates the chirp's excitation and response features over \Res{detbench/seeds} benchmark seeds. Each cell counts the seeds on which that pair meets the target, and ENA's chirp is the best of them. The Mexican-hat wavelet train is the best excitation. It meets the target on at least \Res{detbench/chosenexcitation/minpasses} of the \Res{detbench/seeds} seeds with every response feature, and on every seed with \Res{detbench/passing} of them. No linear frequency sweep, sine or impulse meets the target on any seed. The best configuration without a wavelet train, an \Res{detbench/bestother/excitation} with \Res{detbench/bestother/features} features, meets it on \Res{detbench/bestother/passes} seeds. The Haar and db4~\cite{daubechies1988orthonormal} trains depend strongly on the features. ENA uses the Mexican-hat train with Haar wavelet packets (in bold), which meets the target on every seed.

On the confirmation runs, this chirp meets the target on \Res{detbench/chosen/passes} of \Res{cfg/detbench/confirmseeds} seeds (\Res{stats/ci}\,\% interval \Res{detbench/chosen/lower}--\Res{detbench/chosen/upper}\,\%). It confirms a false change on \Res{detbench/chosen/falseseeds} of them, and confirms every switch at the end of the first episode after it. The share of the worst environment's episodes that fall in its own niche averages \Res{detbench/chosen/puritymean}\,\% over the seeds, and is never below \Res{detbench/chosen/minpurity}\,\%. The chirp meets the target more often than the transition-model detector (\Res{detbench/cmp/chirp/p}). On the \Res{detbench/seeds} ablation seeds it was chosen on, it also meets the target more often than the best configuration without a wavelet train (\Res{detbench/cmp/other/p}).

In the HalfCheetah study itself (Section~\ref{sec:cheetah}), the chirp gives each training environment its own niche in \Res{cheetah/ENA-Combined/recog/distinct} of the \Res{cheetah/n} replicates (Table~\ref{tab:cheetah_detection}, Appendix~\ref{sec:appendix_detection}). A training run, with two switches, confirms \Res{cheetah/ENA-Combined/detect/train/changes/total} changes on average. A test, with four switches, confirms \Res{cheetah/ENA-Combined/detect/test/changes/total}. At test, the training environments are recognized as their own niches in most replicates: Earth-Standard in \Res{cheetah/ENA-Combined/recog/Earth-Standard/Earth-Standard}, Heavy-Rough in \Res{cheetah/ENA-Combined/recog/Heavy-Rough/Heavy-Rough} and Mars-Slippery in \Res{cheetah/ENA-Combined/recog/Mars-Slippery/Mars-Slippery}. High-Wind is recognized as Earth-Standard's niche in \Res{cheetah/ENA-Combined/recog/High-Wind/Earth-Standard} replicates. Ice-Floor is left unrecognized in \Res{cheetah/ENA-Combined/detect/test/Ice-Floor/unrec/mean}\,\% of its episodes.

\begin{table}[htbp]
\caption{Chirp Detector Ablation (30 Seeds)}
\begin{center}
\footnotesize
\setlength{\tabcolsep}{3pt}
\begin{adjustbox}{max width=\columnwidth}
\begin{tabular}{|l|c|c|c|c|c|}
\hline
\textbf{Excitation} & \textbf{Raw} & \textbf{DWT Haar} & \textbf{DWT db4} & \textbf{SWT Haar} & \textbf{Packets Haar} \\
\hline
Linear sweep & 0/30 & 0/30 & 0/30 & 0/30 & 0/30 \\
Exponential sweep & 0/30 & 0/30 & 0/30 & 4/30 & 1/30 \\
Sine & 0/30 & 0/30 & 0/30 & 0/30 & 0/30 \\
Impulse & 0/30 & 0/30 & 0/30 & 0/30 & 0/30 \\
Wavelet train (Haar) & 26/30 & 12/30 & 28/30 & 2/30 & 14/30 \\
Wavelet train (db4) & 0/30 & 12/30 & 17/30 & 13/30 & 26/30 \\
Wavelet train (Mexican hat) & 30/30 & 29/30 & 30/30 & 29/30 & \textbf{30/30} \\
\hline
\end{tabular}
\end{adjustbox}
\label{tab:chirp_ablation}
\end{center}
\end{table}
 
\subsection{Forgetting in Single-Policy Learners: HalfCheetah}
\label{sec:cheetah}
\begin{figure}[htbp]
\centerline{\includegraphics[width=\linewidth]{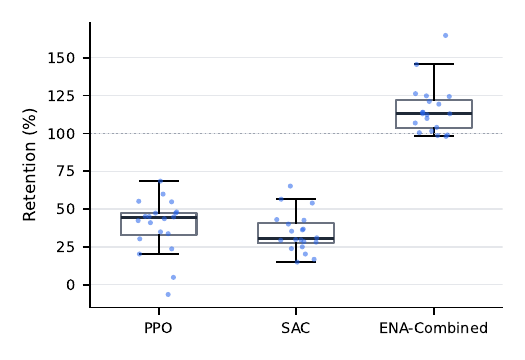}}
\caption{Left-behind retention on HalfCheetah (\Res{cheetah/n} replicates); each dot is one replicate. The dotted line marks 100\%.}
\label{fig:cheetah_retention}
\end{figure}

Tables~\ref{tab:cheetah_retention} and~\ref{tab:cheetah_test} and Fig.~\ref{fig:cheetah_retention} report \Res{cheetah/n} replicates. On the non-stationary stream, ENA-Combined reaches the \Res{cfg/cheetah/solved} threshold in \Res{cheetah/ENA-Combined/test/rel/mean}\% of its test episodes, PPO in \Res{cheetah/PPO/test/rel/mean}\%, and SAC in \Res{cheetah/SAC/test/rel/mean}\%. As explained in Section~\ref{sec:setup}, retention and absolute forgetting are the primary metrics here. Appendix~\ref{app:dynamics} shows the episode-by-episode test curves and the training curves (Figs.~\ref{fig:cheetah_test_curves} and~\ref{fig:cheetah_training_curves}).

\begin{table*}[t]
\caption{HalfCheetah Comparison: Retention of the Environments Left Behind (20 Replicates)}
\begin{center}
\footnotesize
\setlength{\tabcolsep}{3pt}
\begin{adjustbox}{max width=\textwidth}
\begin{tabular}{|l|c|c|c|c|c|c|c|c|c|}
\hline
\textbf{Agent} & \textbf{Earth end} & \textbf{Earth test} & \textbf{Earth ret. (\%)} & \textbf{Heavy end} & \textbf{Heavy test} & \textbf{Heavy ret. (\%)} & \textbf{Retention (\%)} & \textbf{Forgetting} & \textbf{Time (min)} \\
\hline
PPO & 4,543 & 2,208 & 50.1 & 4,793 & 1,302 & 28.3 & 39.2 & 2,913 & 115.3 \\
SAC & 8,635 & 4,353 & 50.8 & 7,544 & 1,301 & 18.0 & 34.4 & 5,263 & 281.1 \\
ENA-Combined & 1,805 & 2,000 & 116.3 & 2,133 & 2,417 & 114.9 & 115.6 & \ensuremath{-}239 & 32.0 \\
\hline
\end{tabular}
\end{adjustbox}
\label{tab:cheetah_retention}
\end{center}
\end{table*}
\begin{table}[htbp]
\caption{HalfCheetah Comparison: Frozen Test (20 Replicates)}
\begin{center}
\footnotesize
\setlength{\tabcolsep}{3pt}
\begin{adjustbox}{max width=\columnwidth}
\begin{tabular}{|l|c|c|c|c|c|}
\hline
\textbf{Agent} & \textbf{Earth} & \textbf{Ice} & \textbf{Heavy} & \textbf{High} & \textbf{Mars} \\
\hline
PPO & 2,208 $\pm$ 1,013 & 2,203 $\pm$ 1,000 & 1,302 $\pm$ 1,087 & 929 $\pm$ 908 & 2,506 $\pm$ 1,187 \\
SAC & 4,353 $\pm$ 981 & 6,774 $\pm$ 2,214 & 1,301 $\pm$ 1,218 & 2,642 $\pm$ 896 & 10,146 $\pm$ 975 \\
ENA-Combined & 2,000 $\pm$ 492 & 1,915 $\pm$ 503 & 2,417 $\pm$ 539 & 996 $\pm$ 511 & 1,773 $\pm$ 375 \\
\hline
\end{tabular}
\end{adjustbox}
\label{tab:cheetah_test}
\end{center}
\end{table}
 
\subsubsection{Forgetting}
The environments left behind are Earth-Standard, trained first, and Heavy-Rough, trained second. PPO ends Heavy-Rough at \Res{cheetah/PPO/end/Heavy-Rough/mean} and scores \Res{cheetah/PPO/test/Heavy-Rough/mean} there at test, retaining \Res{cheetah/PPO/retention/Heavy-Rough/mean}\%. SAC ends it at \Res{cheetah/SAC/end/Heavy-Rough/mean} and scores \Res{cheetah/SAC/test/Heavy-Rough/mean}, retaining \Res{cheetah/SAC/retention/Heavy-Rough/mean}\%. Averaged over both environments left behind, PPO loses \Res{cheetah/PPO/forgetting/left/mean} (\Res{cheetah/PPO/forgetting/left/std}) of its end-of-phase return, and retains \Res{cheetah/PPO/retention/left/mean}\% (\Res{cheetah/PPO/retention/left/std}). SAC loses \Res{cheetah/SAC/forgetting/left/mean} (\Res{cheetah/SAC/forgetting/left/std}), and retains \Res{cheetah/SAC/retention/left/mean}\% (\Res{cheetah/SAC/retention/left/std}).

ENA-Combined's absolute forgetting is \Res{cheetah/ENA-Combined/forgetting/left/mean} (\Res{cheetah/ENA-Combined/forgetting/left/std}), and its retention \Res{cheetah/ENA-Combined/retention/left/mean}\% (\Res{cheetah/ENA-Combined/retention/left/std}). Its test returns, \Res{cheetah/ENA-Combined/test/Earth-Standard/mean} on Earth-Standard and \Res{cheetah/ENA-Combined/test/Heavy-Rough/mean} on Heavy-Rough, exceeded its end-of-phase training returns of \Res{cheetah/ENA-Combined/end/Earth-Standard/mean} and \Res{cheetah/ENA-Combined/end/Heavy-Rough/mean}. On Heavy-Rough at test, it mostly plays Heavy-Rough's gladiators, in \Res{cheetah/ENA-Combined/played/Heavy-Rough/Heavy-Rough} of the \Res{cheetah/n} replicates.

The planned comparisons on retention, Holm-corrected over the two baselines, are:
\begin{itemize}
  \item \textit{PPO:} \Res{cheetah/PPO-vs-ENA-Combined/retention/left/diff} percentage points (\Res{stats/ci}\% CI \Res{cheetah/PPO-vs-ENA-Combined/retention/left/ci/low} to \Res{cheetah/PPO-vs-ENA-Combined/retention/left/ci/high}; $d_z = \Res{cheetah/PPO-vs-ENA-Combined/retention/left/d}$; Holm \Res{cheetah/PPO-vs-ENA-Combined/retention/left/pholm}).
  \item \textit{SAC:} \Res{cheetah/SAC-vs-ENA-Combined/retention/left/diff} percentage points (\Res{stats/ci}\% CI \Res{cheetah/SAC-vs-ENA-Combined/retention/left/ci/low} to \Res{cheetah/SAC-vs-ENA-Combined/retention/left/ci/high}; $d_z = \Res{cheetah/SAC-vs-ENA-Combined/retention/left/d}$; Holm \Res{cheetah/SAC-vs-ENA-Combined/retention/left/pholm}).
\end{itemize}
Absolute forgetting forms a Holm family defined after the runs. On it, ENA-Combined's difference is \Res{cheetah/PPO-vs-ENA-Combined/forgetting/left/diff} against PPO (\Res{cheetah/PPO-vs-ENA-Combined/forgetting/left/ci/low} to \Res{cheetah/PPO-vs-ENA-Combined/forgetting/left/ci/high}; Holm \Res{cheetah/PPO-vs-ENA-Combined/forgetting/left/pholm}), and \Res{cheetah/SAC-vs-ENA-Combined/forgetting/left/diff} against SAC (\Res{cheetah/SAC-vs-ENA-Combined/forgetting/left/ci/low} to \Res{cheetah/SAC-vs-ENA-Combined/forgetting/left/ci/high}; Holm \Res{cheetah/SAC-vs-ENA-Combined/forgetting/left/pholm}).

ENA-Combined's end-of-phase return also includes episodes played by specialists outside its archive, such as the $\epsilon$-greedy selection's picks. Its test plays only gladiators. In \Res{cheetah/archived/n} of the \Res{cheetah/n} replicates, gladiators played at the end of both phases left behind. Measured against the gladiators' end-of-phase episodes alone, ENA-Combined's retention in these replicates is \Res{cheetah/archived/ENA-Combined/retention/left/mean}\% (\Res{cheetah/archived/ENA-Combined/retention/left/std}), and its absolute forgetting \Res{cheetah/archived/ENA-Combined/forgetting/left/mean}. The single-policy learners' end-of-phase returns also include their exploration, since they act deterministically only at test. These values are descriptive.

\subsubsection{Absolute return}
SAC's test returns exceeded ENA-Combined's on every environment but Heavy-Rough. On Mars-Slippery, SAC scores \Res{cheetah/SAC/test/Mars-Slippery/mean} against \Res{cheetah/ENA-Combined/test/Mars-Slippery/mean} (\Res{cheetah/SAC-vs-ENA-Combined/test/Mars-Slippery/p}). On Ice-Floor, it scores \Res{cheetah/SAC/test/Ice-Floor/mean} against \Res{cheetah/ENA-Combined/test/Ice-Floor/mean} (\Res{cheetah/SAC-vs-ENA-Combined/test/Ice-Floor/p}), and on High-Wind \Res{cheetah/SAC/test/High-Wind/mean} against \Res{cheetah/ENA-Combined/test/High-Wind/mean} (\Res{cheetah/SAC-vs-ENA-Combined/test/High-Wind/p}). On Earth-Standard, SAC still scores \Res{cheetah/SAC/test/Earth-Standard/mean} against \Res{cheetah/ENA-Combined/test/Earth-Standard/mean} (\Res{cheetah/SAC-vs-ENA-Combined/test/Earth-Standard/p}), even after losing half of its end-of-phase return there. SAC's end-of-phase returns on the environments left behind, \Res{cheetah/SAC/end/Earth-Standard/mean} and \Res{cheetah/SAC/end/Heavy-Rough/mean}, are also several times ENA-Combined's. On Heavy-Rough at test, ENA-Combined's return is higher than SAC's (\Res{cheetah/SAC-vs-ENA-Combined/test/Heavy-Rough/diff}; \Res{cheetah/SAC-vs-ENA-Combined/test/Heavy-Rough/p}) and PPO's (\Res{cheetah/PPO-vs-ENA-Combined/test/Heavy-Rough/diff}; \Res{cheetah/PPO-vs-ENA-Combined/test/Heavy-Rough/p}). These per-environment comparisons are uncorrected and descriptive (Table~\ref{tab:cheetah_test}).

\subsubsection{Compute}
ENA-Combined trained in \Res{cheetah/ENA-Combined/time/min/mean} minutes per replicate (\Res{cheetah/ENA-Combined/time/min/std}). PPO took \Res{cheetah/PPO/time/min/mean} minutes and SAC \Res{cheetah/SAC/time/min/mean}, on the same instance type (Appendix~\ref{app:compute}). ENA's budget was \Res{cfg/cheetah/budget} training steps per environment. Over the three environments, \Res{cheetah/ENA-Combined/compute/candsteps/mean} of its training steps went to candidate evaluation, Olympiads and chirps, and its acting specialist played the rest (Section~\ref{sec:forgetting_protocol}).

\section{Discussion}
\label{sec:discussion}

\subsection{Answers to the Research Questions}

\textit{RQ1: do the skills mitigate forgetting? Supported on CartPole, for the tested sequences.} In the ablation, the variants that learn Mars and Jupiter are those with Plasticity. Among them, only ENA-Combined maintains performance in both environments at test, and it is the most reliable variant (Section~\ref{sec:ablation}). Adding the archive also changes how the test policy is chosen, so the ablation does not isolate the archive's contribution from the selection's. Without Plasticity, ENA fails to acquire competent policies on Jupiter. The reliability gains observed in the ablation therefore reflect both policy acquisition and subsequent retention. The single-policy learners degrade markedly on Jupiter, the CartPole environment trained just before the last, though they retain performance on Mars. On HalfCheetah, they keep \Res{cheetah/SAC/retention/left/mean}--\Res{cheetah/PPO/retention/left/mean}\% on average of what they reached on the two environments left behind. ENA keeps, on average, its own end-of-phase return there. That return is much lower than theirs (Sections~\ref{sec:forgetting_cartpole} and~\ref{sec:cheetah}). All of these findings hold for the specific implementations, task sequences, and budgets evaluated; alternative settings were not tested.

\textit{RQ1a: is an archive enough by itself? No.} Variants with a Hall of Fame but no detection are not more reliable than the same variants without it, and ENA-HoF is less reliable than ENA-NoSkill. Variants whose archive is organized by detected physics are more reliable than those with an archive but no detection, in both cases. With Plasticity, they are also more reliable than the variant without any archive, and only this combination maintains performance across all learned environments (Section~\ref{sec:ablation}). In this ablation, merely archiving proven specialists proved insufficient: archiving increased reliability only together with detection. Detection, however, brings three things at once: the niche index, a larger archive, and resets at each confirmed change (Section~\ref{sec:variants}). Which of them matters was not isolated. The recall rule also matters. Bayesian policy reuse is more reliable than deploying the recognized niche's highest-Trust gladiator. Niche match, in turn, is not significantly more reliable than ignoring the niche (Section~\ref{sec:selection_ablation}). With either simpler rule, the archive is not more reliable than no archive at all.

\textit{RQ1b: does the memory cost extra interaction? Partly answered.} Without Plasticity, archiving and detecting add no measurable candidate evaluation of their own. Their only measured cost is wall-clock time (Section~\ref{sec:ablation}). With Plasticity, adding the Hall of Fame comes with more candidate steps. The increase is not significant, but the data cannot exclude it, up to \Res{ablation/ENA-Plasticity-vs-ENA-Plasticity+HoF/compute/candsteps/ci/high} steps per replicate. The archive takes no extra candidate evaluation to fill, but it may change how much the failure-driven search spends.

\textit{RQ1c: can the memory be built online, without labels? Yes in both domains, with different drivers.} On CartPole, the transition-model detector separates the training environments, with the fragmentation and false alarms of Table~\ref{tab:detection}. On HalfCheetah, the same kind of detector fails (Section~\ref{sec:chirp}). The chirp, whose response does not depend on the agent, meets the detection target on the benchmark. It also gives each training environment its own niche in every replicate of the HalfCheetah study (Section~\ref{sec:detection_benchmark}). Neither driver uses a gradient.

\textit{RQ2: does failure-driven search pay? Partly supported.} The comparisons that add Plasticity show the largest differences in the ablation, and only the variants with Plasticity learn the harder environments (Section~\ref{sec:ablation}). In the ablation, however, the budget is a number of episodes, so more search also means far more interaction with the environment. Switching Plasticity off also changes the candidate fitness signal, not only its volume. At the minimum depth, a CartPole candidate that survives $h_{\min}$ steps is credited with a full episode (Section~\ref{sec:trust}). The ablation does not separate the two. Searching at full effort in every episode costs an order of magnitude more candidate evaluation (Section~\ref{sec:compute}), a cost set by construction. Whether the ramp reaches what full effort does remains open. In the single replicate run at full effort, full effort was the more reliable. Whether scaling search effort dynamically with failure outperforms distributing identical effort uniformly across iterations remains an open empirical question. The quadratic controller was more reliable than the ramp and used fewer candidate steps, but neither difference is significant (Section~\ref{sec:controller}). These findings therefore support the general principle of scaling search with failure, rather than the specific linear ramp evaluated here.

\subsection{How Detection May Help the Archive}
Detection may help the archive in three ways, and the ablation does not separate them.
\begin{itemize}
\item \textit{The index at recall.} Without niches, the archive is one pool of gladiators proven in different physics. At test time, the selection mechanism then lacks dynamical evidence to distinguish specialists evolved for Jupiter from those evolved for Earth. With detection, every gladiator carries the physics it was proven in. The dynamics likelihoods of the selection~\eqref{eq:bpr} point it to the matching niche, and the gladiators of one niche can no longer be evicted by another niche's.
\item \textit{The capacity.} With detection, the archive holds $K$ gladiators per niche instead of $K$ in all, so it ends training several times larger (Section~\ref{sec:ablation}). During training, its protected copies also hold more population slots, whatever the current physics.
\item \textit{The resets.} Each confirmed change clears the return history and the input statistics, and returns the failure signal to its maximum for the next episodes. Without detection, none of this happens.
\end{itemize}
An archive of matched capacity, with the same resets but without the niche index, would separate the index from the other two. Adding the archive also replaces a random population member by Bayesian policy reuse (Section~\ref{sec:variants}).

Selection rules that use the same niches also differ in reliability. Bayesian policy reuse differs from niche match in two ways. It weighs each episode's dynamics likelihoods into a belief, instead of following the detector's single decision. It also learns from the test returns. The selection ablation does not separate the two. One plausible mechanism, not directly isolated in the experiments, involves the feedback provided by episode returns. They let Bayesian policy reuse abandon a gladiator that fails in its own niche, which niche match cannot do, and across a test phase, niche match evaluates substantially fewer distinct individuals (Section~\ref{sec:selection_ablation}). A gladiator archived for solving a niche could still fail there at test, and without return-based evidence, an ineffective specialist would continue to be deployed. Niche match also inherits the archive's ranking. It plays a niche's gladiator of highest stored Trust, and Trust values stored at different times are not on one scale (Section~\ref{sec:hof}). In the test phases where Bayesian policy reuse settled on another gladiator of the same niche, it was mostly the more reliable (Section~\ref{sec:selection_ablation}). Part of its advantage may therefore come from correcting that ranking with the test returns. The experiments do not separate this from its use of the dynamics evidence.

\subsection{Retention, Adaptation and Absolute Return}
\label{sec:retention}
ENA preserves the behaviors it acquires. Other learners, however, adapt faster or reach higher returns, by margins that vary with the domain. Both gaps match known limits of evolutionary search. Evolutionary algorithms need many samples~\cite{khadka2018evolutionary,pourchot2019cemrl}, and struggle on problems that require optimizing many parameters~\cite{khadka2018evolutionary}. The cost of exploring in parameter space grows with the number of parameters~\cite{vemula2019contrasting}, and random mutations tend to disrupt large networks~\cite{lehman2018safe}. This is why ENA's specialists are small. On CartPole, PPO first solves every training environment sooner than ENA. Most of ENA's steps go to evaluating candidates (Section~\ref{sec:forgetting_protocol}). Against DQN and SAC-Discrete, the order depends on the environment: ENA solves Mars sooner than DQN, and Earth later than both (Section~\ref{sec:forgetting_cartpole}). SAC-Discrete scores slightly higher on the environments it does not forget. ENA, in turn, preserves Jupiter, where every single-policy learner degrades.

On HalfCheetah, the gap is larger. ENA preserves performance from earlier phases, but achieves lower asymptotic returns than SAC within the allocated step budget. The robot has six joints to control. PPO and SAC control them with hidden layers of \Res{cfg/cheetah/baselinenet} units, while ENA's specialists have \Res{cfg/cheetahena/brain} units. Evolving networks as large as the baselines' would mean searching far more parameters, at the cost described above. Even with small networks, most of ENA's steps go to evaluating candidates rather than to its acting specialist's episodes (Section~\ref{sec:forgetting_protocol}). What SAC loses on the environments left behind exceeds the return ENA reached there at the end of each phase (Section~\ref{sec:cheetah}). Most of ENA's retention above 100\% comes from its end-of-phase reference, which includes episodes of specialists outside the archive. Measured against its gladiators' episodes alone, its retention is \Res{cheetah/archived/ENA-Combined/retention/left/mean}\% (Section~\ref{sec:cheetah}).

A specialist archive is hypothesized to provide the greatest advantage when environments recur and relearning is computationally expensive, though this setting was not explicitly evaluated. When an environment is unlikely to return, or a gradient learner can relearn it quickly, a single strong policy that is partly forgotten may be worth more than a weaker one kept intact.

\subsection{Scope of the Comparison}
\label{sec:scope}
The single-policy learners measure how much standard learners forget on these streams. The claim of this paper rests on the skill ablation. The comparison has three limits. First, at test, every agent's weights are frozen, but ENA still routes among its archived specialists with an updating belief. The single-policy learners deploy one policy. The comparison thus contrasts context-indexed specialist preservation with single-policy learning, not two equally equipped systems.

Second, ENA evaluates its candidates in fresh copies of the current environment, which the experiment builds from the environment's own parameters (Appendix~\ref{app:information}). Every candidate step counts in ENA's budget. Still, the single-policy learners play only in the stream itself, and never use such a copy.

Third, the continual-RL methods closest to ENA are not baselines here, for reasons that differ between them. OWL~\cite{kessler2022same} and TeLAPA~\cite{lillo2026beyond} are told where tasks begin and end. OWL trains the head of the task it is given, and infers the task only at test. TeLAPA builds a task's archive, and retrieves a starting point from its archives, at each task boundary it is given. Both build their memory during the stream, but from labeled tasks. On the non-stationary streams evaluated here, these methods would require either explicit task boundaries---which are unavailable to any agent in this study---or an environment detector to infer them. Providing such an inferential mechanism would effectively endow those baselines with ENA's environment-detection capability. ITE~\cite{cully2015robots} builds its archive offline, before deployment. It adapts once to each damage, starting when performance drops below a user-set threshold. It neither keeps a memory across a sequence of conditions, nor recognizes a condition that returns.

RL-CD~\cite{dasilva2006dealing}, MBCD~\cite{alegre2021minimum} and SWOKS~\cite{dick2024statistical} do detect contexts online without labels, and keep a policy per context, so they address the same problem. A comparison with them, however, would be confounded. RL-CD's models are tabular and need a discretized state space, while MBCD and SWOKS learn by gradient descent. A difference could therefore come from the learning algorithm rather than from the memory. Direct comparisons with such algorithms are deferred to future investigations of hybrid architectures combining ENA's modular skills with gradient-based optimization (Section~\ref{sec:conclusion}). One promising formulation would train specialist policies per niche via gradient descent while retaining ENA's environment detection and Bayesian policy reuse.

Finally, wall-clock execution times reflect specific software implementations, measured on a single workstation for CartPole and on dedicated cloud instances per replicate for HalfCheetah. While environment steps and candidate steps represent intrinsic properties of the algorithms, wall-clock time also reflects implementation and hardware utilization efficiency; hence, it does not imply that ENA is inherently less computationally demanding across all execution platforms.

\subsection{Limitations}
\begin{itemize}
\item \textit{Two benchmark families.} Both use related physics changes, with a single morphology in each. Visual, partially observable or structurally different tasks are untested.
\item \textit{Fixed task order.} Each domain uses one training sequence (Mars--Jupiter--Earth; Earth--Heavy-Rough--Mars-Slippery). Ordering effects are untested.
\item \textit{No recurrence during training.} Earlier environments return only at the frozen test. Repeated switches, short dwell times and recall during training are untested.
\item \textit{Test-time choice in the ablation.} Without a Hall of Fame, there are no gladiators to choose from. In that case, ENA-NoSkill and ENA-Plasticity play a random member of their final population in every test episode. The variants with an archive choose among gladiators by Bayesian policy reuse. Adding the Hall of Fame thus changes both what is kept and how the test policy is chosen, and the ablation does not separate the two.
\item \textit{Detection in the ablation.} Switching detection on also gives the archive $K$ gladiators per niche instead of $K$ in all. It also resets the return history and the input statistics at each confirmed change. The ablation does not separate these from the niche index (Section~\ref{sec:variants}).
\item \textit{Interaction in the ablation.} The ablation fixes the number of episodes, not of environment steps. A variant with Plasticity takes many times more steps than one without (Section~\ref{sec:ablation}). Without Plasticity, every candidate is also evaluated for only $h_{\min}$ steps, which changes its fitness signal. The ablation shows what failure-driven search achieves with that interaction. It does not show that allocating effort by failure beats spreading the same effort evenly. Full search was run in a single replicate.
\item \textit{Resets.} The chirp requires resetting the environment at episode boundaries, to play its \Res{cfg/cheetah/chirp/steps}-step excitation from nominal initial states. Candidate evaluation needs more than a reset: a resettable copy of the current physics. Both are natural in simulation, digital twins and automated test rigs. They constrain continuous, non-resettable physical systems.
\end{itemize}
\section{Conclusion and Future Work}
\label{sec:conclusion}

This work investigated whether a small set of modular skills can transform the solutions generated by an evolving population into a reusable memory that mitigates catastrophic forgetting. To address this question, the Evolutionary Neuroplastic Agent (ENA) was introduced. Across the evaluated CartPole sequences, the empirical ablation supports this architecture. Only variants with failure-driven search master the more demanding environments, requiring increased environment interaction. Among these variants, only the architecture that indexes its archive by detected physical regimes preserves acquired competencies across switches. Introducing the archive also alters test-time policy selection, an effect that the current ablation does not decouple. Detection also enlarges the archive and resets the agent's statistics at each change, and the ablation does not separate these from the index either. Policy retrieval via Bayesian policy reuse proved significantly more reliable than relying solely on change detection. The memory is constructed online without external supervision, using detection mechanisms tailored to each domain. Whether dynamically modulating search effort based on failure outperforms uniform effort allocation remains an open question. Single-policy learners trained on identical sequences suffer catastrophic forgetting of previously acquired behaviors. Conversely, they highlight regimes where ENA is less competitive: on CartPole, adaptation is slower than PPO, and on HalfCheetah, asymptotic returns are lower than those of SAC. Both limitations align with known properties of evolutionary search, which exhibits high sample complexity and scales unfavorably with parameter count~\cite{khadka2018evolutionary,vemula2019contrasting}.

The most direct next step is a hybrid of ENA with gradient-based learners. Evolution and gradient descent can be combined in multiple configurations, offering several promising avenues. Gradient learners such as SAC or PPO could serve as local optimizers within each niche to accelerate adaptation and achieve higher asymptotic returns. ENA's environment detection, Plasticity, and physics-indexed Hall of Fame would preserve niche-specific competencies across dynamic switches. Alternatively, the architectures could partition the policy network---with gradient descent optimizing high-dimensional feature representations and evolutionary search tuning low-dimensional control layers---or alternate sequentially. The evolutionary population could also supply diverse rollouts to off-policy replay buffers, as in evolutionary reinforcement learning~\cite{khadka2018evolutionary}. A hybrid architecture would furthermore provide a direct comparison against continual-RL frameworks designed for contextual non-stationarity: OWL~\cite{kessler2022same}, RL-CD~\cite{dasilva2006dealing}, MBCD~\cite{alegre2021minimum}, SWOKS~\cite{dick2024statistical}, and TeLAPA~\cite{lillo2026beyond}.

The modular skills may also transfer directly to those frameworks. OWL and TeLAPA depend on explicit task boundaries. Their authors list detecting boundaries during training~\cite{kessler2022same} and weaker task boundaries~\cite{lillo2026beyond} among future directions. ITE initiates adaptation based on a predefined performance threshold and maintains no memory across sequential conditions~\cite{cully2015robots}. ENA's environment detection identifies transitions and recognizes recurring regimes directly from unassisted observations, providing capabilities that existing methods either assume or lack. Furthermore, a physics-indexed archive could preserve adapted policies for deployment when previously encountered conditions recur.

Several further directions remain open. Repeated recurrence during training and randomized task orders would test the memory beyond the evidence here. For systems that cannot be reset, the probe could be superimposed on the agent's own actions as a small dither. An archive of matched capacity without the niche index would isolate what the index contributes. Evaluating candidates for a fixed horizon, whatever the Plasticity, would separate the volume of search from the fidelity of its fitness signal. The archive could also rank its gladiators by a stationary score, such as their own-niche returns on shared seeds, as the Olympiads do on HalfCheetah. The contributions of the belief's two sources of evidence, the dynamics likelihoods and the test returns, remain to be separated. Specialist selection could also be learned, as a contextual bandit over the detector's evidence. Finally, the population's evaluations are independent and parallelize naturally on accelerators, which would allow larger populations and longer non-stationary streams.
\section*{Author Contributions}

Contributions are listed with the CRediT contributor roles.

\noindent\textbf{Blas Eugenio Vicco Zarate:} Conceptualization, Data curation, Formal analysis, Investigation, Methodology, Software, Visualization, Writing -- original draft, Writing -- review \& editing.

\noindent\textbf{Miguel Hern\'an Vicco:} Conceptualization, Formal analysis, Methodology, Writing -- original draft, Writing -- review \& editing.
\section*{Code Availability}

The full implementation of the Evolutionary Neuroplastic Agent, the experiment runners, the baselines, the LaTeX source of this paper and the scripts that generate its tables and figures are publicly available:
\begin{itemize}
    \item \textbf{GitHub Repository:}\\ \url{https://github.com/blasvicco/ena/tree/v1}
\end{itemize}
The results reported here were produced with the tagged release \texttt{v1}, which fixes the code, configuration and raw replicate-level outputs behind every number.
 
\bibliographystyle{IEEEtran}

\appendices
\section{Information Available to Each Agent}
\label{app:information}
A comparison across agent families is only fair if no agent learns more about the environment than the others. This appendix lists everything each agent receives.

\subsection{The interface}
Every agent, ENA and the baselines alike, interacts with an environment only through the Gymnasium interface. A reset returns an initial observation. A step returns the next observation, the reward, and whether the episode terminated or was truncated by the time limit. In CartPole, the observation is the cart's position and velocity, and the pole's angle and angular velocity. In HalfCheetah, it is the torso's height and angle, the joint angles, and all their velocities, without the forward position.

The physical parameters that distinguish the environments are never part of the observation. In CartPole, these are gravity, the pole's mass and the pole's length. In HalfCheetah, they are gravity, friction, wind and the presence of air. No agent is told when the environment changes, which environment it is in, or how many environments there are. The environment names in the figures and tables are taken from the experiment logs, and are used only for analysis. ENA's niches are its own numbering.

\subsection{Task constants}
ENA reads two constants from each task's Gymnasium registration: the episode time limit, and the solved threshold $R_{\text{solved}}$. They are \Res{cfg/cartpole/maxsteps} steps and \Res{cfg/cartpole/solved} in CartPole, and \Res{cfg/cheetah/maxsteps} steps and \Res{cfg/cheetah/solved} in HalfCheetah. The time limit rescales a truncated candidate return to a full-episode equivalent~\eqref{eq:plasticity}. HalfCheetah episodes never end early, and its candidates are evaluated for at most \Res{cfg/cheetah/maxeval} steps, so every HalfCheetah candidate's return is rescaled. The solved threshold is the utility scale of the specialist selection~\eqref{eq:bpr}. The selection therefore depends on this registered constant, even on HalfCheetah, where no agent reaches it. There, the threshold sets how much one return moves the belief, since the return noise of~\eqref{eq:bpr} is \Res{cfg/ena/bpr/noise} times the threshold in return units. Until \Res{cfg/ena/signal/warmup} returns of the current regime have been seen, it is also the scale against which returns are normalized for Trust. After that, Trust normalizes against the percentiles of recent returns~\eqref{eq:normreturn}. Both constants are public properties of the task, not of the physics. The baselines meet the same time limit through the truncation signal.

\subsection{Environment detection}
\label{sec:appendix_detection}
The detector's input is the sequence of transitions $(s_t, a_t, s_{t+1})$: the observation before acting, the action taken, and the observation after. This is exactly what the agent sees. The detector does not read the reward, the physical parameters or the environment's name. A change of physics changes how the observed velocities evolve, so the current niche's dynamics model~\eqref{eq:dynamics} predicts the new transitions worse. In the ablation, every switch between training environments is confirmed in every replicate, within a few episodes. False alarms also occur, mostly in the first training phase (Table~\ref{tab:detection}).

At test, the detector is frozen and creates no niches. Physics close to a known niche are recognized as that niche. The detector does not need to name an environment, only to find the closest physics it knows. Neptune's physics, for instance, are recognized as Earth's in \Res{ablation/ENA-Combined/recog/Neptune/Earth-Standard} of the \Res{ablation/n} replicates. When no niche explains the transitions within the margin $m$, the episode is left unrecognized, and no niche is assigned. The selection~\eqref{eq:bpr} still weighs every niche by how well its dynamics model explains the episode, so it falls back on the niches whose models explain the episode best.

Recognition is weaker in two cases:
\begin{itemize}
\item \textit{False alarms.} In a test phase, every confirmed change except the first one after a switch is a false alarm. In physics near the margin $m$, the detector can alternate between recognizing a niche and leaving the physics unrecognized. Each confirmation resets the selection's belief to uniform.
\item \textit{Fragmentation.} In some replicates, one training environment is split over two niches.
\end{itemize}
Table~\ref{tab:detection} gives these numbers for the two ablation variants with environment detection. For each phase, it lists the share of unrecognized episodes, the confirmed changes and the episode of the first one. For each training environment, it lists the Hall of Fame niches. Every phase except the run's first follows a real switch. At most one confirmed change per phase is therefore a detection, and the rest are false alarms. The first training phase, Mars, starts with no niche at all. The detector is also reset when the test begins, so the test's Mars phase starts without a current niche, like the run's first. The Brown-dwarf phase is mostly left unrecognized. After it, the detector recognizes Earth by identification, without confirming a change, and Earth shows zero confirmed changes at test.

\begin{table}[htbp]
\caption{Environment Recognition in the Skill Ablation (30 Replicates)}
\begin{center}
\footnotesize
\setlength{\tabcolsep}{3pt}
\begin{adjustbox}{max width=\columnwidth}
\begin{tabular}{|l|l|l|c|c|c|c|}
\hline
\textbf{Agent} & \textbf{Phase} & \textbf{Env.} & \textbf{Unrec. (\%)} & \textbf{Changes} & \textbf{First (ep.)} & \textbf{Niches} \\
\hline
ENA-HoF+EnvDetection & Training & Mars & 0.4 $\pm$ 0.4 & 1.3 $\pm$ 1.9 & 243.0 $\pm$ 138.7 & 1.6 $\pm$ 0.6 \\
ENA-HoF+EnvDetection & Training & Jupiter & 0.2 $\pm$ 0.0 & 1.1 $\pm$ 0.3 & 1.0 $\pm$ 1.0 & 1.1 $\pm$ 0.3 \\
ENA-HoF+EnvDetection & Training & Earth & 0.1 $\pm$ 0.2 & 1.2 $\pm$ 0.4 & 0.5 $\pm$ 0.7 & 1.2 $\pm$ 0.4 \\
ENA-HoF+EnvDetection & Test & Mars & 9.8 $\pm$ 24.3 & 2.0 $\pm$ 10.6 & 6.3 $\pm$ 4.2 & -- \\
ENA-HoF+EnvDetection & Test & Neptune & 1.9 $\pm$ 5.7 & 2.4 $\pm$ 5.8 & 0.2 $\pm$ 0.5 & -- \\
ENA-HoF+EnvDetection & Test & Jupiter & 0.8 $\pm$ 0.8 & 1.0 $\pm$ 0.0 & 0.2 $\pm$ 0.4 & -- \\
ENA-HoF+EnvDetection & Test & Brown & 96.5 $\pm$ 11.4 & 6.8 $\pm$ 11.3 & 0.2 $\pm$ 0.5 & -- \\
ENA-HoF+EnvDetection & Test & Earth & 1.0 $\pm$ 0.9 & 0.0 $\pm$ 0.0 & -- & -- \\
\hline
ENA-Combined & Training & Mars & 0.3 $\pm$ 0.3 & 1.3 $\pm$ 2.5 & 202.4 $\pm$ 174.3 & 1.4 $\pm$ 0.5 \\
ENA-Combined & Training & Jupiter & 0.2 $\pm$ 0.1 & 1.1 $\pm$ 0.4 & 0.8 $\pm$ 0.9 & 1.1 $\pm$ 0.4 \\
ENA-Combined & Training & Earth & 0.0 $\pm$ 0.0 & 1.3 $\pm$ 0.5 & 0.0 $\pm$ 0.0 & 1.3 $\pm$ 0.5 \\
ENA-Combined & Test & Mars & 1.1 $\pm$ 1.1 & 0.8 $\pm$ 1.7 & 29.5 $\pm$ 68.8 & -- \\
ENA-Combined & Test & Neptune & 0.8 $\pm$ 1.1 & 1.2 $\pm$ 1.1 & 0.3 $\pm$ 0.5 & -- \\
ENA-Combined & Test & Jupiter & 0.6 $\pm$ 0.3 & 1.0 $\pm$ 0.0 & 0.2 $\pm$ 0.4 & -- \\
ENA-Combined & Test & Brown & 92.6 $\pm$ 12.6 & 16.4 $\pm$ 18.8 & 0.0 $\pm$ 0.0 & -- \\
ENA-Combined & Test & Earth & 0.4 $\pm$ 0.2 & 0.0 $\pm$ 0.0 & -- & -- \\
\hline
\end{tabular}
\end{adjustbox}
\label{tab:detection}
\end{center}
\end{table}
 
On HalfCheetah, ENA detects with its chirp (Section~\ref{sec:chirp}). The chirp's input is the agent's own fixed action sequence, and the observations it produces after a reset of the environment. These are the same observations the agent sees, so the chirp reads no reward, physical parameter or environment name either. Its steps count in ENA's step budget. The environment's random stream is restored after it, so every agent's episodes start from the same states. Table~\ref{tab:cheetah_detection} gives the same numbers for ENA-Combined in the HalfCheetah study.

\begin{table}[htbp]
\caption{Environment Recognition on HalfCheetah (20 Replicates)}
\begin{center}
\footnotesize
\setlength{\tabcolsep}{3pt}
\begin{adjustbox}{max width=\columnwidth}
\begin{tabular}{|l|l|l|c|c|c|c|}
\hline
\textbf{Agent} & \textbf{Phase} & \textbf{Env.} & \textbf{Unrec. (\%)} & \textbf{Changes} & \textbf{First (ep.)} & \textbf{Niches} \\
\hline
ENA-Combined & Training & Earth & 0.0 $\pm$ 0.0 & 0.0 $\pm$ 0.0 & -- & 1.0 $\pm$ 0.0 \\
ENA-Combined & Training & Heavy & 1.0 $\pm$ 0.7 & 2.1 $\pm$ 1.0 & 0.0 $\pm$ 0.0 & 1.4 $\pm$ 0.6 \\
ENA-Combined & Training & Mars & 1.3 $\pm$ 0.6 & 1.8 $\pm$ 0.8 & 0.7 $\pm$ 2.3 & 1.3 $\pm$ 0.6 \\
ENA-Combined & Test & Earth & 0.0 $\pm$ 0.0 & 0.0 $\pm$ 0.0 & -- & -- \\
ENA-Combined & Test & Ice & 52.2 $\pm$ 13.0 & 6.3 $\pm$ 1.5 & 1.4 $\pm$ 0.5 & -- \\
ENA-Combined & Test & Heavy & 1.1 $\pm$ 1.0 & 0.5 $\pm$ 0.5 & 6.2 $\pm$ 14.7 & -- \\
ENA-Combined & Test & High & 2.5 $\pm$ 0.9 & 1.0 $\pm$ 0.0 & 0.8 $\pm$ 0.7 & -- \\
ENA-Combined & Test & Mars & 4.3 $\pm$ 1.5 & 1.2 $\pm$ 0.4 & 0.1 $\pm$ 0.2 & -- \\
\hline
\end{tabular}
\end{adjustbox}
\label{tab:cheetah_detection}
\end{center}
\end{table}
 
\subsection{Candidate evaluation}
ENA evaluates candidate specialists in fresh instances of the current environment. The experiment's own environment factory builds them from the current environment's physical parameters. ENA passes these parameters to the factory, but uses none of them for a decision. Every candidate step counts in ENA's interaction budget. A candidate evaluation is thus equivalent to an extra episode in the current environment, with the same reset that any episodic agent has. The baselines could spend the same budget in parallel copies of the environment, as vectorized training does. What ENA does assume is a resettable copy of the current physics, in which it can run episodes other than the one being played. Such a copy is free in simulation or with a digital twin. On a physical system, it would cost real time, if it can be had at all.

\subsection{Evaluation and test}
The first-solve checks (Section~\ref{sec:forgetting_protocol}) run on a separate copy of the current environment, for every agent alike. Their steps do not count as training. At test, every agent is frozen, and Section~\ref{sec:selection} lists what still adapts in ENA. Every agent of a replicate meets the same seeded environments.

\section{Hyperparameters}
\label{app:hyperparameters}

All settings are read directly from the code the studies ran, by \texttt{src/tools/paper\_config.py}. Each value is an agent's class default, unless its study script overrides it. The values here therefore cannot drift from the code.

Many of ENA's settings follow from other settings, from the task, or from a study:
\begin{itemize}
\item \textit{Fixed by other settings.} The slope of the linear ramp is $1 - f_{\min}$~\eqref{eq:ramp}. On CartPole, the longest candidate evaluation $h_{\max}$ is the time limit $t_{\max}$. The shares of the population replaced by mutants, crossovers and fresh specialists follow from the replaced share and its split (Section~\ref{sec:search}).
\item \textit{Read from the task.} The time limit $t_{\max}$ and the solved return $R_{\text{solved}}$ come from each task's Gymnasium registration (Appendix~\ref{app:information}).
\item \textit{Chosen or checked by a study.} The Plasticity controller was compared with two alternatives, and kept by a rule fixed before the runs (Section~\ref{sec:controller}). HalfCheetah's chirp was chosen by the detection ablation (Section~\ref{sec:detection_benchmark}). The detector's tolerance $\tau$ and threshold $\Gamma$ were set on preliminary CartPole runs, before the study, so that an unchanged environment raised no alarm (Section~\ref{sec:detection}). ENA's other HalfCheetah settings were chosen on tuning seeds disjoint from the study's replicates.
\end{itemize}
The paper reports no systematic search over ENA's other settings.

Table~\ref{tab:hparams_ena_cartpole} gives the settings of ENA-Combined, as built by \texttt{cartpole\_ena\_ablation.py}. All six ablation variants share them, with the relevant skills switched on or off. Table~\ref{tab:hparams_ena_cheetah} lists only the settings that differ on HalfCheetah, set in \texttt{half\_cheetah.py}.

On HalfCheetah, PPO and SAC take the settings of RL Baselines3 Zoo~\cite{raffin2020rlzoo} (Table~\ref{tab:hparams_cheetah_baselines}). Both have a separate actor and critic. PPO normalizes observations, and scales rewards by the standard deviation of the return. PPO keeps the zoo's constant learning rate. One setting deviates from the zoo: SAC applies its gradient steps in groups of \Res{cfg/cheetahsac/gradientsteps} every \Res{cfg/cheetahsac/updatefrequency} environment steps, for speed, which preserves the zoo's ratio of one update per step.

\begin{table}[htbp]
\caption{ENA Hyperparameters (CartPole)}
\begin{center}\footnotesize
\begin{adjustbox}{max width=\columnwidth}
\begin{tabular}{|l|r|l|r|}
\hline
\textbf{Parameter} & \textbf{Value} & \textbf{Parameter} & \textbf{Value} \\
\hline
Network (hidden) & \Res{cfg/ena/brain} & Population size $N$ & \Res{cfg/ena/pop} \\
Eval depth $h_{\min}$ / $h_{\max}$ & \Res{cfg/ena/smin} / \Res{cfg/ena/smax} & Min.\ candidates $k_{\min}$ & \Res{cfg/ena/search/mincandidates} \\
Return history $|H|$ & \Res{cfg/ena/history} & Exploration $\epsilon$ & \Res{cfg/ena/epsilon} \\
Signal floor $f_{\min}$ & \Res{cfg/ena/signal/min} & & \\
Trust percentiles & \Res{cfg/ena/trust/plow} / \Res{cfg/ena/trust/phigh} & Trust clip & $[\Res{cfg/ena/trust/low}, \Res{cfg/ena/trust/high}]$ \\
Trust blend threshold & \Res{cfg/ena/trust/reset} & Signal warm-up (returns) & \Res{cfg/ena/signal/warmup} \\
Signal percentile & \Res{cfg/ena/signal/percentile}th & Plasticity controller & linear ramp \\
Mutation rate $p_m$ & \Res{cfg/ena/mutation/rate} & Mutation noise $\sigma_m$ / floor & \Res{cfg/ena/mutation/noise} / \Res{cfg/ena/mutation/minnoise} \\
Crossover parent share & \Res{cfg/ena/crossover/share} & Elite (\% of pop.) & \Res{cfg/ena/search/elite} \\
Mutants (\% of pop.) & \Res{cfg/ena/search/mutantspop} & Crossovers (\% of pop.) & \Res{cfg/ena/search/crossoverspop} \\
HoF $K$ (max per niche) & \Res{cfg/ena/hof/K} & HoF reinject ratio & \Res{cfg/ena/hof/reinject} \\
HoF similarity threshold & \Res{cfg/ena/hof/similarity} & HoF gate $z$ / relief & \Res{cfg/ena/hof/gatez} / \Res{cfg/ena/hof/gaterelief} \\
HoF gate $\sigma_T$ floor & \Res{cfg/ena/hof/gatefloor} & HoF gate: sufficient Trust & \Res{cfg/ena/hof/gatetrust} \\
Detection tolerance $\tau$ & \Res{cfg/ena/detect/tau} & Detection margin $m$ & \Res{cfg/ena/detect/margin} \\
Detection recent window & \Res{cfg/ena/detect/recent} & Min.~samples to fit & \Res{cfg/ena/detect/minsamples} \\
Verification samples & \Res{cfg/ena/detect/verify} & CuSum false-alarm run $\Gamma$ & $10^{\Res{cfg/ena/detect/gammaexp}}$ \\
BPR return noise & \Res{cfg/ena/bpr/noise} & BPR switch rate & \Res{cfg/ena/bpr/switch} \\
BPR prior strength & \Res{cfg/ena/bpr/prior} & BPR prior own / other niche & \Res{cfg/ena/bpr/ownprior} / \Res{cfg/ena/bpr/crossprior} \\
\hline
\end{tabular}
\end{adjustbox}
\label{tab:hparams_ena_cartpole}
\end{center}
\end{table}

\begin{table}[htbp]
\caption{ENA Hyperparameters That Differ on HalfCheetah}
\begin{center}\footnotesize
\begin{adjustbox}{max width=\columnwidth}
\begin{tabular}{|l|r|l|r|}
\hline
\textbf{Parameter} & \textbf{Value} & \textbf{Parameter} & \textbf{Value} \\
\hline
Network (hidden) & \Res{cfg/cheetahena/brain}, normc & Eval depth $h_{\max}$ & \Res{cfg/cheetahena/maxevalsteps} \\
Mutation rate $p_m$ & \Res{cfg/cheetahena/mutationrate} & HoF $K$ (max per niche) & \Res{cfg/cheetahena/maxperniche} \\
Hall of Fame & \Res{cfg/cheetahena/halloffamedriver} & Olympiad interval (steps) & \Res{cfg/cheetahena/olympicsinterval} \\
Olympiad episodes & \Res{cfg/cheetahena/olympicsepisodes} & Qualifier seats per niche & \Res{cfg/cheetahena/olympicsqualifierseats} \\
Environment detection & chirp & Chirp steps & \Res{cfg/cheetah/chirp/steps} \\
Chirp excitation & \Res{cfg/cheetah/chirp/excitation} (\Res{cfg/cheetah/chirp/excitationwavelet}) & Chirp transform & \Res{cfg/cheetah/chirp/transform} \\
Chirp wavelet & \Res{cfg/cheetah/chirp/featurewavelet} & Chirp level / bands & \Res{cfg/cheetah/chirp/level} / \Res{cfg/cheetah/chirp/bands} \\
Chirp window (episodes) & \Res{cfg/cheetah/chirp/window} & Chirp min chirps & \Res{cfg/cheetah/chirp/minchirps} \\
Chirp confirm margin & \Res{cfg/cheetah/chirp/margin} & & \\
\hline
\end{tabular}
\end{adjustbox}
\label{tab:hparams_ena_cheetah}
\end{center}
\end{table}

\begin{table}[htbp]
\caption{DQN Hyperparameters (CartPole)}
\begin{center}\footnotesize
\begin{adjustbox}{max width=\columnwidth}
\begin{tabular}{|l|r|l|r|}
\hline
\textbf{Parameter} & \textbf{Value} & \textbf{Parameter} & \textbf{Value} \\
\hline
Learning rate & \Res{cfg/dqn/learningrate} & Discount $\gamma$ & \Res{cfg/dqn/gamma} \\
Batch size & \Res{cfg/dqn/batchsize} & Replay buffer & \Res{cfg/dqn/buffersize} \\
$\varepsilon$ start / end & \Res{cfg/dqn/epsilonstart} / \Res{cfg/dqn/epsilonend} & $\varepsilon$ decay steps & \Res{cfg/dqn/epsilondecaysteps} \\
Target network update (steps) & \Res{cfg/dqn/targetupdatefreq} & Learning starts (steps) & \Res{cfg/dqn/learningstarts} \\
Update every (steps) & \Res{cfg/dqn/trainfreq} & Network & \Res{cfg/challenge/mlpnet} \\
\hline
\end{tabular}
\end{adjustbox}
\label{tab:hparams_dqn}
\end{center}
\end{table}

\begin{table}[htbp]
\caption{PPO Hyperparameters (CartPole)}
\begin{center}\footnotesize
\begin{adjustbox}{max width=\columnwidth}
\begin{tabular}{|l|r|l|r|}
\hline
\textbf{Parameter} & \textbf{Value} & \textbf{Parameter} & \textbf{Value} \\
\hline
Learning rate & \Res{cfg/ppo/learningrate} & Learning rate anneal & \Res{cfg/ppo/anneallr} \\
Steps per update & \Res{cfg/ppo/nsteps} & Batch size & \Res{cfg/ppo/batchsize} \\
Epochs per update & \Res{cfg/ppo/nepochs} & Discount $\gamma$ & \Res{cfg/ppo/gamma} \\
GAE $\lambda$ & \Res{cfg/ppo/gaelambda} & Clip ratio & \Res{cfg/ppo/clipratio} \\
Entropy coefficient & \Res{cfg/ppo/entropycoef} & Value coefficient & \Res{cfg/ppo/valuecoef} \\
Max grad norm & \Res{cfg/ppo/maxgradnorm} & Actor / critic & \Res{cfg/challenge/pponet} \\
Initialization & orthogonal & Normalization & none \\
\hline
\end{tabular}
\end{adjustbox}
\label{tab:hparams_ppo_cartpole}
\end{center}
\end{table}

\begin{table}[htbp]
\caption{SAC-Discrete Hyperparameters (CartPole)}
\begin{center}\footnotesize
\begin{adjustbox}{max width=\columnwidth}
\begin{tabular}{|l|r|l|r|}
\hline
\textbf{Parameter} & \textbf{Value} & \textbf{Parameter} & \textbf{Value} \\
\hline
Learning rate & \Res{cfg/sacdiscrete/learningrate} & Discount $\gamma$ & \Res{cfg/sacdiscrete/gamma} \\
Batch size & \Res{cfg/sacdiscrete/batchsize} & Replay buffer & \Res{cfg/sacdiscrete/buffercapacity} \\
Learning starts (steps) & \Res{cfg/sacdiscrete/learningstarts} & Update every (steps) & \Res{cfg/sacdiscrete/trainfreq} \\
Target update $\tau$ & \Res{cfg/sacdiscrete/tau} & Target entropy & \Res{cfg/sacdiscrete/targetentropyratio}\,$\times\,\log|\mathcal{A}|$ \\
Actor / twin critics & \Res{cfg/challenge/mlpnet} & & \\
\hline
\end{tabular}
\end{adjustbox}
\label{tab:hparams_sacd}
\end{center}
\end{table}

\begin{table}[htbp]
\caption{PPO and SAC Hyperparameters (HalfCheetah)}
\begin{center}\footnotesize
\begin{adjustbox}{max width=\columnwidth}
\begin{tabular}{|l|r|r|}
\hline
\textbf{Parameter} & \textbf{PPO} & \textbf{SAC} \\
\hline
Actor / critic & \Res{cfg/cheetah/baselinenet} & \Res{cfg/cheetah/baselinenet} \\
Learning rate & \Res{cfg/cheetahppo/learningrate} & \Res{cfg/cheetahsac/learningrate} \\
Discount $\gamma$ & \Res{cfg/cheetahppo/gamma} & \Res{cfg/cheetahsac/gamma} \\
Batch size & \Res{cfg/cheetahppo/batchsize} & \Res{cfg/cheetahsac/batchsize} \\
Steps per update / buffer & \Res{cfg/cheetahppo/nsteps} & \Res{cfg/cheetahsac/buffercapacity} \\
Epochs / gradient steps & \Res{cfg/cheetahppo/nepochs} & \Res{cfg/cheetahsac/gradientsteps} per \Res{cfg/cheetahsac/updatefrequency} steps \\
GAE $\lambda$ & \Res{cfg/cheetahppo/gaelambda} & --- \\
Clip ratio & \Res{cfg/cheetahppo/clipratio} & --- \\
Entropy coef.~/ temperature & \Res{cfg/cheetahppo/entropycoef} & auto ($-\dim\mathcal{A}$) \\
Value coef.~/ target update $\tau$ & \Res{cfg/cheetahppo/valuecoef} & \Res{cfg/cheetahsac/tau} \\
Max grad norm & \Res{cfg/cheetahppo/maxgradnorm} & --- \\
Initial log std & \Res{cfg/cheetahppo/logstdinit} & --- \\
Normalization & obs., reward & none \\
Learning starts (steps) & --- & \Res{cfg/cheetahsac/learningstarts} \\
\hline
\end{tabular}
\end{adjustbox}
\label{tab:hparams_cheetah_baselines}
\end{center}
\end{table}

\section{Compute Environment}
\label{app:compute}

\subsection*{CartPole experiments (all studies except HalfCheetah)}
All CartPole studies ran inside a Docker container (\texttt{hub/python/Dockerfile}) on the host machine:
\begin{itemize}
  \item \textbf{Host:} Apple MacBook Pro, Apple M1~Pro chip, 8 cores (6 performance + 2 efficiency), 16\,GB RAM.
  \item \textbf{Container OS:} Linux (ARM64, via Docker Desktop on macOS).
  \item \textbf{Cores:} the runner used 6 of the 8 cores, running up to 6 replicates in parallel, one process each. Within a replicate, the agents trained one after another, and ENA evaluated its candidates on one thread. Wall-clock times are therefore measured under this shared load.
  \item \textbf{Python:} \Res{cfg/software/python} (Docker image \texttt{python:\Res{cfg/software/python}-slim}).
  \item \textbf{Key libraries (pinned):} TensorFlow \Res{cfg/software/tensorflow}, Gymnasium[MuJoCo] \Res{cfg/software/gymnasium}, MuJoCo \Res{cfg/software/mujoco}, NumPy \Res{cfg/software/numpy}, Keras \Res{cfg/software/keras}, SciPy \Res{cfg/software/scipy}, PyWavelets \Res{cfg/software/pywavelets}. Full list in \texttt{app/requirements.txt}.
  \item \textbf{Seeds} (Section~\ref{sec:setup}): \Res{seeds/ablation/first}--\Res{seeds/ablation/last} for the ablation and for the Plasticity-controller and selection studies; \Res{seeds/challenge/first}--\Res{seeds/challenge/last} for the comparison with single-policy learners.
\end{itemize}

\subsection*{HalfCheetah experiment (cloud run)}
The HalfCheetah comparison ran in AWS (eu-west-1), on one \textbf{c8g.xlarge} instance per replicate, \Res{cheetah/n} instances in all:\footnote{Instance type defined in \texttt{infra/src/qos/qos.json}.}
\begin{itemize}
  \item \textbf{Instance:} AWS c8g.xlarge (AWS Graviton4, ARM64).
  \item \textbf{Container:} Linux aarch64 (ARM64 Docker image, \texttt{hub/cloud/Dockerfile}), built from the same base image and pinned \texttt{requirements.txt} as the CartPole environment.
  \item \textbf{Hardware and threads, as each replicate recorded them:} \Res{cheetah/hardware}.
  \item \textbf{Seeds} (Section~\ref{sec:setup}): \Res{seeds/cheetah/first}--\Res{seeds/cheetah/last}.
\end{itemize}

\section{Extended Evaluation and Training Dynamics}
\label{app:dynamics}

This appendix shows the episode-by-episode test curves of both domains, and HalfCheetah's training curves. The numbers they summarize are in Section~\ref{sec:results}.

\subsection{Frozen Test on CartPole}
\begin{figure*}[t]
\centerline{\includegraphics[width=0.85\textwidth]{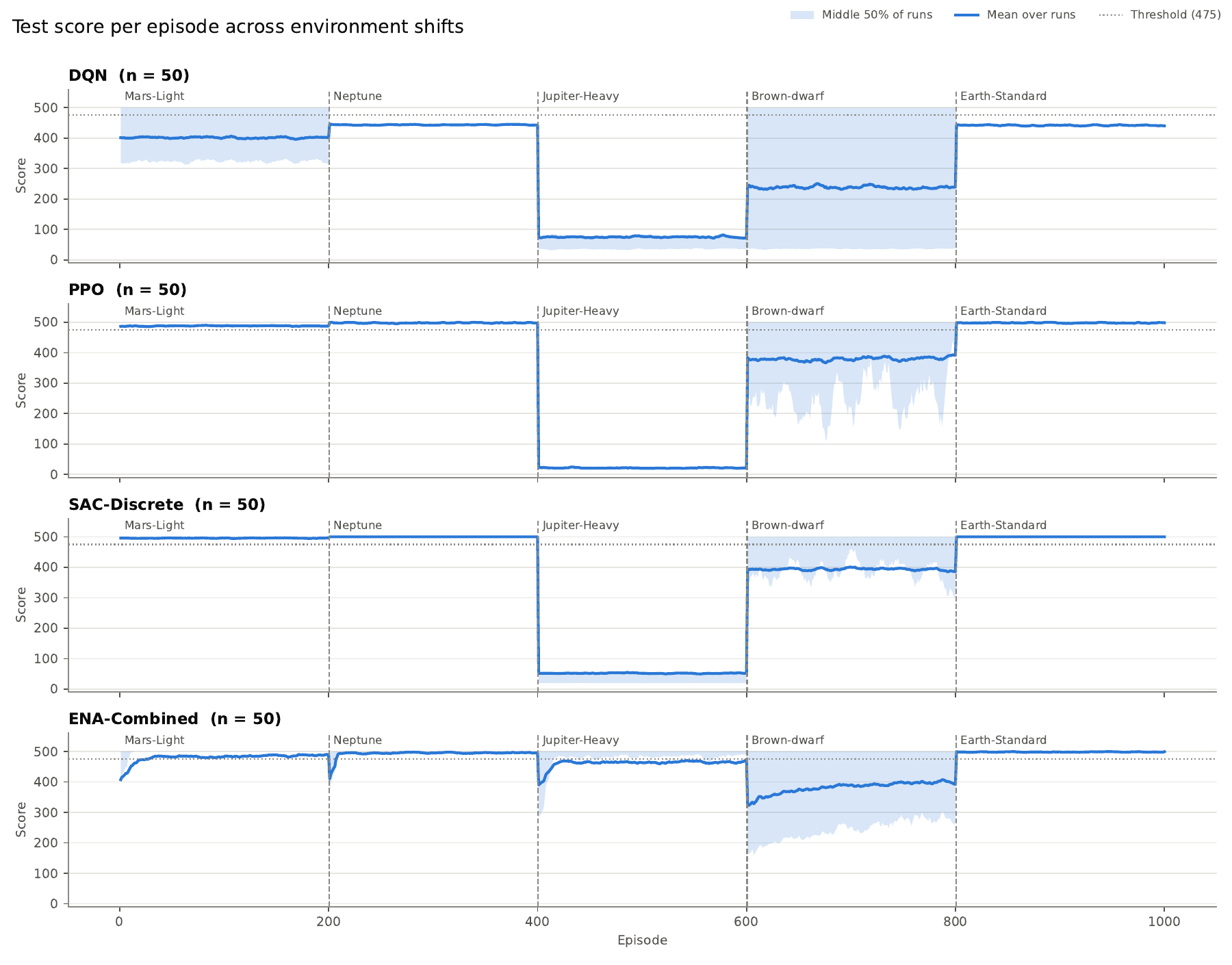}}
\caption{Frozen test on CartPole: mean return per episode across \Res{challenge/n} replicates (line) and the middle \Res{cfg/plot/band}\% of replicates (band). Dashed lines mark environment switches; Neptune and Brown-dwarf were never seen in training.}
\label{fig:challenge_curves}
\end{figure*}

Fig.~\ref{fig:challenge_curves} shows the frozen test of the CartPole comparison. Each agent plays \Res{cfg/cartpole/test} episodes in each of five environments: Mars-Light $\to$ Neptune $\to$ Jupiter-Heavy $\to$ Brown-dwarf $\to$ Earth-Standard. The line is the mean return over the \Res{challenge/n} replicates, and the band the middle \Res{cfg/plot/band}\% of replicates. On Jupiter-Heavy, the environment trained just before the last, the single-policy learners' returns stay low for the whole phase. ENA-Combined's return there stays close to the success threshold. On Brown-dwarf, ENA-Combined's mean return rises during the phase. The factors driving this increase were not investigated.

\subsection{Frozen Test on HalfCheetah}
\begin{figure*}[t]
\centerline{\includegraphics[width=0.85\textwidth]{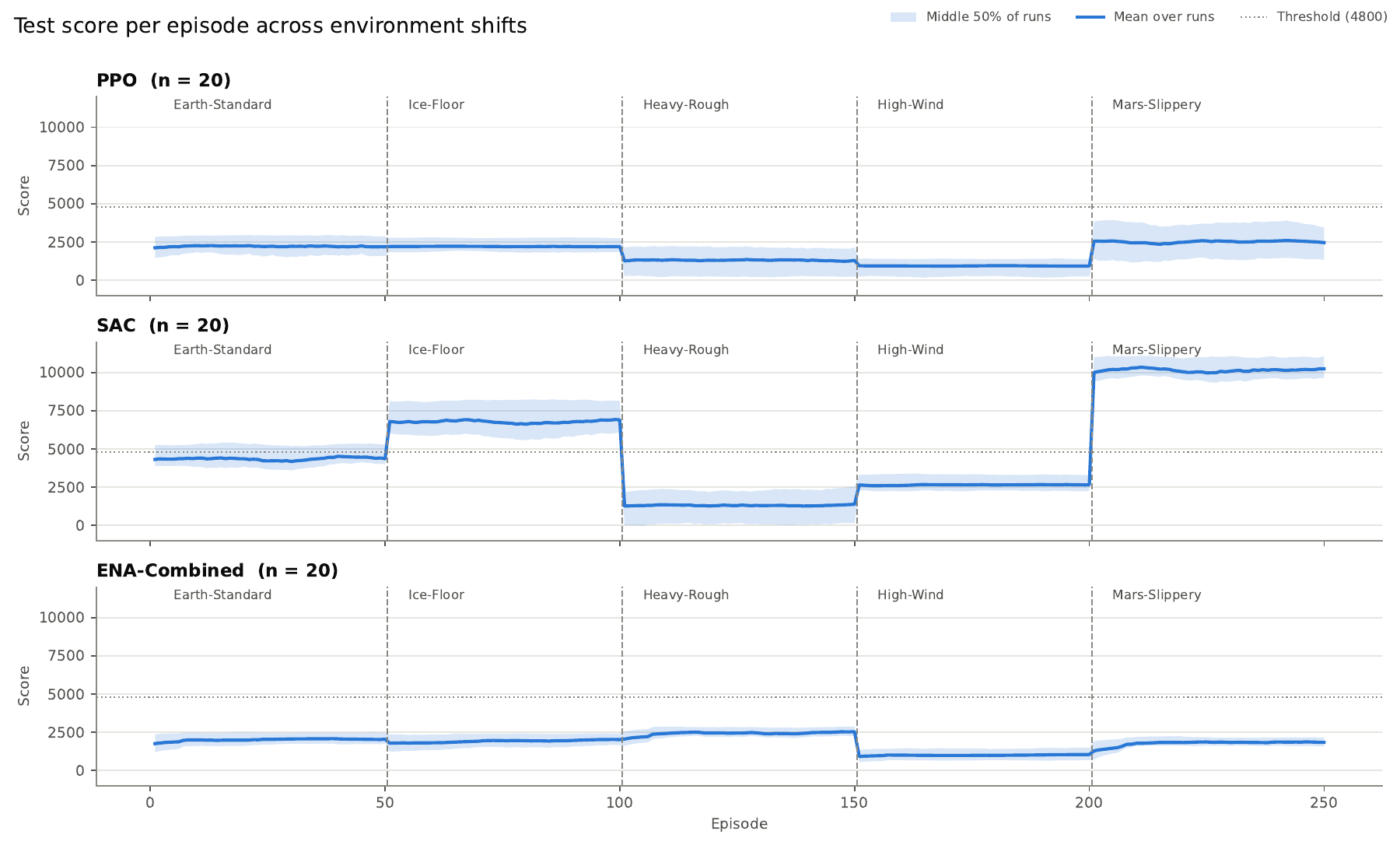}}
\caption{Frozen test on HalfCheetah: mean return per episode across \Res{cheetah/n} replicates (line) and the middle \Res{cfg/plot/band}\% of replicates (band). Dashed vertical lines mark environment switches; Ice-Floor and High-Wind were never seen in training.}
\label{fig:cheetah_test_curves}
\end{figure*}

Fig.~\ref{fig:cheetah_test_curves} shows the frozen test on HalfCheetah, over the \Res{cheetah/n} replicates. Each agent plays \Res{cfg/cheetah/test} episodes in each of five environments: Earth-Standard $\to$ Ice-Floor $\to$ Heavy-Rough $\to$ High-Wind $\to$ Mars-Slippery. On Heavy-Rough, the environment trained just before the last, PPO's and SAC's returns drop. ENA-Combined's return there is its highest. Among the unseen environments, SAC's return on Ice-Floor is far above the other agents', and ENA-Combined's lowest return is on High-Wind.

\subsection{Training Dynamics on HalfCheetah}
\begin{figure*}[t]
\centerline{\includegraphics[width=0.85\textwidth]{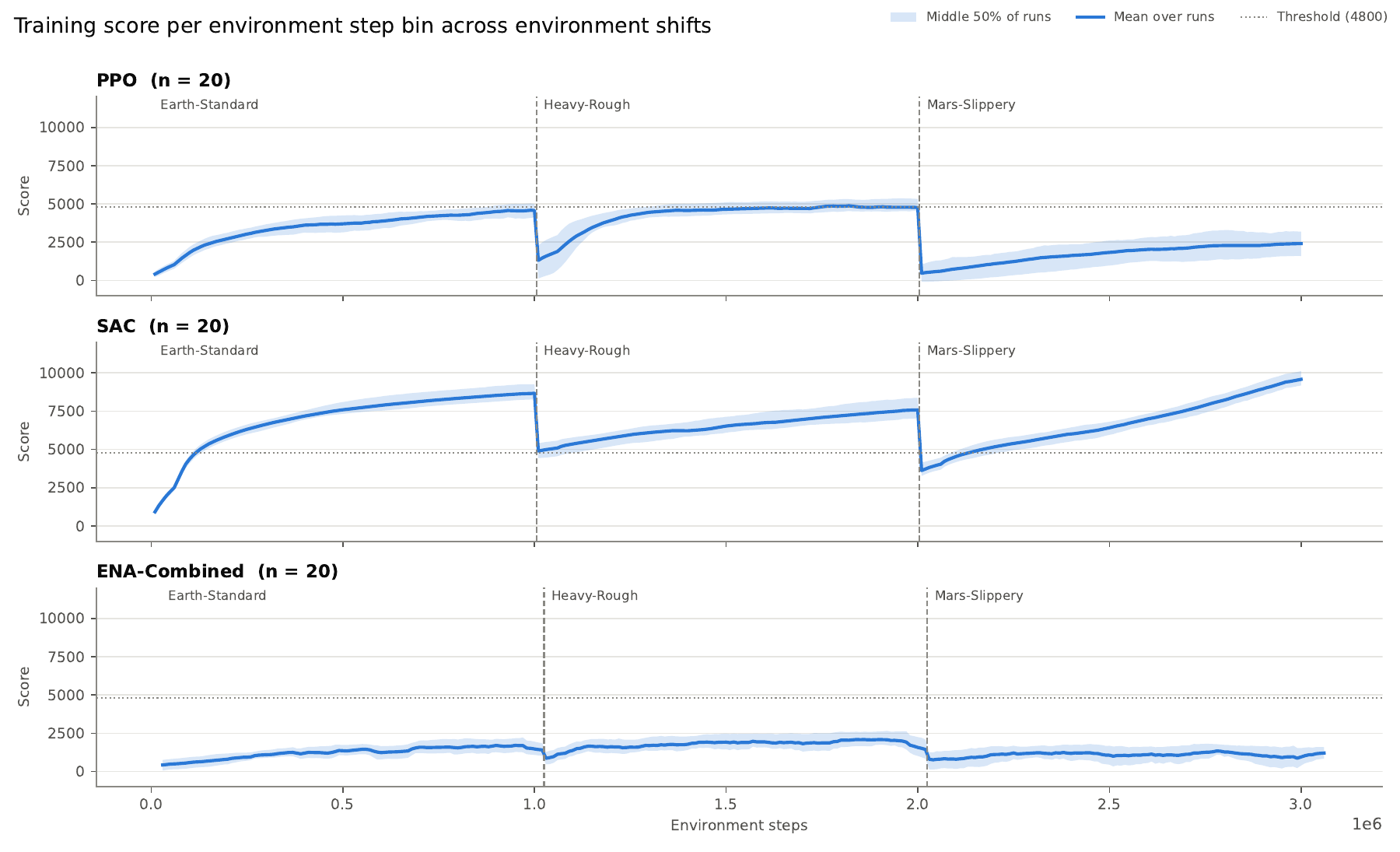}}
\caption{Training return on HalfCheetah against environment steps (\Res{cheetah/n} replicates; line is the mean, band the middle \Res{cfg/plot/band}\% of replicates). Dashed vertical lines mark the switches between the three training environments (Earth-Standard $\to$ Heavy-Rough $\to$ Mars-Slippery).}
\label{fig:cheetah_training_curves}
\end{figure*}

Fig.~\ref{fig:cheetah_training_curves} shows the training return of the three agents against environment steps, \Res{cfg/cheetah/budget} per training environment. The returns are binned and averaged over the \Res{cheetah/n} replicates. SAC's return climbs within each phase, drops at each switch, and climbs again. It ends its three phases at \Res{cheetah/SAC/end/Earth-Standard/mean}, \Res{cheetah/SAC/end/Heavy-Rough/mean} and \Res{cheetah/SAC/end/Mars-Slippery/mean}. ENA-Combined's training return stays far lower. It ends its phases at \Res{cheetah/ENA-Combined/end/Earth-Standard/mean}, \Res{cheetah/ENA-Combined/end/Heavy-Rough/mean} and \Res{cheetah/ENA-Combined/end/Mars-Slippery/mean}.

\end{document}